\documentclass[11pt]{article}
\pdfoutput=1
\usepackage{pdfsync}
\usepackage[OT1]{fontenc}
\usepackage{subfigure}
\usepackage{mathtools}

\usepackage{amsmath,amssymb}
\usepackage{bm}
\usepackage[usenames]{color}
\usepackage{amsthm}
\usepackage{algorithm}
\usepackage{algorithmic}
\usepackage{booktabs}
\usepackage{multirow} 
\usepackage{wrapfig}

\usepackage[colorlinks,
linkcolor=red,
anchorcolor=blue,
citecolor=blue
]{hyperref}

\usepackage{mathrsfs}

\usepackage{fullpage}
\usepackage[protrusion=true,expansion=true]{microtype}

\usepackage{smile}
\usepackage{bbm}

\usepackage{caption} 
\newcommand{\ie}{\textit{i}.\textit{e}.}
\newcommand{\eg}{\textit{e}.\textit{g}.}
\newcommand{\st}{\textit{s}.\textit{t}.}

\newcommand{\ifcomments}{\iftrue}

\newcommand{\tabincell}[2]{\begin{tabular}{@{}#1@{}}#2\end{tabular}}

\usepackage{xcolor,colortbl}
\definecolor{DarkBlue}{RGB}{22,54,93}

\newcommand{\ignore}[1]{}

\begin{document}
\title{\huge Attacking Adversarial Attacks as A Defense}

\author
{   Boxi Wu \thanks{This work was done during an internship at Tencent.} \thanks{College of Computer Science, Zhejiang University, China.} ,  
	~~Heng Pan \footnotemark[1]~\footnotemark[2] ,
	~~Li Shen \thanks{JD Explore Academy, China.},
	~~Jindong Gu \footnotemark[1]~\thanks{University of Munich.},
	~~Shuai Zhao \footnotemark[2], \\
	~~Zhifeng Li \thanks{Tencent Data Platform, China.},
	~~Deng Cai \footnotemark[2],
	~~Xiaofei He \footnotemark[2],
	~~Wei Liu \footnotemark[5]
}
\date{}
 
\maketitle

\begin{abstract}
It is well known that adversarial attacks can fool deep neural networks with imperceptible perturbations. Although adversarial training significantly improves model robustness, failure cases of defense still broadly exist. In this work, we find that the adversarial attacks can also be vulnerable to small perturbations. Namely, on adversarially-trained models, perturbing adversarial examples with a small random noise may invalidate their misled predictions. After carefully examining state-of-the-art attacks of various kinds, we find that all these attacks have this deficiency to different extents. Enlightened by this finding, we propose to counter attacks by crafting more effective defensive perturbations. Our defensive perturbations leverage the advantage that adversarial training endows the ground-truth class with smaller local Lipschitzness. By simultaneously attacking all the classes, the misled predictions with larger Lipschitzness can be flipped into correct ones. We verify our defensive perturbation with both empirical experiments and theoretical analyses on a linear model. On CIFAR10, it boosts the state-of-the-art model from $66.16\%$ to $72.66\%$ against the four attacks of AutoAttack, including $71.76\%$ to $83.30\%$ against the Square attack. On ImageNet, the top-1 robust accuracy of FastAT is improved from $33.18\%$ to $38.54\%$ under the 100-step PGD attack.

\end{abstract}

\section{Introduction}
\label{sec:intro}

\begin{wrapfigure}{r}{0.48\linewidth}
	\centering
	\includegraphics[width=\linewidth]{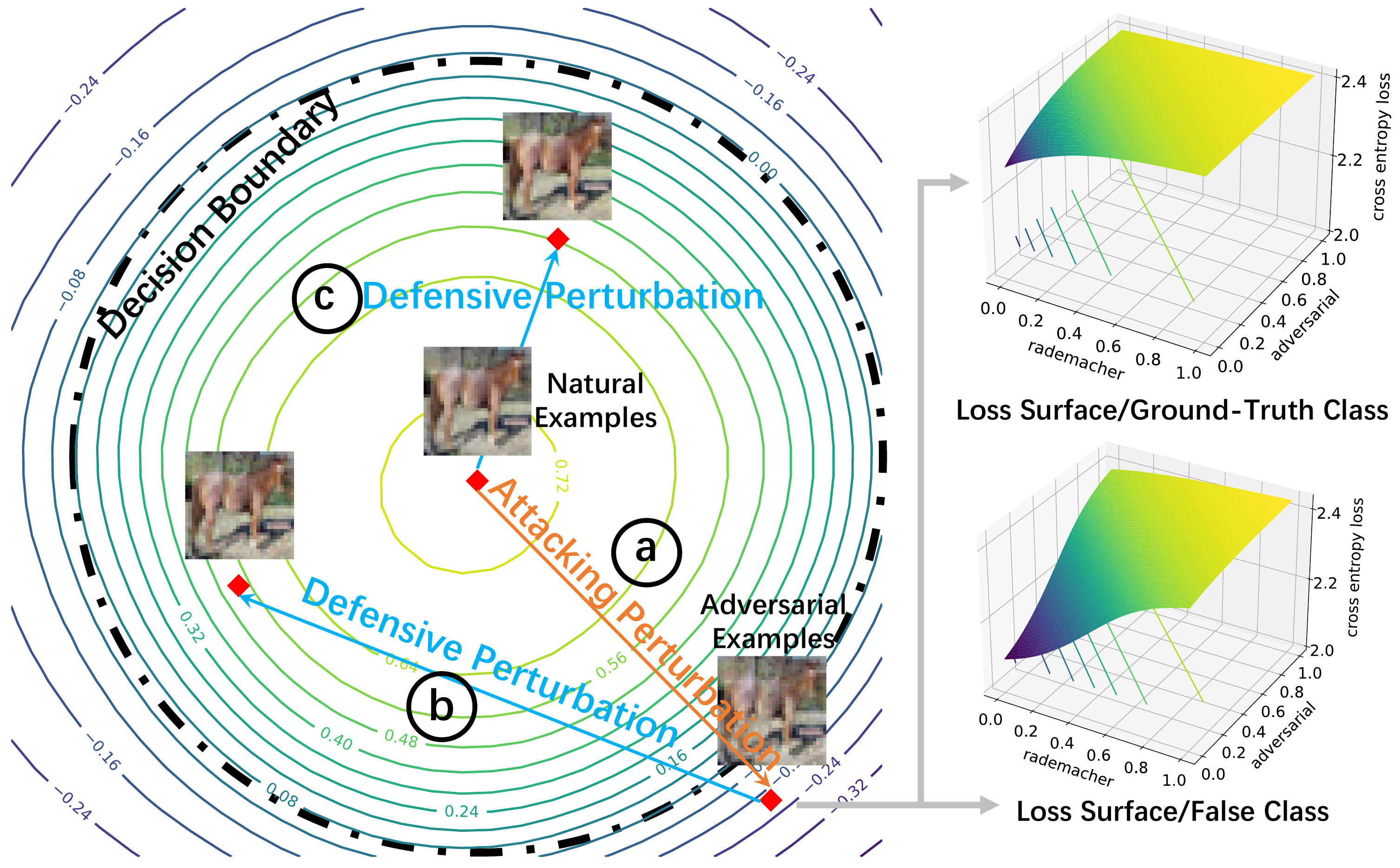}
	\caption{\textbf{Left:} The loss landscape~\cite{visualizing-landscape} around a natural example. An attacking perturbation intends to push the example out of the decision boundary (dot-dashed line), while a defensive perturbation intends to pull it back. \textbf{Right:} For the adversarial example on the left, we plot its two loss surfaces for the class of the false prediction and the ground-truth class.}
	\label{fig:main-idea}
	\vskip -0.1in
\end{wrapfigure}

Deep neural networks are first found to be vulnerable to adversarial attacks in \cite{Szegedy13}. That is, malicious attackers can deceive deep networks into a false category by adding a small perturbation onto the natural examples, as shown by Figure~\ref{fig:main-idea}(a). From then on, huge efforts have been made to counter this intriguing deficit~\cite{jsma,distillation,Xie18,sap}. After an intense arms race between attacks and defenses, the methodology of adversarial training~\cite{fgsm,pgd,trades} withstands the examination of time and becomes the state-of-the-art defense. Yet, even with adversarial training, the robustness against attacks is still far from satisfying, considering the huge gap between natural accuracy and robust accuracy. On the attacking side, various attacks targeting at different scenarios are kept being proposed, \eg, white-box attacks~\cite{cw,deepfool}, black-box attacks~\cite{square}, and hard-label attacks~\cite{rays}.

In this work, after carefully examining 25 leading robust models against 9 most powerful and representative attacks, we find that the adversarial attacks themselves are not robust, either. Specifically, on these adversarially trained models, if we add small random perturbations onto the deceptive adversarial examples before feeding them to networks, the false predictions of adversarial examples can be reverted to the correct predictions of natural examples. Empirical evaluations suggest that such a vulnerability against perturbations varies among different kinds of attacks. For instance, on the adversarially-trained model of TRADES~\cite{trades}, the successful rate of DeepFool~\cite{deepfool} will be decreased from $41.85\%$ to $34.52\%$. For the state-of-the-art black-box attack of Square~\cite{square}, the degradation will be $40.55\%$ to $32.21\%$. In contrast, attacks like PGD~\cite{pgd} and C\&W~\cite{cw} exhibit stronger resistance to random perturbations, and the degradation is generally 
less than $4.0\%$.

The above finding offers a new insight for enhancing adversarially trained models: can defenders find a defensive perturbation that is more effective than a random noise (Figure~\ref{fig:main-idea}(b)) and thus breaks attacks like PGD and C\&W? In this work, we propose to accomplish this by leveraging a common understanding of adversarial training. That is, the ground-truth predictions are stabilized during training and thus may have a smoother loss surface and smaller local Lipschitzness~\cite{Ross18,parseval,jarn,gu2019saliency,width}. This protection on the ground-truth class broadly exists in the vicinity of natural examples as well as the adversarial ones. In Figure~\ref{fig:main-idea}~Right, though deep networks wrongly classify the adversarial example, the loss surface for the false prediction is much steeper than that of the ground-truth class. Thus, if we simultaneously attack all the classes under the same attacking steps, unprotected false predictions with larger Lipschitzness will be pushed further away, while the ground-truth class may well survive. On natural examples (Figure~\ref{fig:main-idea}(c)), since the above pattern also holds, attacking all the classes barely affects their accuracy (See Section~\ref{sec:exp}) .

The above manner of attacking all the classes resembles the concept of hedge fund in financial risk control, where people invest in contradictory assets to contain the risk of an isolated investment. Thus, we term our method Hedge Defense. Hedge Defense creates perturbations tailored to each example and is more effective than the indiscriminate random noise. On CIFAR10~\cite{cifar}, it improves the state-of-the-art model~\cite{uncover} from $66.16\%$ to $72.66\%$ against AutoAttack~\cite{autoattack}, an ensemble of 4 most powerful attacks from different tracks, including an improvement from $71.76\%$ to $83.30\%$ against the state-of-the-art black-box attack of Square~\cite{square}. On ImageNet~\cite{imagenet}, it improves the top-1 robust accuracy of FastAT~\cite{fastat} from $33.18\%$ to $38.54\%$ against the 100-step PGD~\cite{pgd} attack. We first introduce Hedge Defense in Section~\ref{sec:method} and then present empirical evaluations in Section~\ref{sec:exp}. In Section~\ref{sec:theory}, we interpret Hedge Defense on a linear model for a better understanding and discuss its feasibility and possible limitations in Section~\ref{sec:discu}. Finally, we conclude our work in Section~\ref{sec:conclu}.

\section{Related Work}
\label{sec:related}

\paragraph{Adversarial Attacks}

Since the breakthrough finding in \cite{Szegedy13}, various adversarial attacks have been proposed, which can be approximately divided into two categories, white-box and black-box attacks. White-box attacks~\cite{gu2021effective,cw,multitarget} are allowed to get access to the model. Thus, they essentially maximize the loss value by repeatedly perturbing the input. For black-box attacks, where attackers cannot get access to gradient information, attacks are usually conducted via transferring adversarial examples~\cite{Tramer17} or extensive queries~\cite{zoo,square,rays}. To the best of our knowledge, all of these attacks are assumed to acquire the label in advance. In Appendix~\ref{app:label-free}, we show that the reverse process of Hedge Defense can attack deep models without knowing the label as a by-product of our work.

\paragraph{Adversarial Training}

At the beginning of developing robust models, plenty of defensive methods with different insights are proposed~\cite{distillation,jsma}, some of which even do not require training~\cite{Guo18}. Yet, \cite{obfuscated} found that most of these defenses are the falsehood originating from the obfuscated gradient problem, which can be circumvented by a tailored adaptation. Meanwhile, a series of defenses, dubbed adversarial training~\cite{ensemble,awp,mart}, withstand the above testing and have become the most solid defense so far. Adversarial training conducts attacks during training and directly trains deep models with the generated adversarial examples, allowing it to fuse the upcoming attacks in the future. Yet, empirical results suggest that adversarial training alone is not enough to learn highly robust models.

\paragraph{Alternative Defenses}

Alternative defenses are proposed~\cite{llr,jarn}. In particular, \cite{Guo18} and \cite{Xie18} adopt image pre-processing to counter attacks, which has been proven to be non-robust against white-box attacks~\cite{obfuscated}. \cite{PalV20,PinotERCA20} study attacks and defenses under Game Theory and theoretically analyze that defenders can improve the deterministic manner of predictions by adding a uniform noise. \cite{PalV20} only testifies the uniform perturbation on MNIST~\cite{mnist} against the PGD attack, where the improvement is restricted. This paper is the first work to extensively study various kinds of attacks and point out their non-robust deficiency. In addition, \cite{rmc} proposes to dynamically modify networks during inference time. Their method requires collective testing and has a different optimization criterion from ours. We compare Hedge Defense with these closely-related methods in more details in Appendix~\ref{app:more-related}.
\section{Hedge Defense}
\label{sec:method}

\paragraph{Preliminaries.}
For a classification problem with $C$ classes, we denote the input as $\xb$ and the ground-truth label as $y$. Then, a robust model~\cite{Tramer20} is expected to output $y$ for any input within $\mathbb{B}(\xb,\epsilon_\text{a}) = \{ \xb' \ | \ \| \xb'  - \xb  \|_p \leq \epsilon_\text{a}\}$, which defines the neighborhood area around $\xb$ with the attacking radius $\epsilon_\text{a}$ under $\ell_p$ norm. And the most fundamental form of adversarial attacks can be described as finding an adversarial example $\xb_\text{adv}$ that maximizes the loss value for the label $y$~\cite{Szegedy13}:
\begin{small}
\begin{align}
\xb_\text{adv} = 
\argmax\limits_{\xb' \in \mathbb{B}(\xb,\epsilon_\text{a})}
\mathcal{L}(f(\xb'),y)
= 
\argmax\limits_{\xb' \in \mathbb{B}(\xb,\epsilon_\text{a})}
-\log f_y(\xb').
\label{eq:attack} 
\end{align}
\end{small}
\!\!$\mathcal{L}(\cdot,\cdot)$ represents the adopted loss function, whose specific form may vary for different attacks. We choose the most popular choice of the cross-entropy loss for demonstration. $f(\cdot)$ denotes the deep model whose output is the predicted distribution, and $f_y(\cdot)$ is its scalar probability for class $y$.

\subsection{Defense via Attacking All Classes}

\begin{wrapfigure}{r}{0.48\textwidth}
\vspace{-1.2cm}
\begin{minipage}{0.47\columnwidth}
\begin{algorithm}[H]
\small
\caption{\ Hedge Defense}
\label{alg:hd}
 \begin{algorithmic}[1]
    \STATE {{\bfseries Input}: the coming input $\xb'$,  the number of iterations $K$, the step size $\eta$,  the deep network $f(\cdot)$,  and the defensive radius $\epsilon_\text{d}$.}
    \STATE{\emph{// $ \cU(\mathbf{-1},\mathbf{1}) $ generates a uniform noise}}
    \STATE {{\bfseries Initialization}: $\xb''_0 \gets \xb' + \epsilon_\text{d} \ \cU(\mathbf{-1},\mathbf{1}) $}.
    \FOR{$k = 1 \dots K$}
    \STATE{\emph{// Sum the losses for all the classes,}}
    \STATE{\emph{// then update $\xb''_k$ with gradient ascending.}}
    \STATE{$\xb''_k \gets \xb''_{k-1} + $ \\ $ \quad \eta\cdot\sign(\nabla_{\xb''_{k-1}}\sum_{c=1}^{C} \cL(f(\xb''_{k-1}),c))$;}
    \STATE{\emph{// $\Pi$ is the projection operator.}}
    \STATE{$\xb''_k \gets \Pi_{\mathbb{B}(\xb',\epsilon_\text{d})} (\xb''_k)$;}
    \ENDFOR
    \STATE {{\bfseries Output}: the safer prediction $f(\xb''_K)$.}
 \end{algorithmic}
\end{algorithm}
\vspace{-0.3cm}
\end{minipage}
\end{wrapfigure}

For the defensive purpose, we consider adding an extra perturbation on all the coming inputs before making the prediction so that we may dodge the dangerous adversarial examples. In practice, defenders do not know whether the coming input is adversarial or not. Thus, for a general sense of robustness, a defense should barely affect an arbitrary $\xb'$ within $\mathbb{B}(\xb,\epsilon_\text{a})$ meanwhile improve the resistance against $\xb_\text{adv}$. The defensive perturbation should also be constrained with a defensive radius $\epsilon_\text{d}$, so that the image is not substantially changed. Thus, for any $\xb'$ within $\mathbb{B}(\xb,\epsilon_\text{a})$, we search in examples of $\xb''$ within $\mathbb{B}(\xb',\epsilon_\text{d})$ to find a safer example $\xb_\text{hed}$. Specifically, we achieve this by maximizing the summation of the cross-entropy losses of all the classes:
\begin{small}
\begin{align}\label{eq:defense}
\xb_\text{hed} = 
\argmax\limits_{\xb'' \in \mathbb{B}(\xb',\epsilon_\text{d})}
\sum_{c=1}^{C}
\cL(f(\xb''),c),
\quad
\forall\xb' \in \mathbb{B}(\xb,\epsilon_\text{a}).
\end{align}
\end{small}
\!\!When simultaneously attacking all the classes, the computed gradients for the input example will be dominated by those that attack the false classes since they have larger local Lipschitzness and can thus be more easily attacked. Note that, the soft-max cross-entropies of different classes contradict each other: a smaller logit of the false class means a larger loss of its own but a smaller loss value of the correct class. Thus, the loss value of the ground-truth class actually becomes smaller. On natural examples, because the above analysis can also hold, Hedge Defense barely affects their accuracy. We describe the detail of Hedge Defense in Algorithm~\ref{alg:hd}. In Section~\ref{sec:theory}, we will further explain its mechanism on the more precise linear model for a binary classification problem.

Taking a deeper look at Eq.\eqref{eq:defense}, the scheme of attacking all the classes can be reformulated as: 
\begin{small}
\begin{align}\label{eq:defense-as-kl}
\sum_{c=1}^{C}
\mathcal{L}(f(\xb''),c)
=
C \cdot
\sum_{c=1}^{C}
\Big[
\frac{1}{C}
\log 
\frac{1/C}{f_c(\xb'')}
\Big]
-
\log \frac{1}{C}
=
C \cdot
\mathrm{KL}(\ub,f(\xb'')) + \mathrm{H}(\ub),
\end{align}
\end{small}
\!\!where $\mathrm{H}(\ub)$ denotes the entropy of the uniform distribution $\ub$. Thus, Hedge Defense is equivalent to maximizing the KL-Divergence to the uniform distribution. This offers another perspective to understand Hedge Defense. As adversarial attacks usually result in low-confidence predictions~\cite{admi,rce}, Hedge Defense may force the input to stay away from those uncertain predictions and return to the sharper predictions of natural examples. In addition, minimizing the KL-Divergence to the uniform distribution is a popular choice of training out-of-distribution examples~\cite{oe}, examples that belong to none of the considered categories. Thus, maximizing this metric as Eq.~\eqref{eq:defense-as-kl} can also be interpreted as seeking a better in-distribution prediction.

\section{Attacking State-of-the-Art Attacks}
\label{sec:exp}

\subsection{Evaluations with Robust Models on Benchmark Datasets}

This section extensively examines state-of-the-art robust models and attacks, investing in how the adversarial examples react to perturbations. We mainly consider the more typical $\ell_{\infty}$ norm attacks here and investigate the weaker $\ell_2$ norm attacks in Appendix~\ref{app:l2}. We present evaluations with the metric of robust accuracy, which is opposite to the successful rate of attacks.

\subsubsection{Empirical Results on CIFAR10 and CIFAR100}

\label{sec:cifar}

\begin{table*}[t!]
	\caption{Robust accuracy~($\%$) before perturbation (-), after random noise (Random), and after Hedge Defense (Hedge) on CIFAR10. Eight robust models are showed here. Evaluations are repeated three times with the mean values presented. Extra models and variances are provided in Appendix~\ref{app:cifar-extra}.}
	\label{table:cifar10}
	\centering
	\vskip 0.05in
	\resizebox{\textwidth}{!}{
	\begin{tabular}{ cccccccccccccc } 
		\toprule
		Model & Method 
		& \tabincell{c}{Nat-\\Acc.} 
		& PGD  & C\&W 
		& \tabincell{c}{Deep\\Fool}
		& \tabincell{c}{APGD\\CE}  
		& \tabincell{c}{APGD\\T}
		& FAB & Square & RayS 
		& \tabincell{c}{Auto\\Attack}
		& \tabincell{c}{Worst\\Case} \\
		\midrule
		\midrule
		\multirow{3}{*}{\tabincell{c}{WA+\\SiLU}~\cite{uncover}} 
		& -       
		& 91.10  
		& 69.16  & 67.48  & 71.23  & 68.24  & 66.17  
		& 66.70  & 71.76  & 72.03  & 66.16  & 66.15 \\
        & Random 
        & 90.82  
        & 69.34  & 71.35  & 75.65  & 69.46  & 67.45  
        & 78.82  & 77.58  & 79.23  & 67.76  & 66.24  \\
        & \textbf{Hedge}   
        & 90.62  
        & \textbf{71.43}  & \textbf{78.42}  & \textbf{79.64}  & \textbf{73.86} & \textbf{73.00}    
        & \textbf{82.28}  & \textbf{83.30}  & \textbf{82.03}  & \textbf{72.66} & \textbf{68.61} \\
		\midrule
		\multirow{3}{*}{AWP~\cite{awp}} 
		& -       
		& 88.25  
		& 64.14  & 61.44  & 64.81  & 63.56  & 60.49  
		& 60.97  & 66.15  & 66.93  & 60.50  & 60.46   \\
        & Random 
        & 87.91  
        & 64.18  & 65.01  & 70.41  & 64.53  & 61.54  
        & 73.37  & 72.47  & 74.89  & 62.00  & 60.42   \\
        & \textbf{Hedge}    
        & 86.98 
        & \textbf{66.16} & \textbf{71.94} & \textbf{73.83} & \textbf{68.18} & \textbf{66.47} 
        & \textbf{76.37} & \textbf{76.69} & \textbf{76.80} & \textbf{66.68} & \textbf{62.72} \\ 
		\midrule
		\multirow{3}{*}{RST~\cite{rst}} 
		& -       
		& 89.69 
		& 63.17  & 61.74  & 66.69  & 62.01  & 60.14  
		& 60.66  & 66.91  & 67.61  & 60.13  & 60.08   \\
        & Random 
        & 89.50  
        & 63.47  & 65.90  & 71.16  & 63.42  & 61.49          
        & 75.48  & 74.51  & 77.02  & 61.89  & 60.09   \\
        & \textbf{Hedge}    
        & 88.64 
        & \textbf{66.38} & \textbf{73.89} & \textbf{75.25} & \textbf{68.37} & \textbf{68.09} 
        & \textbf{78.48} & \textbf{79.24} & \textbf{78.98} & \textbf{67.46} & \textbf{63.10}  \\ 
		\midrule
		\multirow{3}{*}{\tabincell{c}{Pre-\\Training}~\cite{pretrain}} 
		& -       
		& 87.11 
		& 58.19  & 57.09  & 59.77  & 57.52  & 55.32  
		& 55.68  & 62.38  & 63.32  & 55.30  & 55.28    \\
        & Random 
        & 86.93 
        & 58.52  & 61.79  & 70.4   & 58.59  & 56.45          
        & 71.99  & 71.16  & 74.56  & 56.79  & 55.32   \\
        & \textbf{Hedge}    
        & 87.43 
        & \textbf{62.01} & \textbf{73.12} & \textbf{75.50} & \textbf{65.63} & \textbf{65.38} 
        & \textbf{76.62} & \textbf{78.32} & \textbf{78.65} & \textbf{64.45} & \textbf{59.17} \\ 
		\midrule
		\multirow{3}{*}{MART~\cite{mart}} 
		& -       
		& 87.50  
		& 63.49  & 59.62  & 63.14  & 61.83  & 56.73          
		& 57.39  & 64.88  & 65.66  & 56.74  & 56.70          \\
        & Random 
        & 87.08 
        & 63.59  & 63.82  & 72.26  & 63.02  & 58.10           
        & 73.04  & 72.11  & 74.71  & 58.75  & 56.99          \\
        & \textbf{Hedge}   
        & 86.18 
        & \textbf{64.56} & \textbf{71.71} & \textbf{73.86} & \textbf{67.00} & \textbf{63.30}  
        & \textbf{74.35} & \textbf{76.09} & \textbf{75.60} & \textbf{63.42} & \textbf{59.35} \\ 
		\midrule
		\multirow{3}{*}{HYDRA~\cite{hydra}} 
		& -       
		& 88.98 
		& 60.95  & 62.19  & 64.66  & 59.86  & 57.67  
		& 58.41  & 65.02  & 65.61  & 57.67  & 57.63          \\
        & Random 
        & 88.80  
        & 61.42  & 66.08  & 69.50  & 61.25  & 59.03          
        & 74.60  & 73.26  & 76.20  & 59.55  & 57.85          \\
        & \textbf{Hedge}    
        & 87.82 
        & \textbf{64.19} & \textbf{73.05} & \textbf{73.43} & \textbf{66.15} & \textbf{65.56} 
        & \textbf{76.94} & \textbf{77.58} & \textbf{77.36} & \textbf{65.07} & \textbf{61.13} \\ 
		\midrule
		\multirow{3}{*}{TRADES~\cite{trades}} 
		& -       
		& 84.92 
		& 55.83  & 54.47  & 58.15  & 55.08  & 53.09  
		& 53.56  & 59.45  & 59.69  & 53.09  & 53.06          \\
        & Random 
        & 84.69 
        & 56.03  & 58.65  & 65.48  & 56.06  & 54.22          
        & 69.33  & 67.79  & 71.29  & 54.51  & 53.02          \\
        & \textbf{Hedge}    
        & 83.99 
        & \textbf{58.99} & \textbf{69.98} & \textbf{71.30}  & \textbf{62.24} & \textbf{62.22} 
        & \textbf{74.31} & \textbf{74.71} & \textbf{73.94}  & \textbf{61.28} & \textbf{56.21} \\  
		\midrule
		\multirow{3}{*}{AT~\cite{pgd}} 
		& -       
		& 83.23 
		& 46.97  & 57.51  & 55.84  & 47.97  & 50.99  
		& 72.68  & 77.71  & 60.93  & 47.29  & 38.81  \\
        & Random 
        & 82.99 
        & 47.06  & 57.77  & 56.86  & 47.92  & 51.23          
        & 72.73  & 77.66  & 82.94  & 47.27  & 40.55   \\
        & \textbf{Hedge}    
        & 81.03 
        & \textbf{48.31} & \textbf{63.22} & \textbf{63.52} & \textbf{52.38} & \textbf{57.67} 
        & \textbf{74.89} & \textbf{77.93} & \textbf{80.47} & \textbf{51.82} & \textbf{42.87} \\
		\midrule
		\midrule
		\multirow{3}{*}{Standard} 
		& -       
		& 94.78 
		& 00.00  & 00.00  & 00.83  & 00.00  & 00.00  
		& 00.04  & 00.39  & 00.04  & 00.00  & 00.00  \\
        & Random 
        & 91.32 
        & 00.00          & \textbf{15.97} & 74.57   & 01.16   & 01.16           
        & \textbf{84.37} & \textbf{56.41} & 79.15   & 01.20   & 00.00               \\
        & \textbf{Hedge}    
        & 91.50
        & 00.00  & 15.66  & \textbf{74.71} & \textbf{01.23}  & \textbf{01.21}  
        & 84.29  & 56.37  & \textbf{79.43} & \textbf{01.25}  & 00.00    \\
		\bottomrule
	\end{tabular}}
	\vskip -0.1in
\end{table*}

On CIFAR10, we select the top-20 officially-released robust models on RobustBench~\cite{robustbench} to demonstrate the effectiveness of our method. We show seven of them here and put the rest in Appendix. We also test models trained with the basic adversarial training (AT) and standard training for comparison. Details of these models are shown in Appendix~\ref{app:rb-models}. On the attacking side, we select nine most representative and powerful attacks from different tracks, including six white-box attacks (PGD~\cite{pgd}, CW~\cite{cw}, DeepFool~\cite{deepfool}, APGD-CE~\cite{autoattack}, APGD-T~\cite{autoattack}, FAB~\cite{fab}), two black-box attacks (Square~\cite{square}, RayS~\cite{rays}), and one ensemble attack (AutoAttack~\cite{autoattack}). All attacks are constrained by $\epsilon_\text{a}=8/255$. We introduce different attacks in detail in Appendix~\ref{app:attacks}. 

For each model in Table~\ref{table:cifar10}, we compare the robust accuracy of three scenarios: predictions before perturbation, predictions after random perturbation, and predictions after hedge perturbation. By default, we discuss $\epsilon_\text{d} = \epsilon_\text{a}$ as in \cite{PalV20} and empirically test different values in Section~\ref{sec:radius}. Hedge Defense is implemented with $20$ iterations of optimization with a step size of $4/255$. This is the same configuration as the PGD attack. We first evaluate each attack independently. Then, we compute the worst-case robust accuracy of all attacks. Namely, for a certain testing example, as long as one of the nine attacks succeeds, we consider the defense as a failure. From Table~\ref{table:cifar10}, we can observe that most attacks are vulnerable to both random perturbation and hedge perturbation. And the capability to resist perturbations differs hugely among them. Here we briefly analyze possible causes of their non-robustness. A complete case study can be found in Appendix~\ref{app:case-study}.

\textbf{Square and RayS:} Square and RayS are black-box attacks based on extensively querying deep networks. They tend to stop queries once they find a successful adversarial example so that the computation budget can be saved. Yet, such a design makes attacks stop at the adversarial examples that are closer to the unharmful example, which is the example one step earlier before the successful attacking query. This explains why they are so vulnerable to defensive perturbation. The black-box constraint of not knowing the gradient information of the model may also prevent them from finding more powerful adversarial examples that are far away from the decision boundary.

\begin{small}
\vskip -0.15in
\begin{align}
\textbf{FAB:} \
\xb_\text{fab} = \argmin\limits_{\xb'} { \| \xb' -\xb \|_p} \quad
\st
\argmax\limits_{c}f_c(\xb') \neq y. 
\end{align}
\end{small}
\!\!\textbf{FAB and DeepFool:} FAB can be considered as a direct improvement of DeepFool. Unlike other white-box attacks that search adversarial examples within $\mathbb{B}(\xb,\epsilon_\text{a})$, FAB tries to find a minimal perturbation for attacks, as shown above. However, this aim of finding smaller attacking perturbations also encourages them to find adversarial examples that are too close to the decision boundary and thus become very vulnerable to defensive perturbation. Thus, even on the undefended standard model where they generate smaller perturbations than on robust models, they can still be hugely nullified by random perturbation, and the robust accuracy increases from less than $1.0\%$ to $74.57\%$ and $84.37\%$.

\begin{small}
\vskip -0.15in
\begin{align}
\textbf{C\&W:} \
\xb_\text{cw} = 
\argmax\limits_{\xb' \in \mathbb{B}(\xb,\epsilon_\text{a})}
(-z_y(\xb')+\max\limits_{i\neq y}z_{i}(\xb')).
\end{align}
\end{small}
\!\!\textbf{PGD and C\&W:}
Attacks like PGD and C\&W show better resistance against perturbations. However, with Hedge Defense, the robust accuracy can still by improved by $3.0\% $ to $11.0\%$. Interestingly, several commonly believed more powerful attacks turn out to be weaker when facing perturbations. For instance, C\&W enhances attacks by increasing the largest false logit $z_{i}(\xb') (i\neq y)$ and decreasing the true logit $z_y(\xb')$. Without Hedge Defense, it is more powerful ($54.47\%$ robust accuracy of TRADES) than the PGD attack ($55.83\%$). With Hedge Defense, it becomes less effective ($69.98\%$ vs $58.99\%$). This indicates that its efficacy comes at a price of less robustness against defensive perturbation. And the effect of directly increasing the false logit can be weakened if the false class is attacked. Similar analyses can also be applied to the attacks of APGD-CE and APGD-T.

\begin{table*}[t!]
	\caption{Robust accuracy~($\%$) on CIFAR100. All settings align with CIFAR10 in Table~\ref{table:cifar10}.}
	\label{table:cifar100}
	\centering
	\vskip 0.05in
	\resizebox{\textwidth}{!}{
	\begin{tabular}{ cccccccccccccc } 
		\toprule
		Model & Method 
		& \tabincell{c}{Nat-\\Acc.} 
		& PGD  & C\&W 
		& \tabincell{c}{Deep\\Fool}
		& \tabincell{c}{APGD\\CE}  
		& \tabincell{c}{APGD\\T}
		& FAB & Square & RayS 
		& \tabincell{c}{Auto\\Attack}
		& \tabincell{c}{Worst\\Case} \\
		\midrule
		\midrule
		\multirow{3}{*}{\tabincell{c}{WA+\\SiLU}~\cite{uncover}} 
		& -       & 69.15    & 40.84          & 39.46          & 40.74             & 40.32            & 37.33           & 37.65          & 42.97           & 43.15          & 37.29                & 37.26          \\ 
		& Random & 69.02             & 40.97          & 43.61          & 52.22             & 41.35            & 38.75           & 54.59          & 51.08           & 55.09          & 38.79                & 37.33          \\
		& \textbf{Hedge}    & 68.67             & \textbf{43.25} & \textbf{55.94} & \textbf{57.82}    & \textbf{46.75}   & \textbf{46.47}  & \textbf{59.69} & \textbf{59.81}  & \textbf{59.75} & \textbf{45.00}          & \textbf{39.78} \\ 
		\midrule
		\multirow{3}{*}{AWP~\cite{awp}} 
		& -       & 60.38             & 34.13          & 31.41          & 31.33             & 33.37            & 29.18           & 29.47          & 34.57           & 34.79          & 29.16                & 29.15          \\
		& Random & 60.45    & 34.21          & 35.24          & 44.42             & 34.53            & 30.56           & 45.59          & 42.94           & 47.82          & 30.75                & 29.25          \\
		& \textbf{Hedge}    & 59.37             & \textbf{36.69} & \textbf{46.01} & \textbf{48.13}    & \textbf{40.18}   & \textbf{38.10}   & \textbf{49.49} & \textbf{49.82}  & \textbf{50.34} & \textbf{37.75}       & \textbf{32.29} \\
		\midrule
		\multirow{3}{*}{TRADES~\cite{trades}} 
		& -       & 57.34    & 25.19          & 32.83          & 31.97             & 34.51            & 37.24           & 52.74          & \textbf{54.06}           & 30.01          & 33.94                & 19.61          \\
		& Random & 55.60              & 25.30           & 33.34          & 33.97             & 34.11            & 36.73           & 51.41          & 52.36           & 55.59          & 33.78                & 22.00             \\
		& \textbf{Hedge}    & 56.04             & \textbf{29.75} & \textbf{44.09} & \textbf{45.33}    & \textbf{39.25}   & \textbf{42.04}  & \textbf{53.27} & 54.01  & \textbf{55.79} & \textbf{38.72}       & \textbf{26.59} \\ 
		\midrule
		\multirow{3}{*}{AT~\cite{pgd}} 
		& -       & 58.61    & 25.34          & 35.01          & 31.93             & 32.05            & 35.56           & 52.62          & 55.11           & 32.31          & 31.70                 & 19.72          \\
		& Random & 58.49             & 25.43          & 35.53          & 35.54             & 32.09            & 35.57           & 52.58          & \textbf{55.00}              & \textbf{58.72}          & 31.74                & 21.61          \\ 
		& \textbf{Hedge}    & 57.66             & \textbf{28.23} & \textbf{43.73} & \textbf{45.02}    & \textbf{35.79}   & \textbf{40.08}  & \textbf{53.24} & 54.99  & 57.22 & \textbf{35.54}       & \textbf{24.80}  \\
		\bottomrule
	\end{tabular}}
	\vskip -0.1in
\end{table*}

For natural accuracy on robust models, Hedge Defense is slightly lower than the directly predicting approach in most cases. This aligns with the common understanding that robustness may be at odds with natural accuracy~\cite{tes19,trades,obfuscated}. Compared with the improvements on robust accuracy, such a degradation of natural accuracy is acceptable. One exception is the robust model of Pre-Training Method~\cite{pretrain}, where Hedge Defense actually achieves better natural accuracy. On CIFAR100, we evaluate four robust models in Table~\ref{table:cifar100}. Hedge Defense again achieves huge improvements. In particular, it improves the state-of-the-art model against AutoAttack from $37.29\%$ to $45.00\%$.

\subsubsection{Evaluations of Fast Adversarial Training on ImageNet}

We further test Hedge Defense on the challenging ImageNet dataset. Because adversarial training is a computationally consuming algorithm, it is important to reduce its complexity to apply to very large datasets like ImageNet. In particular, Fast Adversarial Training~\cite{fastat} is proposed for this purpose and achieves promising results. Thus, we choose this algorithm and evaluate it on different architectures against the PGD attack of different numbers of iterations ($10/50/100$). Each trial is restarted 10 times until the attack succeeds. We adopt the officially released ResNet50 model of \cite{fastat}. Also, we train ResNet101 with the official code to compare models with different capacities. $\epsilon_\text{a}$ is set to $2/255$ or $4/255$ as in \cite{fastat}. In Table~\ref{table:imagenet}, Hedge Defense improves robust accuracy in all cases. As the model gets deeper, Hedge Defense achieves more improvement.

\begin{table*}[t!]
	\caption{Top-1 robust accuracy ($\%$) for Fast Adversarial Training on ImageNet.}
	\label{table:imagenet}
	\centering
	\vskip 0.05in
	\resizebox{0.94\textwidth}{!}{
	\begin{tabular}{ ccccccccc } 
		\toprule
		& &
		& \multicolumn{3}{c}{$\epsilon_\text{a}=2/255$}
		& \multicolumn{3}{c}{$\epsilon_\text{a}=4/255$}\\
		 \cmidrule(r){4-6}
		 \cmidrule(r){7-9}
		Method & Model & Hedge
		& PGD-10 & PGD-50 & PGD-100
		& PGD-10 & PGD-50 & PGD-100 \\
		\midrule
		\multirow{4}{*}{Fast AT~\cite{fastat}} 
		& \multirow{2}{*}{ResNet-50}
		& -
		& 43.44 & 43.40 & 43.38
		& 30.81 & 30.17 & 30.13  \\ 
		& 
		& $\checkmark$
		& \textbf{45.38} & \textbf{45.44}& \textbf{45.43}
		& \textbf{34.42} & \textbf{34.63} & \textbf{34.71} \\ 
		\cmidrule(r){2-9}
		& \multirow{2}{*}{ResNet-101} 
		& -
		& 44.69 & 44.62 & 44.60
		& 34.02 & 33.26 & 33.18 \\ 
		& 
		& $\checkmark$
		& \textbf{47.00} & \textbf{47.07} & \textbf{47.08}
		& \textbf{38.36} & \textbf{38.50} & \textbf{38.54} \\ 
		\bottomrule
	\end{tabular}}
	\vskip -0.1in
\end{table*}


\begin{figure*}[t!]
		\centering
\resizebox{0.98\textwidth}{!}{
		\subfigure[The Trade-off of Defensive Perturbation]{
			\label{fig:trade-off}
			\includegraphics[width=0.43\linewidth]{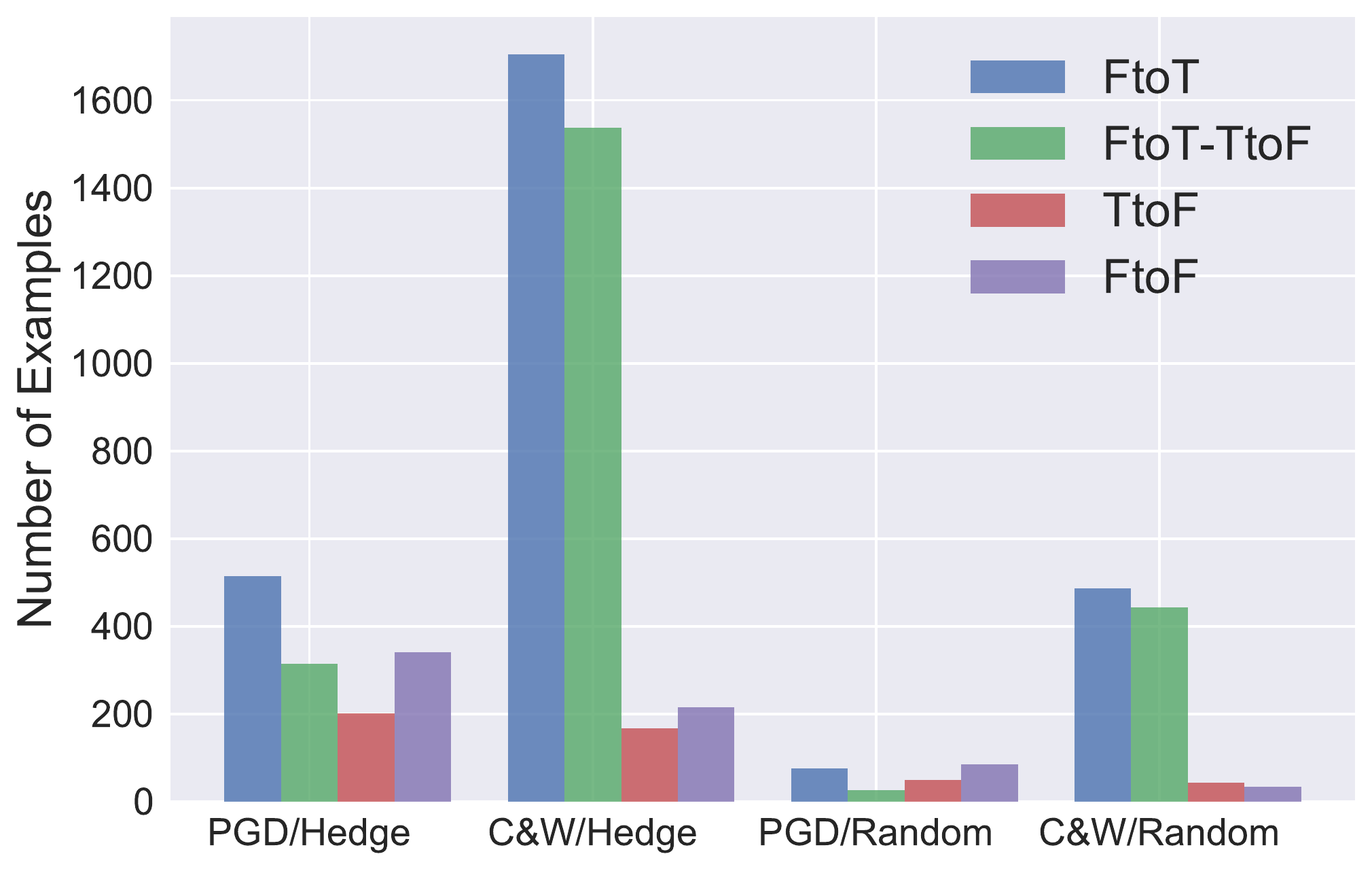}}
		\subfigure[Lipschitzness Difference]{
			\label{fig:lips-diff}
			\includegraphics[width=0.30\linewidth]{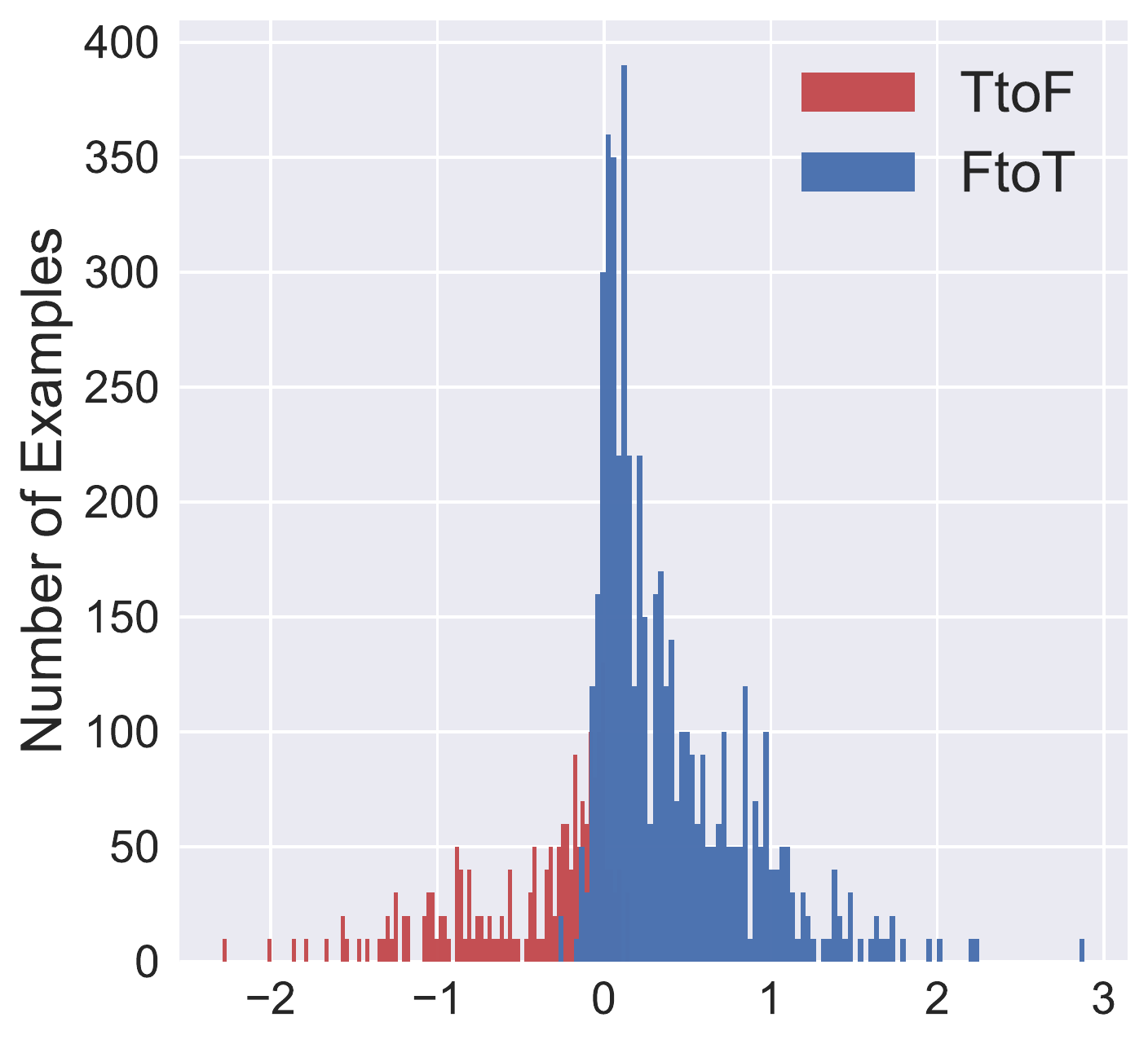}}
		\subfigure[Visualization of Examples]{
			\label{fig:visual}
			\includegraphics[width=0.45\linewidth]{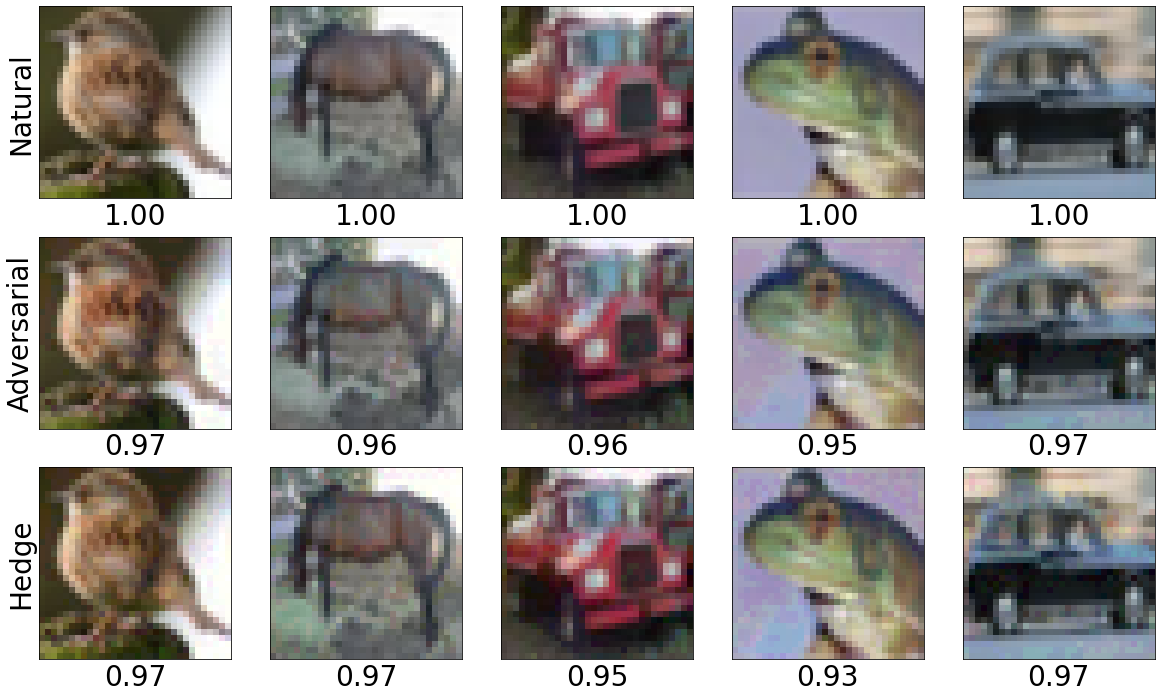}}
}
		\caption{Analytical Experiments: (a) The number of examples for FtoT, TtoF, FtoT-TtoF, and FtoF (Section~\ref{sec:trade-off}). (b) The distribution of Lipschitzness difference for FtoT and TtoF (Section~\ref{sec:lips-diff}). (c) The visualization for natural, adversarial, and hedge examples with their SSIM scores (Section~\ref{sec:visual}).}
		\label{fig:analy}
		\vskip -0.1in
\end{figure*}

\subsection{Analytical Experiments for Defensive Perturbation}
In this section, we investigate multiple features of defensive perturbation. We adopt the official model of TRADES~\cite{trades} on CIFAR10 for demonstration in this part.

\subsubsection{The Trade-off of Defensive Perturbation}
\label{sec:trade-off}

We have observed that Hedge Defense achieves overall improvements on robustness. Now we investigate how it influences each testing example. Specifically, Hedge Defense may either convert an originally true prediction to a false one (\textbf{TtoF}) or convert a false prediction to a true class (\textbf{FtoT}). Then, the overall improvement is the number of FtoT subtracts the number of TtoF. (\textbf{FtoT-TtoF}). In addition, we count the number of examples that are originally falsely predicted before defensive perturbations and are converted to another false class after defensive perturbations (\textbf{FtoF}). This metric indicates a partial failure of the targeted attack, which is expected to deceive deep networks into a specific false class rather than an arbitrary one. We show these statistics of Hedge Defense against different attacks in Figure~\ref{fig:trade-off}. It is shown that Hedge Defense improves robustness because it helps much more examples of FtoT, compared with the relatively fewer examples of TtoF. We also provide the statistics of random perturbation for comparison.

\subsubsection{Local Lipshitzness and Hedge Defense}
\label{sec:lips-diff}

Here we empirically verify the statement that Hedge Defense takes effect because the ground-truth class has smaller Lipshitzness. On FtoT examples where Hedge Defense brings a benefit, we compute the approximated Lipshitzness between the false class and the ground-truth class. As shown in Figure~\ref{fig:lips-diff} (blue), this difference is generally positive and verifies that the ground-truth class does have smaller Lipshitzness. We further evaluate the same metric on the few examples of TtoF and observe a generally negative difference (red). This again verifies our theory and shows that Hedge Defense actually hurts robustness if the Lipshitzness of the ground-truth class is not smaller. 

\subsubsection{Visualization and Image Similarity}
\label{sec:visual}

To provide a better sense of how Hedge Defense perturbs the images, we visualize the three types of examples, \ie, natural examples $\xb$, adversarial examples $\xb_\text{adv}$, and hedge examples $\xb_\text{hed}$. To show the similarity between different examples, we also compute the SSIM~\cite{ssim} score of adversarial and hedge examples with respect to natural examples. As shown in Figure~\ref{fig:visual}, both adversarial and hedge examples are visually consistent with natural examples. Their SSIM scores are generally higher than $0.90$, indicating a high level of similarity (The score of $1.0$ means it is exactly the same image). More visualizations, including defensively perturbed natural examples, can be seen in Appendix~\ref{app:more-visual}.


\begin{figure*}[t!]
		\centering

		\subfigure[]{
			\label{fig:noe-step}
			\includegraphics[width=0.23\linewidth]{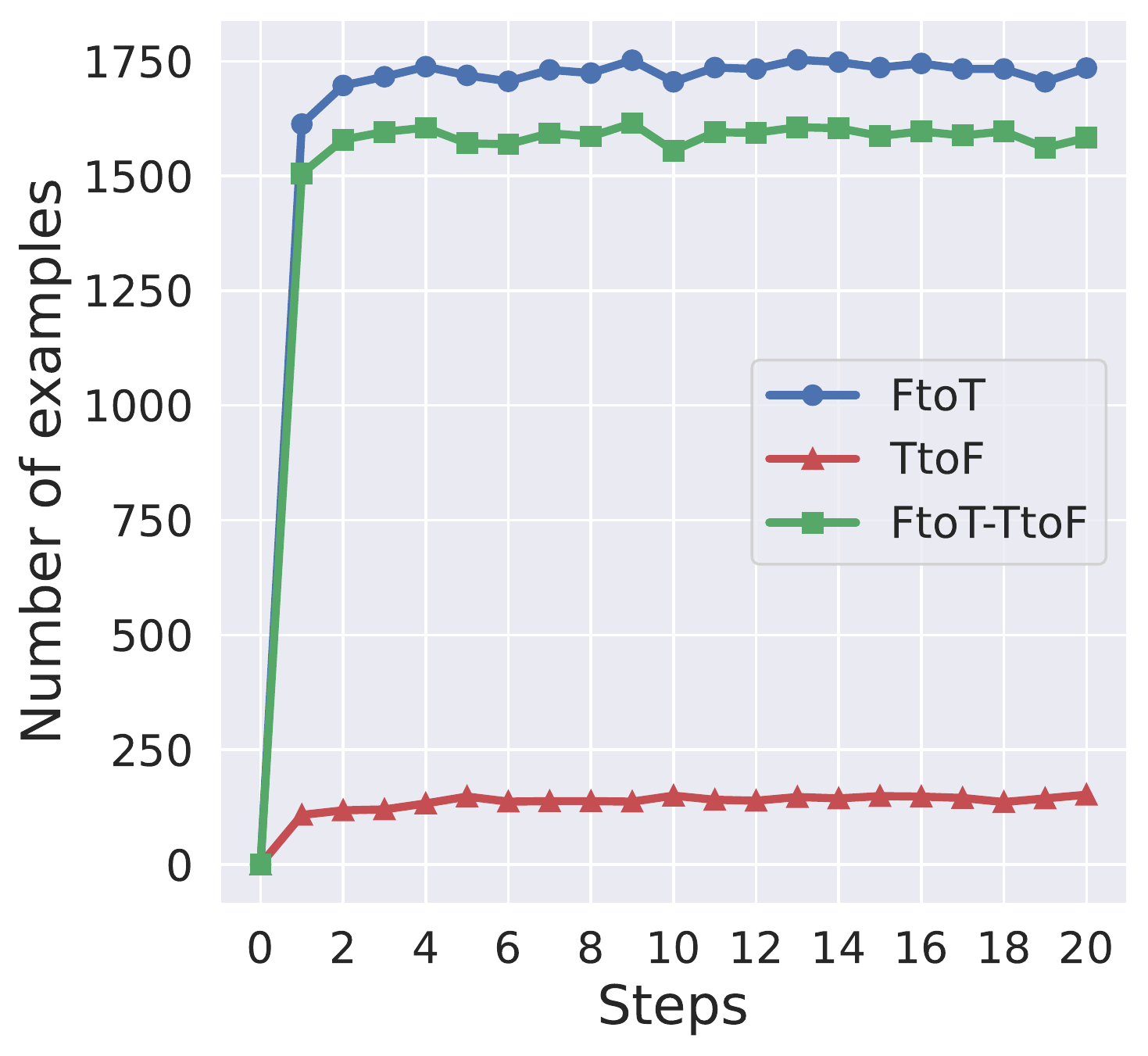}}
		\subfigure[]{
			\label{fig:noe-dr}
			\includegraphics[width=0.23\linewidth]{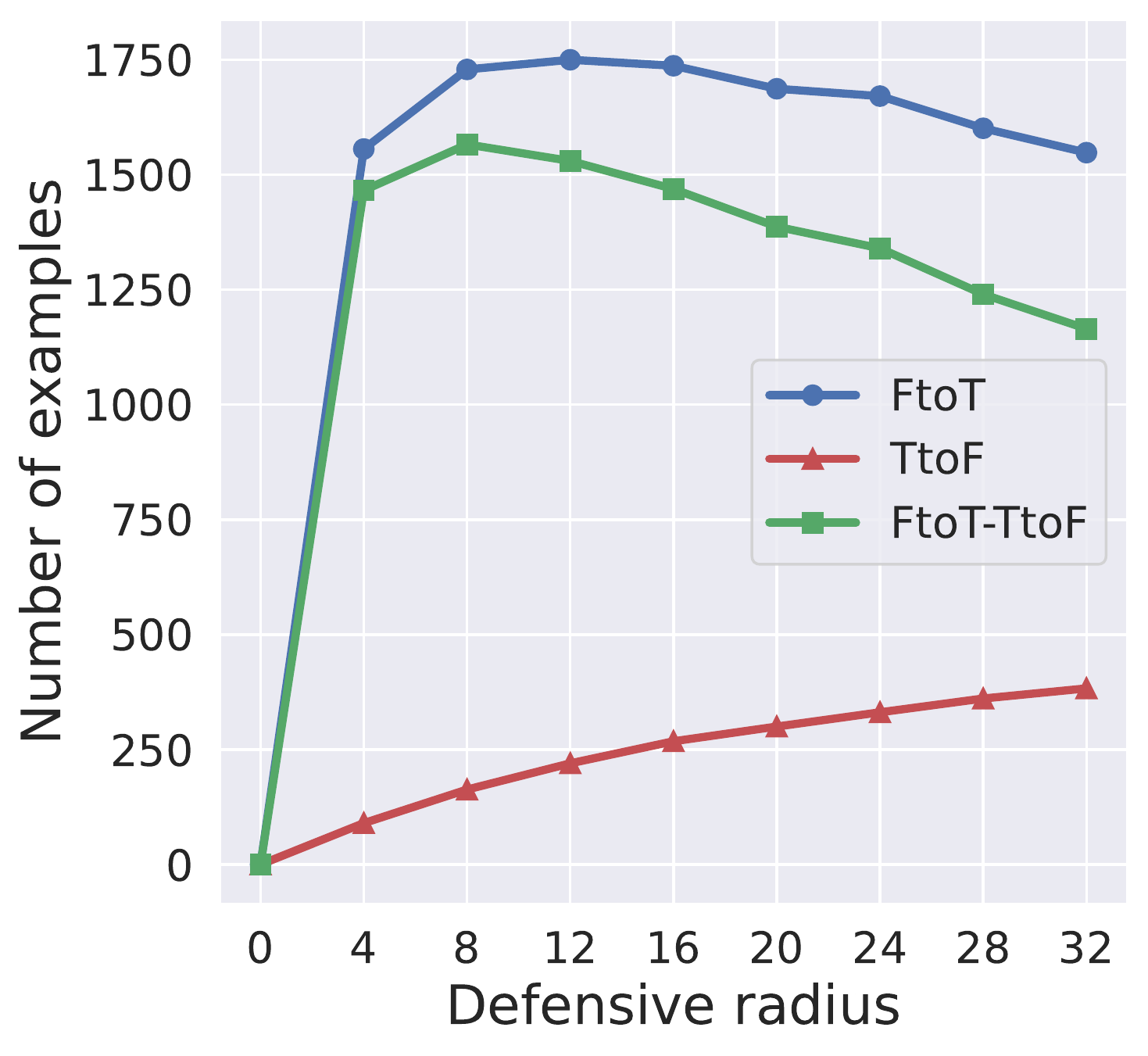}}
		\subfigure[]{
			\label{fig:noe-ar}
			\includegraphics[width=0.23\linewidth]{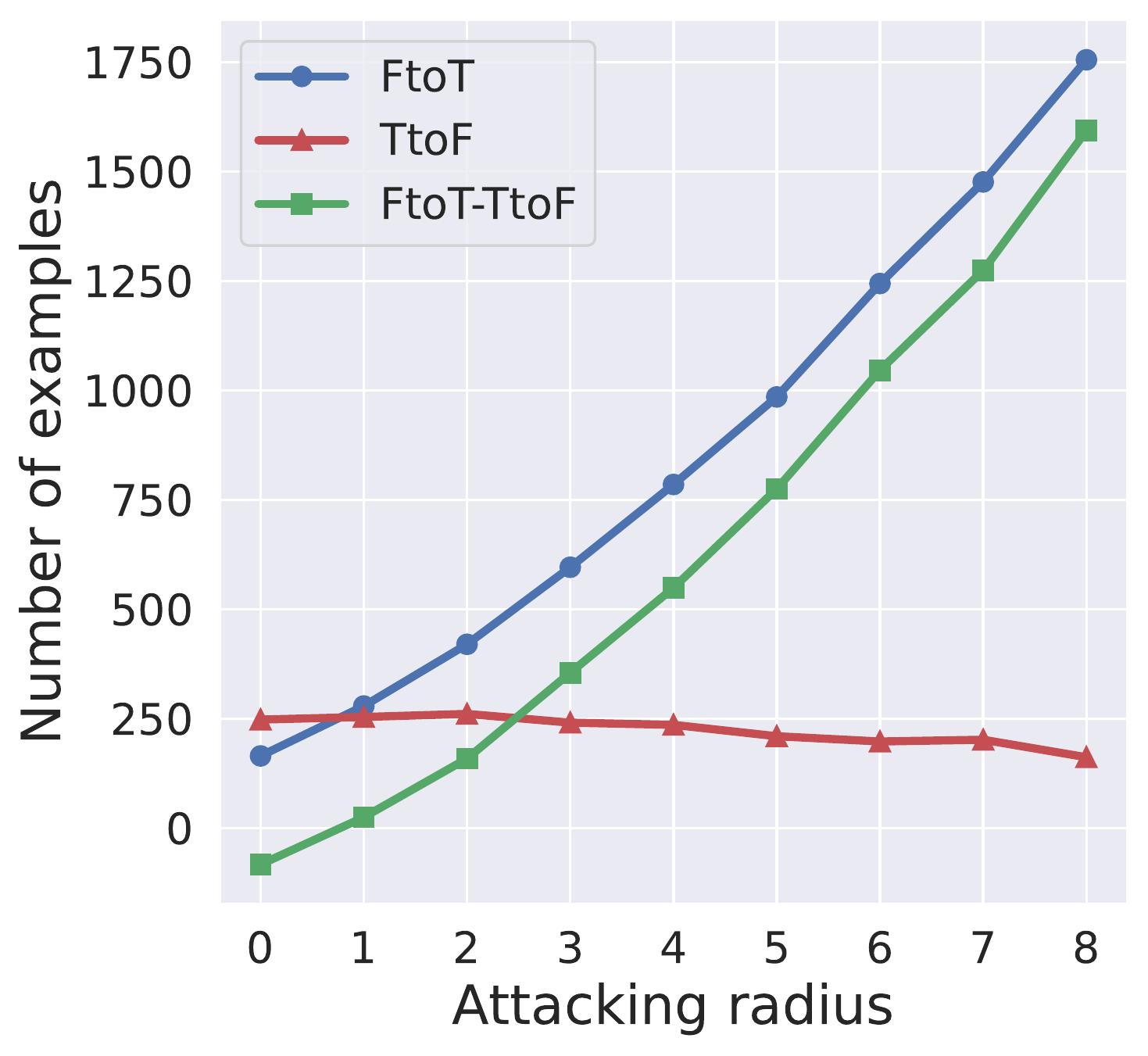}}
		\subfigure[]{
			\label{fig:ar-dr}
			\includegraphics[width=0.25\linewidth]{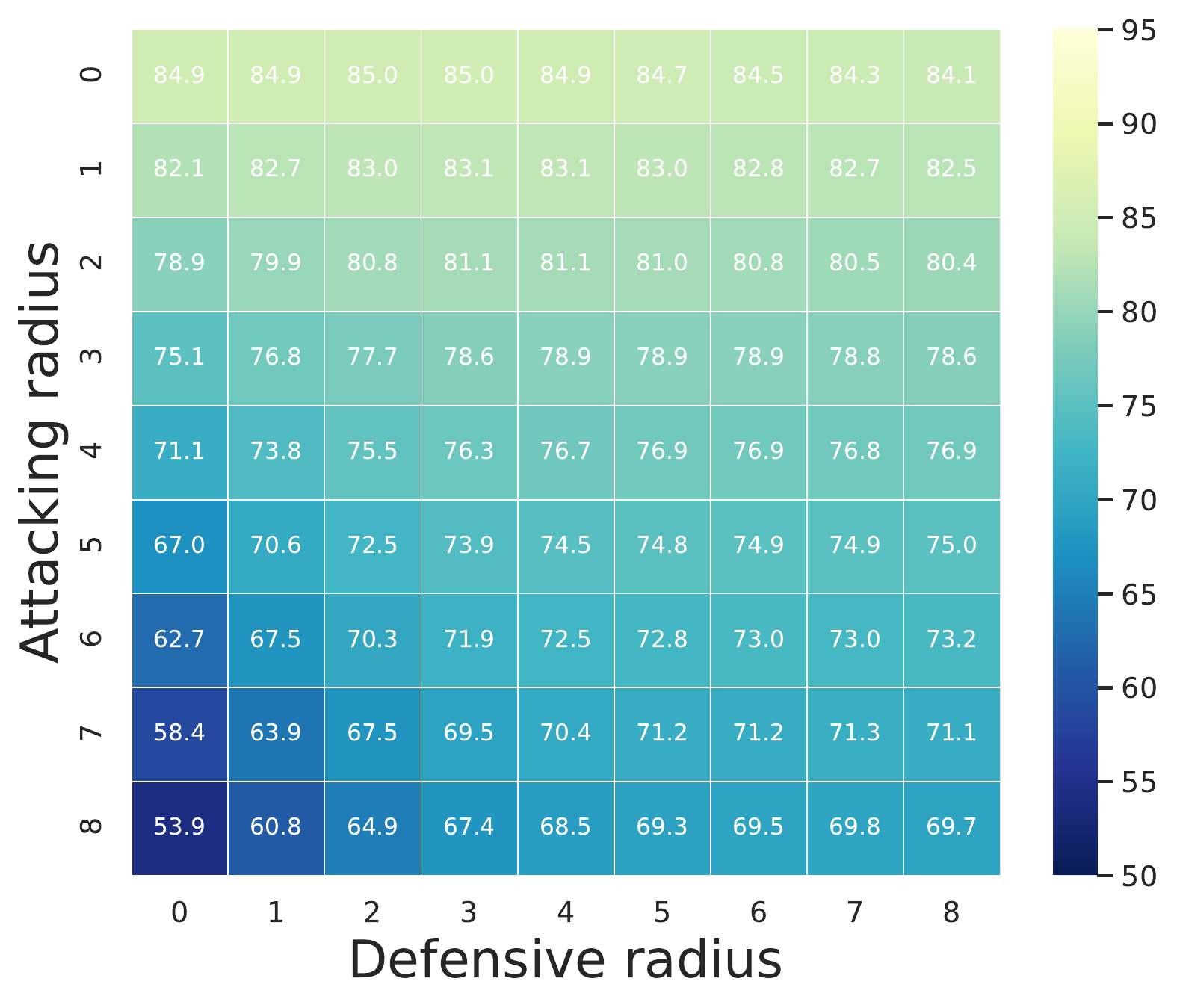}}
	
		\caption{(a) Number of examples for FtoT, TtoF, and FtoT-TtoF against Hedge Defense iteration steps (Section~\ref{sec:noe-step}). (b,c,d) Ablation study on the perturbation radii (/255) of $\epsilon_\text{a}$ and $\epsilon_\text{d}$ (Section~\ref{sec:radius}). }
		\label{fig:three}
		\vskip -0.1in
\end{figure*}

\subsection{Ablation Study for Defensive Perturbation}
In this section, we study the hyper-parameters of optimization steps and perturbation radius. The TRADES~\cite{trades} model on CIFAR10 is evaluated against C\&W~\cite{cw} attack.

\subsubsection{Ablation Study on the Complexity of Hedge Defense}
\label{sec:noe-step}

Although Hedge Defense requires extra computation, its complexity is the same as attacks since both of them utilize similar optimization techniques. Thus, defenders are only required to have the same computation budget against attackers. Moreover, the complexity of Hedge Defense mainly comes from the number of iterations ($K$ in Algorithm~\ref{alg:hd}). Thus, we plot the effect of Hedge Defense against its iteration times in Figure~\ref{fig:noe-step}. It is shown that Hedge Defense converges very fast. Thus, defenders can use our method with only a few steps of optimizations.

\subsubsection{Ablation Study on Attacking and Defensive Perturbation Radius}
\label{sec:radius}

Here we explore how the two perturbation radii of $\epsilon_\text{a}$ and $\epsilon_\text{d}$ influence each other in an extensive manner. We first set $\epsilon_\text{a}=8/255$, and investigate how different values of $\epsilon_\text{d}$ influence robust accuracy. As shown in Figure~\ref{fig:noe-dr}, the efficacy of Hedge Defense is not very sensitive to the value of $\epsilon_\text{d}$, considering obvious improvements for $\epsilon_\text{d}$ between $4/255 $ and $32/255$. 
We then set $\epsilon_\text{d}=8/255$ and investigate different values of $\epsilon_\text{a}$~\cite{Dong20}. As shown in Figure~\ref{fig:noe-ar}, the unwanted examples of TtoF stay relatively unchanged as $\epsilon_\text{a}$ increases, indicating TtoF is more related to $\epsilon_\text{d}$ instead of $\epsilon_\text{a}$. On the other hand, the other two metrics steadily increase with $\epsilon_\text{a}$. This suggests Hedge Defense performs better when attacks get stronger. $\epsilon_\text{a}>8/255$ is not tested since it may destroy semantic information of the example. Eventually, we plot a complete view for the values of the two radii in Figure~\ref{fig:ar-dr}. The color represents the value of robust accuracy under different $\epsilon_\text{a}$ and $\epsilon_\text{d}$.

\section{Interpretation of Hedge Defense}
\label{sec:theory}

We have shown that Hedge Defense can hugely improve the robustness of deep networks on real-world problems. To get a clearer understanding of its working mechanism, we give a detailed interpretation of Hedge Defense on a locally linear model in this section. We first construct a binary classification problem, where the ground-truth function is a robust model. Then, we add a small bias on the ground-truth function to create a naturally accurate but non-robust model. Finally, we show that Hedge Defense can convert this non-robust model to a robust counterpart, on the condition that the true class has a smaller magnitude of gradient, which is equivalent to Lipshitzness for linear models.

\begin{figure*}[t]
		\centering
		\subfigure[]{
			\label{fig:4a}
			\includegraphics[width=0.18\linewidth]{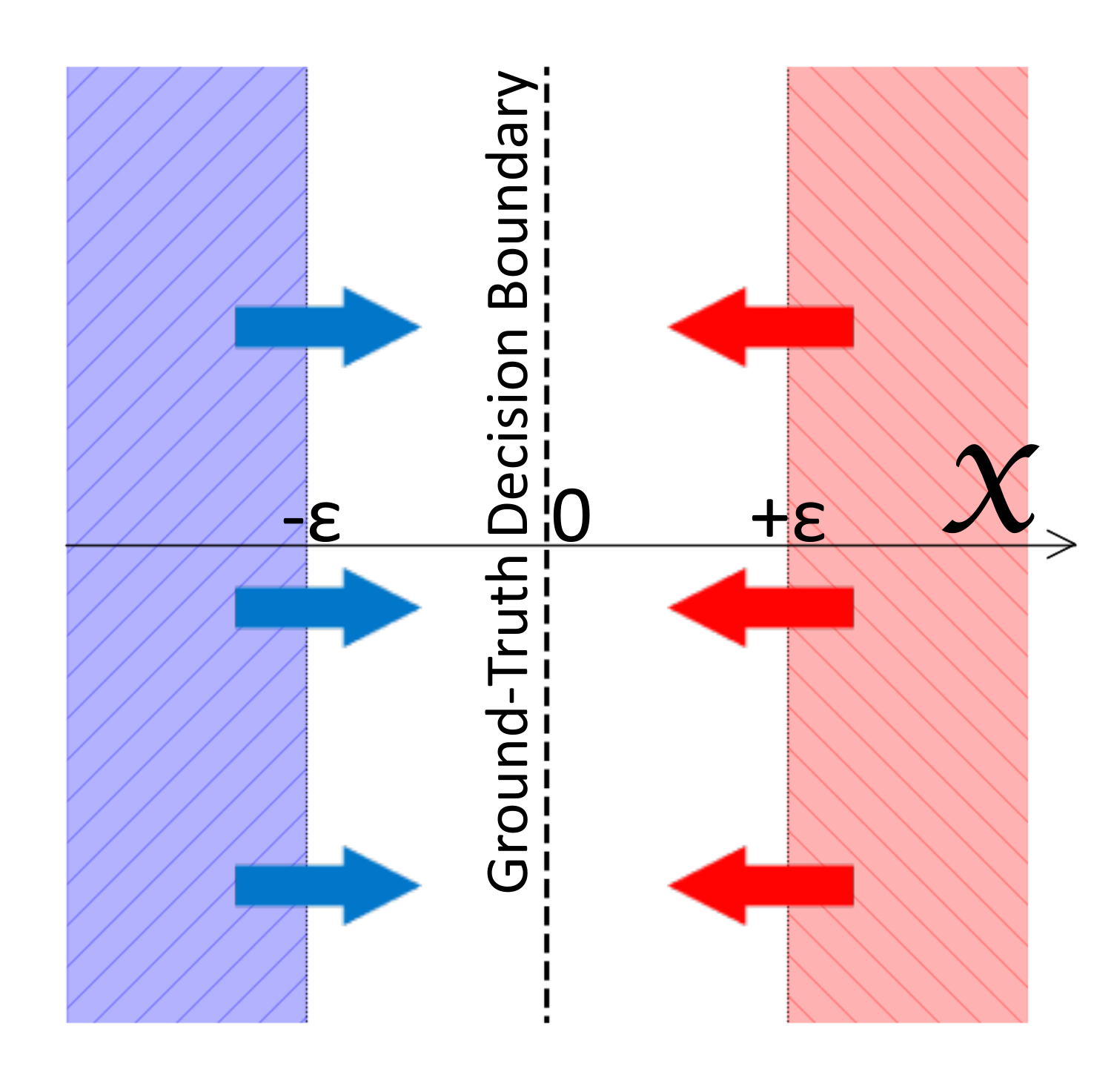}}
		\subfigure[]{
			\label{fig:4b}
			\includegraphics[width=0.18\linewidth]{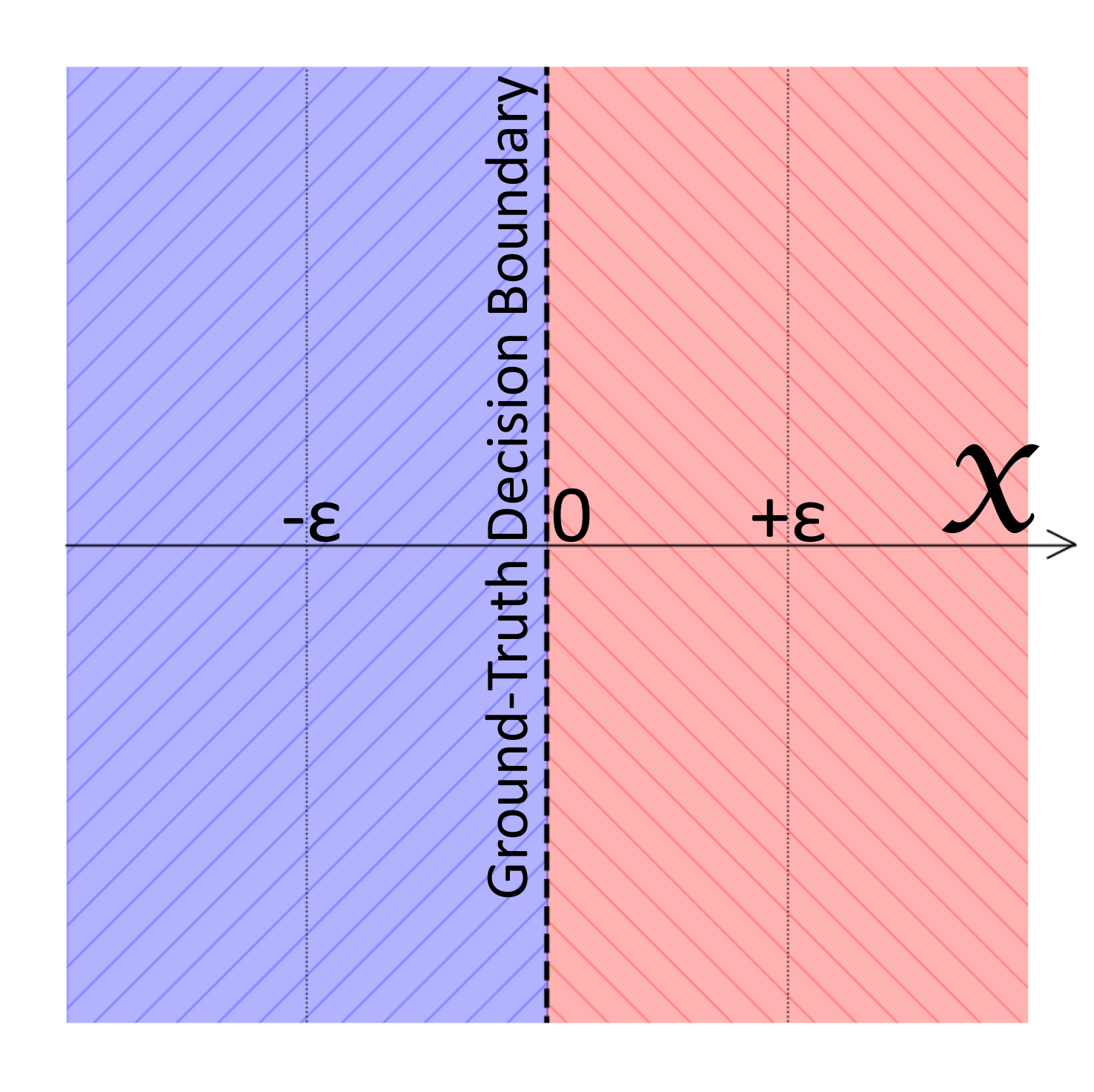}}
		\subfigure[]{
			\label{fig:4c}
			\includegraphics[width=0.18\linewidth]{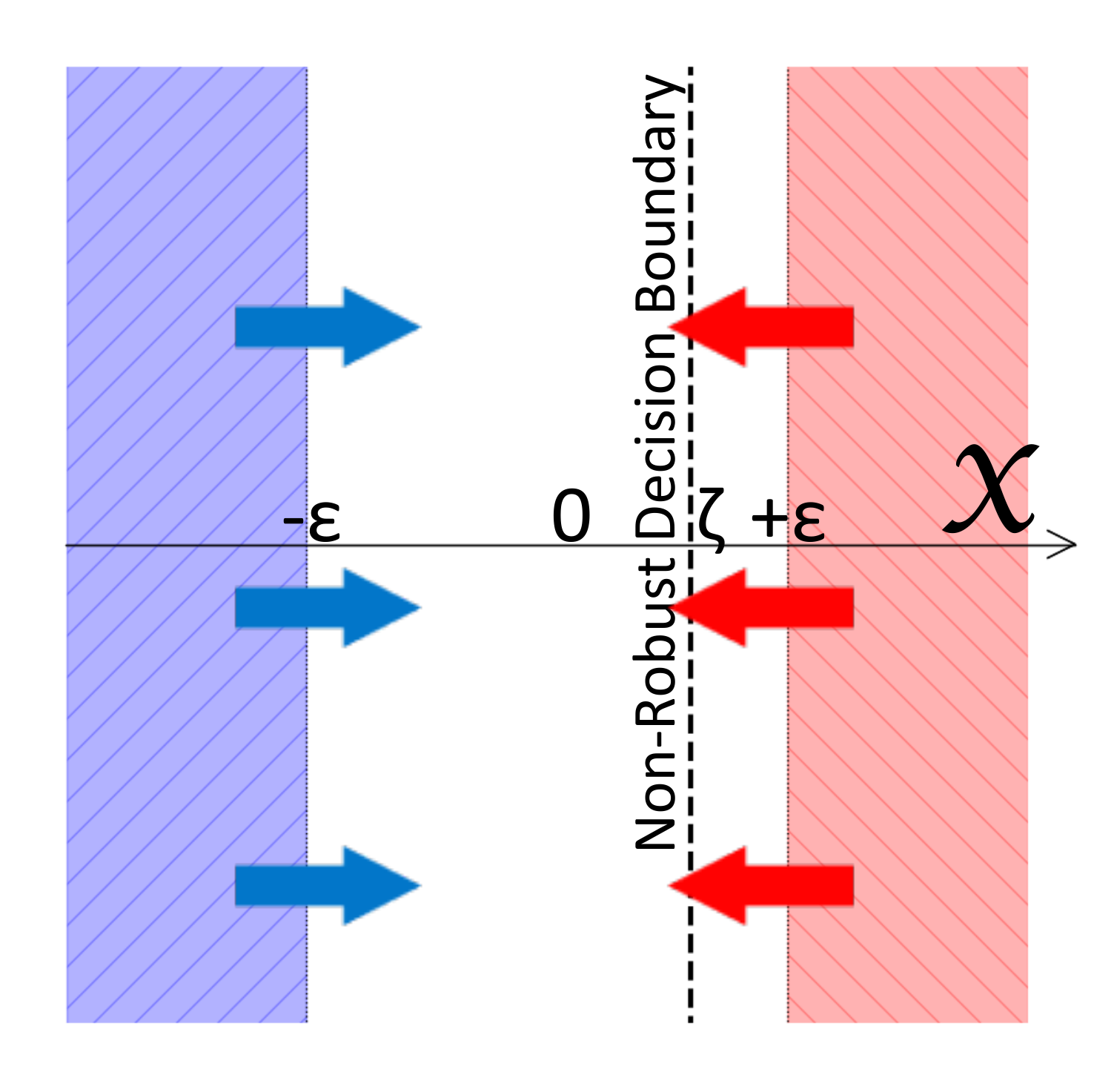}}
		\subfigure[]{
			\label{fig:4d}
			\includegraphics[width=0.18\linewidth]{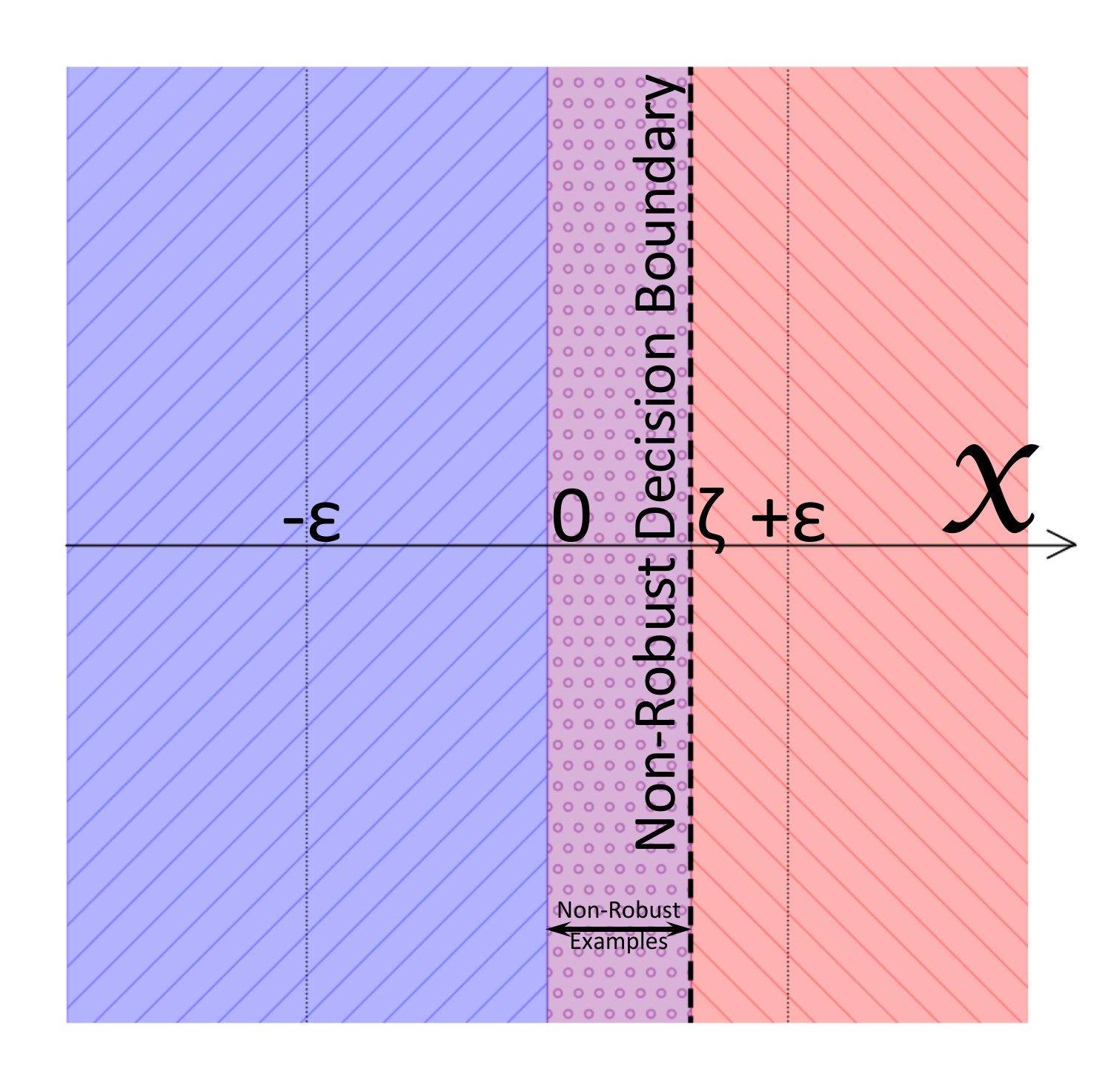}}
		\subfigure[]{
			\label{fig:4e}
			\includegraphics[width=0.18\linewidth]{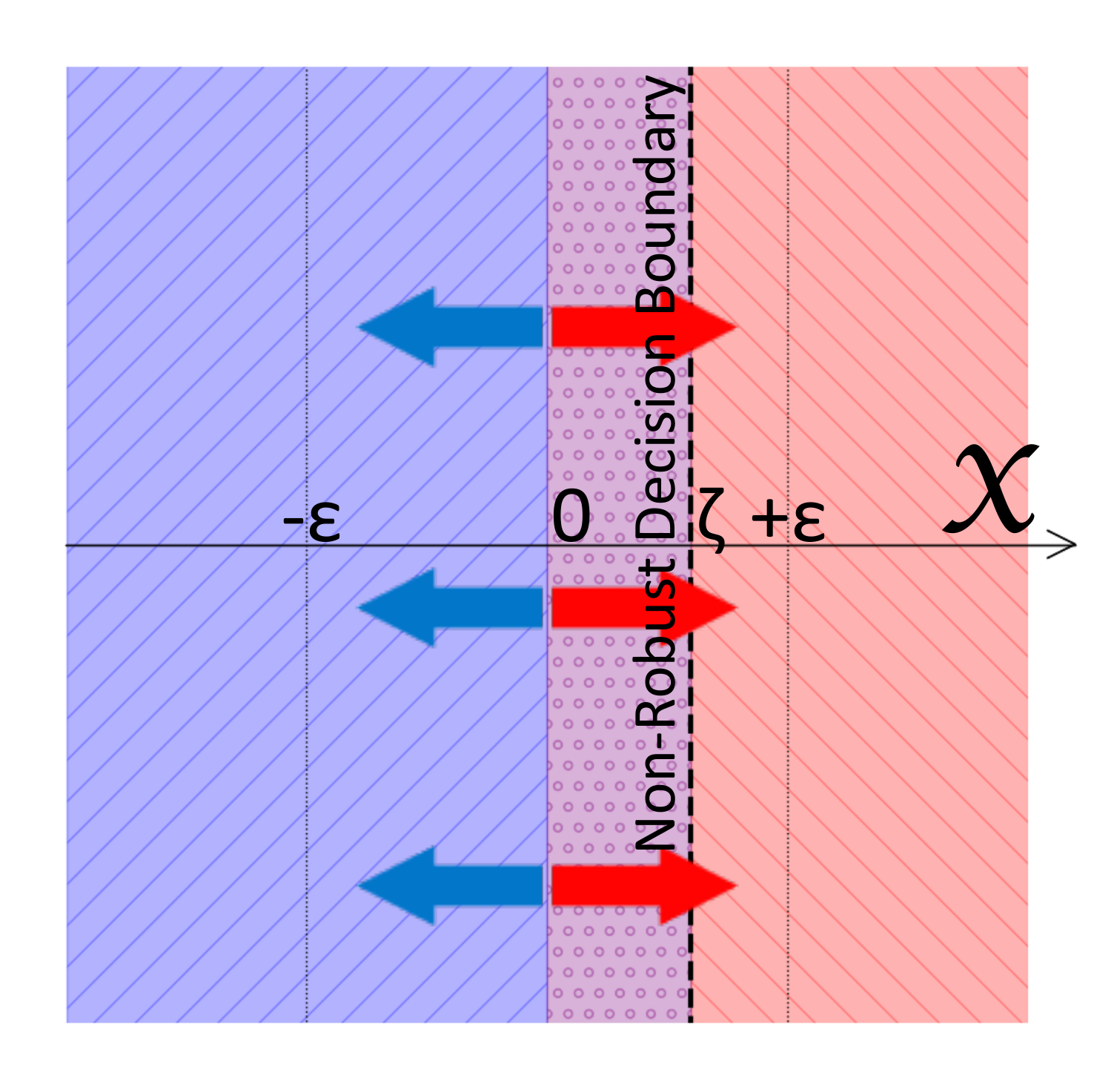}}
		\caption{We represent examples of the two classes with red ($+1$) and blue ($-1$). The arrows represent the direction of attacking or defensive perturbations. (a) Examples before attacks. (b) Example after attacks for the ground-truth model. (c) Attacking a non-robust model. (d) Examples after attacks for the non-robust model. The purple region represents the examples that are attacked successfully and wrongly classified. (e) The direction of defensive perturbation on adversarial examples.}
		\vskip -0.1in
\end{figure*}

\subsection{A Binary Classification Task}
We consider the classification for the two classes of $\{+1,-1\}$. Without loss of generality, we adopt the scalar $x$ to represent the input to demonstrate our idea, and all the analyses in this section can be easily extended to multi-dimensional cases. Each class has its own score function: $f_\text{+1}(x)$ for $y=+1$ and $f_\text{-1}(x)$ for $y=-1$ as Eq.~\eqref{linear-binary-classification}. The label $y$ for the input $x$ is determined by these two score functions: $y=\sign(f_\text{+1}(x)-f_\text{-1}(x))\in\{+1,-1\}$. Thus, $y=+1\ \text{if} \ x > 0$, and $y=-1\ \text{if} \ x < 0$.
\begin{equation}
\label{linear-binary-classification}
\small
f_\text{+1}(x) = 
\begin{cases}
k_t x \quad \text{if} \ x > 0  \\
k_f x \quad \text{if} \ x < 0
\end{cases}, \quad
f_\text{-1}(x) = 
\begin{cases}
-k_f x \quad \text{if} \ x > 0  \\
-k_t x \quad \text{if} \ x < 0
\end{cases}, \quad
\ k_f \gg k_t > 0.
\end{equation}
We denote the gradient of the true class as $k_t$, which is the gradient of $f_\text{+1}(x)$ if $y=+1$ and the gradient of $f_\text{-1}(x)$ if $y=-1$, and $k_f$ as the false counterpart. As stated in Section~\ref{sec:method}, Hedge Defense relies on the condition that the magnitude of $k_t$ is much smaller than $k_f$, which is $\ k_f \gg k_t > 0$. We set $f_\text{+1}(x)$ in a reverse proportion to $f_\text{-1}(x)$ so that the two classes exclude each other.

\subsection{Construct a Non-Robust Classifier under Adversarial Attacks}

Suppose the attacking perturbation is bounded by $\epsilon>0$. Then, the standard FGM attack~\cite{fgsm,PalV20} generates the attacking perturbation on the opposite direction of the gradient for the true class:
\begin{equation}
\footnotesize
a_\text{FGM}(x) 
=
\frac{-\epsilon\nabla f_\text{+1}(x)}
{\|\nabla f_\text{+1}(x)\|}
=
\frac{-\epsilon k_t}{\|k_t\|}
=-\epsilon,
\ \text{if} \ y=+1 ;\quad
a_\text{FGM}(x) 
=
\frac{-\epsilon\nabla f_\text{-1}(x)}
{\|\nabla f_\text{-1}(x)\|}
=
\frac{\epsilon k_t}{\|-k_t\|}
=\epsilon,
\ \text{if}\ y=-1.
\end{equation}
As shown by the arrows in Figure~\ref{fig:4a}, the attack drags natural examples closer to the decision boundary. Under this attack, a robust example~\cite{PalV20} $x$ should has the same prediction as examples within $(x-\epsilon,x+\epsilon)$. Apparently, only $x\in(-\infty,-\epsilon)\cup(+\epsilon,+\infty)$ satisfies this requirement, while $x\in (-\epsilon,\epsilon)$ is intrinsically non-robust and too close to the decision boundary~\cite{rst}. Thus, we do not consider $x$ from $(-\epsilon,\epsilon)$, as shown by the blank in Figure~\ref{fig:4a}. In Figure~\ref{fig:4b}, we show the input after the attack in Figure~\ref{fig:4a}. The examples are not dragged across the decision boundary of the ground-truth function $\sign(f_\text{+1}(x)-f_\text{-1}(x))$, making it a robust solution for the learning task.

Next, we consider the situation where we have learned a non-robust model so that Hedge Defense can provide an extra protection. To this end, suppose we learn an estimated function $\hat{f}_\text{+1}(x)$ with a small bias: $\hat{f}_\text{+1}(x)=f_\text{+1}(x)-\zeta(k_t+k_f)$, where $\zeta \in (0,\epsilon)$, and $\hat{f}_\text{-1}(x)$ holds still to $f_\text{-1}(x)$. This assumption is reasonable since learning a bias frequently happens in many learning algorithms. Then, the estimated decision boundary will shift from $x=0$ to $x=\zeta$. The estimated classifier will always output the correct label for $x\in(-\infty,-\epsilon)\cup(+\epsilon,+\infty)$. Thus, the learned classifier is a naturally accurate discriminator.
However, when examples are attacked, as shown by the arrows in Figure~\ref{fig:4c}, $x \in (\epsilon,\epsilon+\zeta)$ will be perturbed into the range of $(0,\zeta)$ and then be wrongly classified by the estimated decision boundary, as shown by the purple region in Figure~\ref{fig:4d}. Thus, the learned classifier is not a robust discriminator like the ground-truth. In summary, we now have a classifier that is naturally accurate but non-robust. This resembles real-world tasks like CIFAR10, where we can learn a deep network that can accurately predict natural examples but fail to resist adversarial ones.

\subsection{Apply Hedge Defense to the Non-Robust Classifier} 

Consider that defenders can also add a defensive perturbation bounded by $\epsilon>0$ after attacks. 
For a coming input, natural or adversarial, Hedge Defense generates a defensive perturbation as Eq.~\eqref{eq:defense}:
\begin{equation}
\footnotesize
d_\text{HD}(x)=\frac{-\epsilon(\nabla \hat{f}_\text{+1}(x) + \nabla \hat{f}_\text{-1}(x))}{\|\nabla \hat{f}_\text{+1}(x) + \nabla \hat{f}_\text{-1}(x)\|}
=
\begin{cases}
-\epsilon(k_t-k_f)/\| k_t-k_f \|    
\approx \epsilon k_f/\|-k_f\|
=\epsilon
\quad \text{if} \ x > 0  \\
-\epsilon(k_f-k_t)/\|k_f-k_t\|    
\approx -\epsilon k_f/\|k_f\|
=-\epsilon       
\quad \text{if} \ x < 0
\end{cases}.
\notag
\end{equation}
Notice that, while attackers are allowed to utilize the ground-truth classifier, defenders can merely utilize the estimated one. The direction of the defensive perturbation generated by $d_\text{HD}(\cdot)$ is shown in Figure~\ref{fig:4e}. The defensive perturbation is opposite to the attacking perturbation and will eventually push examples back to the positions before attacks occur. This allow the estimated classifier to correctly make a decision. Given the above illustration, we can tell that Hedge Defense highly relies on the condition that the true class has a smaller magnitude of gradient. If $k_t>k_f$, Hedge Defense actually worsens the prediction and generates perturbations on the same direction of adversarial perturbations. This explains the rare cases of TtoF in Section~\ref{sec:exp}, where deep models correctly classify adversarial examples but fail on hedge examples. Extra analyses are shown in Appendix~\ref{app:more-theory}.

\section{Discussion}
\label{sec:discu}

\textbf{From the Attacking Perspective:}  In this work, we have shown that adversarial attacks can also be vulnerable to perturbations. One may wonder, if attackers have known that a defensive perturbation will be added before prediction, can they design better attacks? For instance, based on our analyses in Section~\ref{sec:exp}, attacks like Square may keep querying to find more robust adversarial examples, and FAB may consider keeping a proper distance to the decision boundary. We have also tried several possible adaptions to enhance the white-box attacks. Our defense can survive from all these defense-aware attacks (see Appendix~\ref{app:attempt-attack}). Specifically, to resist defensive perturbation, we consider seeking a local space that is either full of misled predictions or against the condition that the ground-truth class has a smaller Lipschitzness. Both can be harder than simply finding a point-wise falsehood.

\textbf{From the Defensive Perspective:} Whenever a new defense comes out, the first thing that occurs to us is: Is this new method just an illusion of improper evaluations? Can it be naively broken by tailored attacks? Although we cannot prove the certified robustness~\cite{certified} of Hedge Defense, for now, we can still get some insights about its solidity from the following aspects: 1) In a thorough evaluation of robustness, defenses should never interfere with the attacking stage. Our evaluations obey this rule. Specifically, we did not add any defensive perturbation to the input when generating adversarial examples because it can hugely worsen the performance of attacks (see Appendix~\ref{app:attempt-attack}). 2) Hedge Defense does not fall into any category of the obfuscated gradient~\cite{obfuscated}. We present a detailed analysis in Appendix~\ref{app:not-obfuscated}. 3) \cite{PalV20} has theoretically justified that randomly perturbing the coming input before predicting can outperform the manner of directly predicting the coming input, and our analyses in Section~\ref{sec:theory} also partially confirm the reliability of Hedge Defense.

\section{Conclusion}
\label{sec:conclu}

The contribution of this work is two-fold. 1) We point out the novel finding that adversarial attacks can also be vulnerable to perturbations. 2) Enlightened by the above finding, we develop the algorithm of Hedge Defense that can more effectively break attacks and enhance adversarially trained models. Both empirical results and theoretical analyses support our method. Our work fights back attacks with the same methodology and sheds light on a new direction of defenses. Future work may consider using criteria other than attacking all the classes to find the way back to the correct predictions. Moreover, with Hedge Defense, defenders may not need to guarantee that the model can correctly classify all the examples within the local space. We only need to ensure that some conditions, \ie, the ground-truth class has smaller local Lipshitzness, exist to allow us to find those safer examples.


\bibliographystyle{ims}
\bibliography{references}
\newpage

\appendix

\section{Appendix Guideline}

In Appendix~\ref{app:more-exp}, we first provide extra detailed configurations about our empirical evaluations in Section~\ref{sec:exp}, including the detail of the robust models that we tested, a complete introduction of the nine adversarial attacks we selected, and a complete version of our case studies on different attacks in Section~\ref{sec:cifar}. Then, we present extra experimental results, including a complete version of Table~\ref{table:cifar10} with more robust models, the variance of the robust accuracy reported in Table~\ref{table:cifar10}, different visualizations for Section~\ref{sec:visual}, evaluations on $\ell_2$ norm attacks, reversing Hedge Defense as a label-free attack, the attempts that we made to counter Hedge Defense, and several extra experiments aiming at verifying the solidity of Hedge Defense.

In Section~\ref{sec:theory}, we have presented a linear model for a binary classification learning task. In Appendix~\ref{app:more-theory}, we will further explore this simplified model, including discussing the legitimacy of several assumptions we have made, how Hedge Defense reacts to natural examples, and what will happen when the assumptions do not hold.

In Appendix~\ref{app:more-related}, we introduce the details of several closely related works and their relations with Hedge Defense. Then, in Appendix~\ref{app:not-obfuscated}, we present the four categories of obfuscated gradient and analyze why Hedge Defense does not fall into any of them. Notably, in the original paper of the obfuscated gradient phenomena~\cite{obfuscated}, the authors conduct a case study for several previously believed solid defenses and break them with various approaches. Several of these examined defenses are also closely related to Hedge Defense as in Appendix~\ref{app:more-related}. Thus, we will illustrate why Hedge Defense can withstand the examinations that appeared in \cite{obfuscated} and thus outperform previous methods. Finally, in  Appendix~\ref{sec:discu}, we provide extra discussions about the newly proposed Hedge Defense.

\section{Extra Experimental Details and Extra Experiments}
\label{app:more-exp}

\subsection{Details of the Robust Models}
\label{app:rb-models}

Here we introduce the details of the 25 robust models we have tested.
Most of these state-of-the-art robust models are generated by adversarial training~\cite{fgsm,pgd} or similar training schemes~\cite{backward,self}. In particular, the basic Adversarial Training (AT)~\cite{pgd} is a very important work that was proposed a few years ago. Since it is not submitted to the RobustBench, we train its model with a PyTorch implementation and provide it in Table~\ref{table:cifar10}. Together with the other 7 robust models, Table~\ref{table:cifar10} includes 8 robust models. Also, RobustBench provides an unprotected model without adversarial training (Standard). We use this model official model of RobustBench for demonstration. Results on the rest 17 robust models will be shown in Appendix~\ref{app:cifar-extra}.

\subsection{Details of the Nine Attacks}
\label{app:attacks}

\paragraph{PGD:} PGD was first proposed in \cite{pgd}. Comparing with FGSM~\cite{fgsm}, the most important difference is that PGD uses multi-step optimization techniques. This modification is significant and followed by almost all the other white-box attacks. In our settings of evaluation, we set $20$ iterations of optimization with a step-size of $4/255$, and a random start of attacks~\cite{fastat} is implemented via uniformly sampling an example in $\mathbb{B}(\xb,\epsilon_\text{a})$.  PGD adopts the cross-entropy loss for optimization. Many following works investigate other choices of loss design~\cite{cw,autoattack} and achieve consistent improvements. Thus, recent works have considered PGD as a relatively weaker attack. In our work, we have shown that although PGD achieves worse attacking performance in the first place, it does show superiority against perturbations and thus becomes better than other attacks. In Algorithm~\ref{alg:pgd}, we present the detail of the PGD attack. Notice that Algorithm~\ref{alg:pgd} generally resembles Algorithm~\ref{alg:hd}, except for their criterion to be optimized completely different from each other.
\begin{algorithm}[H]
\small
\caption{\ Projected Gradient Descent}
\label{alg:pgd}
 \begin{algorithmic}[1]
    \STATE {{\bfseries Input}: the coming input $\xb$, the label $y$, the number of iterations $K$, the step size $\eta$, the deep network $f(\cdot)$,  and the defensive radius $\epsilon_\text{a}$.}
    \STATE{\emph{// $ \cU(\mathbf{-1},\mathbf{1}) $ generates a uniform noise}}
    \STATE {{\bfseries Initialization}: $\xb'_0 \gets \xb + \epsilon_\text{a} \ \cU(\mathbf{-1},\mathbf{1}), K=20, \eta=4/255, \epsilon_\text{a}=8/255 $}.
    \FOR{$k = 1 \dots K$}
    \STATE{$\xb'_k \gets \xb'_{k-1} +  \quad \eta\cdot\sign(\nabla_{\xb'_{k-1}} \cL(f(\xb'_{k-1}),y))$;}
    \STATE{$\xb'_k \gets \Pi_{\mathbb{B}(\xb,\epsilon_\text{a})} (\xb'_k)$;}
    \ENDFOR
    \STATE {{\bfseries Output}: the adversarial example $\xb'_K$.}
 \end{algorithmic}
\end{algorithm}

\paragraph{APGD-CE:} APGD-CE also uses the cross-entropy loss for optimization like PGD. The difference is that APGD-CE uses adaptive optimization step size and achieves better performance. Specifically, on each optimization step of $\xb'$, the hyper-parameters of $\eta$ will be updated according to a series of intuitive designs. The details are shown in Algorithm~\ref{alg:apgd}.
\begin{algorithm}[H]
\small
\caption{\ Auto Projected Gradient Descent}
\label{alg:apgd}
 \begin{algorithmic}[1]
    \STATE {{\bfseries Input}: the coming input $\xb$, the label $y$, the number of iterations $K$, the step size $\eta$, the deep network $f(\cdot)$, the update checkpoint $W=\{w_0, w_1, ... w_n\}$, the hyper-parameters $\alpha$, $\rho$, and the defensive radius $\epsilon_\text{a}$.}
    \STATE{\emph{// $ \cU(\mathbf{-1},\mathbf{1}) $ generates a uniform noise}}
    \STATE {{\bfseries Initialization}: $\epsilon_\text{a}=8/255$}.
    \STATE{$\xb'_0 \gets \xb$;}
    \STATE{$\xb'_1 \gets \xb'_0 + \eta\cdot\sign(\nabla_{\xb'_{0}} \cL(f(\xb'_0),y))$;}
    \STATE{$f_\text{max} \gets \max(f(\xb'_0),f(\xb'_1))$;}
    \STATE{$x_\text{max} \gets \xb'_0$ if $f_\text{max}\equiv f(\xb'_0)$ else $x_\text{max} \gets \xb'_1$;}
    \FOR{$k = 2 \dots K$}
    \STATE{$\zb'_k \gets \Pi_{\mathbb{B}(\xb,\epsilon_\text{a})} ( \xb'_{k-1} + \eta\cdot\sign(\nabla_{\xb'_{k-1}} \cL(f(\xb'_{k-1}),y)))$;}
    \STATE{$\xb'_k \gets \Pi_{\mathbb{B}(\xb,\epsilon_\text{a})} (\xb'_{k-1} + \alpha(\zb'_k - \xb'_{k-1}) + (1-\alpha)(\xb'_{k-1}-\xb'_{k-2}) )$;}
    \IF{$f(\xb'_k)>f_\text{max}$}
    \STATE {$x_\text{max} \gets \xb'_k$ and $f_\text{max} \gets f(\xb'_k)$.}
    \ENDIF
    \IF{$k\in W$ and $k\equiv w_j$}
    \IF{$\sum^{i=w_{j-1}}_{w_j-1} \mathbf{1}(f(x_{i+1})>f(x_i)) < \rho (w_j-w_{j-1})$ \OR ( $\eta_{w_{j-1}}\equiv \eta_{w_{j}}$ \AND $f_\text{max}^{w_{j-1}}  \equiv  f_\text{max}^{w_{j}} )$}
    \STATE {$\eta \gets \eta/2$ and $\xb'_k \gets x_\text{max}$.}
    \ENDIF
    \ENDIF
    \ENDFOR
    \STATE {{\bfseries Output}: the adversarial example $\xb'_K$.}
 \end{algorithmic}
\end{algorithm}

\paragraph{C\&W:} PGD intends to maximize the soft-max cross-entropy loss. In contrast, C\&W directly attacks the logits before the softmax layer. In previous works, it shows steadily better performance than the standard attack of PGD. Nevertheless, our works have shown that the adversarial examples generated by C\&W can be more vulnerable to perturbations. We will discuss this in detail in the next section.
\begin{equation}
\small
\textbf{C\&W:} \
\xb_\text{cw} = 
\argmax\limits_{\xb' \in \mathbb{B}(\xb,\epsilon_\text{a})}
(-z_y(\xb')+\max\limits_{i\neq y}z_{i}(\xb')).
\end{equation}

\paragraph{APGD-DLR and APGD-T:} Like C\&W, APGD-DLR also attacks the logits. \cite{autoattack} argues that the CW loss is not scaling invariant, and thus an extreme re-scaling could in principle be used to induce gradient masking. Thue, they introduce the difference between the largest logit $z_{\pi_1}(\xb')$ and the third-largest logit $z_{\pi_3}(\xb')$ to counter the potential scaling. APGD-T is its targeted version. It iterates among all the false classes as the targeted attacking aim and selects the worst-case among them. APGD-T can be more time-consuming but also more effective.
\begin{equation}
\small
\textbf{DLR:} \
\xb_\text{DLR} = 
\argmax\limits_{\xb' \in \mathbb{B}(\xb,\epsilon_\text{a})}
\left(
-
\frac{z_y(\xb')-\max\limits_{i\neq y}z_{i}(\xb')}
{z_{\pi_1}(\xb')-z_{\pi_3}(\xb')}
\right).
\end{equation}

\paragraph{FAB and DeepFool:} These two attacks are based on pushing examples to the other side of the decision boundary. The reason they intend to minimize the distance to the natural example $\| \xb' -\xb \|_p$ is to create adversarial examples that are more difficult to be detected by either visual clues or adversarial examples detection techniques. 

\begin{equation}
\small
\textbf{FAB:} \
\xb_\text{fab} = \argmin\limits_{\xb'} { \| \xb' -\xb \|_p} \quad
\st
\argmax\limits_{c}f_c(\xb') \neq y. 
\end{equation}

\paragraph{Square:} Square is the state-of-the-art black-box attack based on extensively querying the model. Because it cannot get access to the gradient information of the model, it has to establish some assumptions about the landscape of deep neural networks and conduct searching based on the assumption. Although slightly worse than the white-box attacks and more time-consuming, its performance on the successful attacking rate is still overwhelming and significantly endangers the deployment of deep networks.

\paragraph{RayS:} RayS is the state-of-the-art hard-label black-box attack. Namely, black-box attacks like Square can utilize the predicted distribution of networks (the output of the softmax layer), while the hard-label attack can only use the discrete prediction of the model. This makes hard-label attacks the most difficult attacking tasks currently, and RayS has to search adversarial examples based on more restricted assumptions. Its performance on adversarially-trained models is slightly lower than Square's, but the total successful attacking rate is still promising.

\subsection{Case Study for Section~\ref{sec:exp}}
\label{app:case-study}

In Section~\ref{sec:cifar}, we briefly analyze why different attacks react differently to random or defensive perturbations. Here we offer a complete version of it.

\textbf{Square and RayS:} Square and RayS are black-box attacks based on extensively querying deep networks. Typically, they need to conduct about 1k to 40k queries to attack a single example. On an NVIDIA V100 GPU, typical white-box attacks like PGD take a few seconds to attack the $10000$ examples of the CIFAR10 test set, while Square and RayS may cost $20$ to $30$ hours in different settings. Therefore, they tend to stop queries once they find a successful adversarial example to save the computation budget. Nevertheless, such a design makes attacks stop at the adversarial examples closer to the unharmful example, which is the example one step earlier before the successful attacking query. This explains why they are so vulnerable to defensive perturbations. The black-box constraint of not knowing the gradient information of the model may also prevent them from finding more powerful adversarial examples that are far away from the decision boundary. Explicitly speaking, to counter defensive perturbations, black-box attacks may well need to find a region-wise falsehood, a small local space that is full of false predictions. Without knowing the gradient information and the landscape of the loss surface, this aim can be tough to fulfill. 

\textbf{FAB:} Unlike other white-box attacks that search adversarial examples within $\mathbb{B}(\xb,\epsilon_\text{a})$, FAB tries to find a minimal perturbation for attacks, as shown above. As discussed previously, this design allows FAB to generate more undetectable adversarial examples. However, this aim of finding smaller attacking perturbations also encourages them to find adversarial examples that are too close to the decision boundary and thus become very vulnerable to defensive perturbations. Thus, even on the undefended standard model where they generate smaller perturbations than on robust models, they can still be hugely nullified by random perturbation, and the robust accuracy increases from less than $1.0\%$ to $74.57\%$ and $84.37\%$. It is easy to tell that FAB is vulnerable to Hedge Defense primarily due to its intrinsic design. With a proper adaption, FAB may alleviate the non-robustness to the emerge defensive perturbations, although questions like how they can keep a proper distance to the decision boundary and whether the new design degrades their performance remains unclear.

\textbf{PGD and C\&W:} Attacks like PGD and C\&W show better resistance against perturbations. However, with Hedge Defense, the robust accuracy can still be improved by $3.0\% $ to $11.0\%$. Interestingly, several commonly believed more powerful attacks turn out to be weaker when facing perturbations. For instance, C\&W enhances attacks by increasing the largest false logit $z_{i}(\xb') (i\neq y)$ and decreasing the true logit $z_y(\xb')$. Without Hedge Defense, it is more powerful ($54.47\%$ robust accuracy of TRADES) than the PGD attack ($55.83\%$). With Hedge Defense, it becomes less effective ($69.98\%$ vs. $58.99\%$). This indicates that its efficacy comes at a price of less robustness against defensive perturbation. Furthermore, the effect of directly increasing the false logit can be weakened if the false class is attacked. 

\textbf{APGD-CE and APGD-T:} APGD-CE is a direct improvement for PGD. However, even after random perturbation, the improvement brought by APGD-CE can be nullified compared with the standard PGD. APGD-T does bring consistent improvements in many cases. However, APGD-T or APGD-DLR may also lag behind the previously believed weaker PGD attack from time to time. Thus, future works may still need to put effort into examining how these intuitively designed losses help attacks. 

Finally, we want to compare the robustness of attacks and the robustness of defenses with a unified perspective. As studies in previous works~\cite{tes19,trades}, adversarial robustness may be at odds with natural accuracy. Similarly, we can observe that stronger attacks may bring a higher successful attacking rate but less robustness against defensive perturbation in our evaluations. This important finding gives defenses a chance to fight back and break attacks with their methodology. The above analyses show that state-of-the-art attacks are not robust due to various reasons. In the future, when they alleviate the problem and enhance their attacking robustness, they may not attack robust models at the currently high successful rate, resembling that robust deep networks all have lower natural accuracy.

\newpage
\subsection{Extra Results of Table~\ref{table:cifar10}}
\label{app:cifar-extra}

Here we provide extra results for our evaluations in Table~\ref{table:cifar10}, including the variances for the robust accuracy evaluations in Table~\ref{table:cifar10} and evaluations on the rest 17 robust models. Besides the reference of the original work, we also provide the IDs of these models on RobustBench. Notice that, for the eight robust models and the one standard model, we evaluate three times, reporting the average value in Table~\ref{table:cifar10} and presenting the variances in Table~\ref{table:cifar10-var}. For the rest 17 robust models in Table~\ref{table:cifar10-extra}, we only conduct a single time of evaluation.

As shown below, compared with the improvement brought by random perturbation and Hedge Defense, the variances of them are primarily around $0.1\%$, indicating their improvement is stable and reliable. Later we will show that, even with EOT~\cite{obfuscated}, the improvements of defensive perturbation are still solid.

\begin{table*}[h!]
	\caption{The variance for Table~\ref{table:cifar10}.}
	\label{table:cifar10-var}
	\centering
	\vskip -0.1in
	\resizebox{0.8\textwidth}{!}{
	\begin{tabular}{ cccccccccccccc } 
		\toprule
		Model & Method 
		& PGD  & C\&W 
		& \tabincell{c}{Deep\\Fool}
		& \tabincell{c}{APGD\\CE}  
		& \tabincell{c}{APGD\\T}
		& FAB & Square & RayS 
		& \tabincell{c}{Auto\\Attack}
		& \tabincell{c}{Worst\\Case} \\
		\midrule
		\midrule
		\multirow{3}{*}{\tabincell{c}{WA+\\SiLU}~\cite{uncover}} 
		& -       & 0.04 & 0.02 & 0.00    & 0.01 & 0.01 & 0.00    & 0.05 & 0    & 0.01 & 0.01 \\
        & Uniform & 0.03 & 0.09 & 0.10  & 0.03 & 0.10  & 0.09 & 0.14 & 0.12 & 0.03 & 0.03 \\
        & Hedge   & 0.04 & 0.06 & 0.04 & 0.00    & 0.04 & 0.03 & 0.05 & 0.03 & 0.05 & 0.05 \\
		\midrule
		\multirow{3}{*}{AWP~\cite{awp}} 
		& -       & 0.03 & 0.02 & 0.00 & 0.02 & 0.01 & 0.01 & 0.04 & 0.00 & 0.00 & 0.01 \\
        & Uniform & 0.07 & 0.08 & 0.03 & 0.00 & 0.05 & 0.02 & 0.10 & 0.12 & 0.08 & 0.10 \\
        & Hedge   & 0.05 & 0.10 & 0.03 & 0.05 & 0.02 & 0.14 & 0.10 & 0.05 & 0.01 & 0.01 \\
		\midrule
		\multirow{3}{*}{RST~\cite{rst}} 
		& -       & 0.02 & 0.01 & 0.00 & 0.01 & 0.01 & 0.01 & 0.02 & 0.00 & 0.02 & 0.01 \\
        & Uniform & 0.07 & 0.13 & 0.03 & 0.11 & 0.00 & 0.07 & 0.15 & 0.11 & 0.04 & 0.04 \\
        & Hedge   & 0.08 & 0.09 & 0.06 & 0.06 & 0.06 & 0.07 & 0.04 & 0.01 & 0.03 & 0.03 \\
		\midrule
		\multirow{3}{*}{\tabincell{c}{Pre-\\Training}~\cite{pretrain}} 
		& -       & 0.04 & 0.01 & 0.00 & 0.01 & 0.02 & 0.02 & 0.03 & 0.00 & 0.00 & 0.02 \\
        & Uniform & 0.03 & 0.07 & 0.06 & 0.07 & 0.07 & 0.06 & 0.06 & 0.28 & 0.06 & 0.03 \\
        & Hedge   & 0.02 & 0.11 & 0.05 & 0.06 & 0.04 & 0.06 & 0.10 & 0.04 & 0.10 & 0.04 \\
		\midrule
		\multirow{3}{*}{MART~\cite{mart}} 
		& -       & 0.04 & 0.03 & 0.00 & 0.01 & 0.01 & 0.03 & 0.02 & 0.00 & 0.01 & 0.02 \\
        & Uniform & 0.10 & 0.10 & 0.19 & 0.04 & 0.09 & 0.06 & 0.14 & 0.12 & 0.08 & 0.03 \\
        & Hedge   & 0.02 & 0.04 & 0.02 & 0.07 & 0.06 & 0.03 & 0.05 & 0.06 & 0.06 & 0.02 \\ 
		\midrule
		\multirow{3}{*}{HYDRA~\cite{hydra}} 
		& -       & 0.03 & 0.05 & 0.00 & 0.02 & 0.00 & 0.01 & 0.01 & 0.00 & 0.01 & 0.01 \\
        & Uniform & 0.07 & 0.13 & 0.09 & 0.05 & 0.05 & 0.16 & 0.17 & 0.27 & 0.01 & 0.05 \\
        & Hedge   & 0.04 & 0.07 & 0.04 & 0.03 & 0.05 & 0.05 & 0.07 & 0.03 & 0.04 & 0.03 \\
		\midrule
		\multirow{3}{*}{TRADES~\cite{trades}} 
		& -       & 0.03 & 0.01 & 0.00 & 0.02 & 0.01 & 0.02 & 0.03 & 0.00 & 0.01 & 0.00 \\
        & Uniform & 0.04 & 0.08 & 0.05 & 0.08 & 0.06 & 0.06 & 0.2  & 0.24 & 0.04 & 0.02 \\
        & Hedge   & 0.02 & 0.07 & 0.03 & 0.03 & 0.05 & 0.05 & 0.02 & 0.01 & 0.03 & 0.03 \\
		\midrule
		\multirow{3}{*}{AT~\cite{pgd}} 
		& -       & 0.03 & 0.10 & 0.07 & 0.04 & 0.10 & 0.10 & 0.09 & 0.00 & 0.01 & 0.05 \\
        & Uniform & 0.00 & 0.29 & 0.10 & 0.04 & 0.04 & 0.10 & 0.17 & 0.01 & 0.08 & 0.10 \\
        & Hedge   & 0.01 & 0.08 & 0.14 & 0.04 & 0.02 & 0.04 & 0.01 & 0.08 & 0.08 & 0.07 \\
		\midrule
		\multirow{3}{*}{Standard} 
		& -       & 0.00 & 0.00 & 0.00 & 0.00 & 0.00 & 0.01 & 0.04 & 0.00 & 0.00 & 0.00 \\ 
        & Uniform & 0.00 & 0.23 & 0.21 & 0.01 & 0.04 & 0.20 & 0.34 & 0.35 & 0.07 & 0.00 \\ 
        & Hedge   & 0.00 & 0.42 & 0.23 & 0.05 & 0.02 & 0.19 & 0.23 & 0.25 & 0.02 & 0.00 \\
		\bottomrule
	\end{tabular}}
	\vskip -0.1in
\end{table*}

\textbf{IDs on RobustBench (not including AT):} 

Gowal2020Uncovering\_70\_16\_extra 

Wu2020Adversarial\_extra 

Carmon2019Unlabeled 

Hendrycks2019Using 

Wang2020Improving 

Sehwag2020Hydra 

Zhang2019Theoretically 

Standard


\newpage

\begin{table*}[h!]
	\caption{Seventeen extra robust models for Table~\ref{table:cifar10}}
	\label{table:cifar10-extra}
	\centering
	\vskip 0.05in
	\resizebox{0.8\textwidth}{!}{
	\begin{tabular}{ cccccccccccccc } 
		\toprule
		Model & Method 
		& \tabincell{c}{Nat-\\Acc.} 
		& PGD  & C\&W 
		& \tabincell{c}{Deep\\Fool}
		& \tabincell{c}{APGD\\CE}  
		& \tabincell{c}{APGD\\T}
		& FAB & Square & RayS 
		& \tabincell{c}{Auto\\Attack}
		& \tabincell{c}{Worst\\Case} \\
		\midrule
		\midrule
		\multirow{3}{*}{\cite{uncover}} & -              & 89.48             & 66.59          & 64.46          & 67.62             & 65.92            & 63.25           & 63.77          & 69.07           & 69.3           & 63.24                & 63.22          \\
                           & Random         & 89.21             & 66.58          & 68.08          & 71.89             & 66.90            & 64.35           & 75.54          & 75.30           & 77.12          & 64.80                & 63.08          \\
                           & \textbf{Hedge} & 89.16             & \textbf{69.04} & \textbf{76.18} & \textbf{77.26}    & \textbf{71.47}   & \textbf{70.33}  & \textbf{80.15} & \textbf{81.11}  & \textbf{79.99} & \textbf{70.07}       & \textbf{65.70} \\
                           \midrule
        \multirow{3}{*}{\cite{geometry}} & -              & 89.36             & 68.15          & 61.18          & 64.09             & 66.37            & 59.64           & 60.16          & 66.18           & 66.82          & 59.61                & 59.61          \\
                           & Random         & 89.33             & 68.06          & 65.33          & 74.79             & 67.54            & 60.93           & 74.92          & 73.56           & 76.47          & 61.65                & 59.74          \\
                           & \textbf{Hedge} & 90.16             & \textbf{71.03} & \textbf{77.8}  & \textbf{81.45}    & \textbf{73.30}   & \textbf{69.08}  & \textbf{82.54} & \textbf{82.44}  & \textbf{83.07} & \textbf{69.89}       & \textbf{64.43} \\
                           \midrule
        \multirow{3}{*}{\cite{uncover}} & -              & 88.70             & 56.38          & 55.38          & 60.42             & 55.49            & 53.56           & 54.09          & 61.96           & 63.01          & 53.57                & 53.52          \\
                           & Random         & 88.61             & 56.60          & 60.55          & 69.20             & 56.68            & 55.06           & 72.32          & 71.56           & 74.81          & 55.19                & 53.59          \\
                           & \textbf{Hedge} & 88.77             & \textbf{58.58} & \textbf{69.45} & \textbf{73.56}    & \textbf{59.84}   & \textbf{59.35}  & \textbf{76.67} & \textbf{77.11}  & \textbf{78.06} & \textbf{58.61}       & \textbf{55.44} \\
                           \midrule
        \multirow{3}{*}{\cite{awp}} & -              & 85.36             & 59.66          & 57.51          & 60.87             & 59.02            & 56.52           & 57.04          & 61.56           & 61.84          & 56.53                & 56.51          \\
                           & Random         & 85.22             & 59.80          & 61.49          & 68.71             & 59.99            & 57.82           & 71.61          & 69.74           & 72.59          & 58.15                & 56.67          \\
                           & \textbf{Hedge} & 84.89             & \textbf{62.14} & \textbf{70.47} & \textbf{72.57}    & \textbf{65.07}   & \textbf{64.06}  & \textbf{75.04} & \textbf{75.13}  & \textbf{74.93} & \textbf{63.87}       & \textbf{59.15} \\
                           \midrule
        \multirow{3}{*}{\cite{boosting}} & -              & 85.14             & 62.50          & 56.27          & 58.64             & 61.73            & 54.30           & 54.71          & 61.59           & 61.62          & 54.27                & 54.24          \\
                           & Random         & 84.86             & 62.74          & 60.45          & 68.40             & 62.74            & 55.54           & 69.08          & 68.95           & 72.09          & 55.93                & 54.44          \\
                           & \textbf{Hedge} & 85.59             & \textbf{66.04} & \textbf{71.42} & \textbf{74.05}    & \textbf{70.79}   & \textbf{65.48}  & \textbf{76.06} & \textbf{77.74}  & \textbf{76.25} & \textbf{66.26}       & \textbf{58.15} \\
                           \midrule
        \multirow{3}{*}{\cite{learnable}} & -              & 88.70             & 56.38          & 55.38          & 60.42             & 55.49            & 53.56           & 54.09          & 61.96           & 63.01          & 53.57                & 53.52          \\
                           & Random         & 88.61             & 56.60          & 60.55          & 69.20             & 56.68            & 55.06           & 72.32          & 71.56           & 74.81          & 55.19                & 53.59          \\
                           & \textbf{Hedge} & 88.77             & \textbf{58.58} & \textbf{69.45} & \textbf{73.56}    & \textbf{59.84}   & \textbf{59.35}  & \textbf{76.67} & \textbf{77.11}  & \textbf{78.06} & \textbf{58.61}       & \textbf{55.44} \\
                           \midrule
        \multirow{3}{*}{\cite{friendly}} & -              & 84.52             & 57.78          & 55.37          & 59.01             & 56.96            & 54.10           & 54.58          & 59.84           & 60.16          & 54.06                & 54.06          \\
                           & Random         & 84.35             & 58.09          & 59.47          & 64.66             & 58.18            & 55.26           & 69.42          & 68.12           & \textbf{71.32} & 55.85                & 54.25          \\
                           & \textbf{Hedge} & 81.76             & \textbf{59.96} & \textbf{64.76} & \textbf{66.66}    & \textbf{62.74}   & \textbf{61.09}  & \textbf{70.25} & \textbf{70.79}  & 69.42          & \textbf{61.33}       & \textbf{55.97} \\
                           \midrule
        \multirow{3}{*}{\cite{overfitting}} & -              & 85.34             & 57.98          & 56.21          & 58.46             & 57.25            & 53.92           & 54.31          & 61.74           & 61.66          & 53.94                & 53.87          \\
                           & Random         & 85.13             & 58.10          & 60.15          & 69.09             & 58.20            & 55.24           & 69.36          & 69.66           & 72.53          & 55.45                & 53.81          \\
                           & \textbf{Hedge} & 85.85             & \textbf{62.16} & \textbf{73.72} & \textbf{75.42}    & \textbf{66.69}   & \textbf{65.79}  & \textbf{76.65} & \textbf{78.7}   & \textbf{77.70} & \textbf{64.75}       & \textbf{58.59} \\
                           \midrule
        \multirow{3}{*}{\cite{self}} & -              & 83.48             & 56.80          & 54.76          & 58.65             & 56.08            & 53.38           & 53.90          & 58.88           & 59.47          & 53.35                & 53.31          \\
                           & Random         & 83.40             & 56.86          & 58.83          & 64.29             & 57.04            & 54.77           & 69.25          & 67.01           & 70.63          & 54.74                & 53.40          \\
                           & \textbf{Hedge} & 82.25             & \textbf{59.66} & \textbf{68.03} & \textbf{68.84}    & \textbf{62.55}   & \textbf{61.78}  & \textbf{72.33} & \textbf{72.72}  & \textbf{71.98} & \textbf{61.16}       & \textbf{56.34} \\
                           \midrule
        \multirow{3}{*}{\cite{learnable}} & -              & 88.22             & 55.23          & 54.83          & 61.12             & 54.30            & 52.90           & 53.51          & 61.22           & 62.59          & 52.87                & 52.82          \\
                           & Random         & 88.05             & 55.63          & 60.23          & 68.67             & 55.56            & 54.54           & 72.11          & 70.98           & 74.07          & 54.39                & 52.95          \\
                           & \textbf{Hedge} & 88.57             & \textbf{57.72} & \textbf{69.13} & \textbf{72.57}    & \textbf{58.66}   & \textbf{58.79}  & \textbf{76.18} & \textbf{76.17}  & \textbf{77.61} & \textbf{57.74}       & \textbf{54.63} \\
                           \midrule
        \multirow{3}{*}{\cite{chen2020adversarial}} & -              & 86.04             & 54.86          & 52.86          & 56.47             & 54.35            & 52.12           & 52.51          & 58.47           & 59.77          & 52.09                & 52.08          \\
                           & Random         & 85.86             & 55.14          & 57.16          & 63.15             & 55.44            & 53.29           & 68.78          & 67.47           & 72.01          & 53.48                & 51.97          \\
                           & \textbf{Hedge} & 85.99             & \textbf{55.97} & \textbf{62.93} & \textbf{66.97}    & \textbf{56.84}   & \textbf{55.14}  & \textbf{72.26} & \textbf{71.60}  & \textbf{74.60} & \textbf{55.47}       & \textbf{53.01} \\
                           \midrule
        \multirow{3}{*}{\cite{backward}} & -              & 85.32             & 55.49          & 53.84          & 57.90             & 54.06            & 51.59           & 52.29          & 58.62           & 59.44          & 51.57                & 51.55          \\
                           & Random         & 85.19             & 55.88          & 59.11          & 65.85             & 55.63            & 53.32           & \textbf{71.78} & 68.66           & 72.47          & 53.67                & 51.99          \\
                           & \textbf{Hedge} & 82.52             & \textbf{58.67} & \textbf{67.3}  & \textbf{68.99}    & \textbf{61.30}   & \textbf{60.92}  & 71.68          & \textbf{72.30}  & \textbf{71.54} & \textbf{60.09}       & \textbf{55.31} \\
                           \midrule
        \multirow{3}{*}{\cite{Sitawarin}} & -              & 86.84             & 54.41          & 54.01          & 57.17             & 53.25            & 51.46           & 51.92          & 59.75           & 60.53          & 51.42                & 51.38          \\
                           & Random         & 86.69             & 54.86          & 59.19          & 69.52             & 54.54            & 52.91           & 70.66          & 69.54           & 73.01          & 53.08                & 51.39          \\
                           & \textbf{Hedge} & 86.11             & \textbf{57.66} & \textbf{70.57} & \textbf{73.51}    & \textbf{60.31}   & \textbf{60.94}  & \textbf{74.21} & \textbf{76.52}  & \textbf{75.95} & \textbf{59.13}       & \textbf{54.58} \\
                           \midrule
        \multirow{3}{*}{\cite{robustness}} & -              & 87.03             & 53.49          & 53.25          & 56.48             & 52.09            & 49.93           & 50.36          & 58.34           & 59.79          & 49.87                & 49.84          \\
                           & Random         & 86.99             & 53.76          & 58.70          & 70.38             & 53.39            & 51.63           & 70.89          & 69.00           & 73.26          & 51.86                & 50.04          \\
                           & \textbf{Hedge} & 86.00             & \textbf{56.83} & \textbf{70.96} & \textbf{72.30}    & \textbf{59.73}   & \textbf{59.74}  & \textbf{73.39} & \textbf{76.13}  & \textbf{75.01} & \textbf{57.96}       & \textbf{53.42} \\
                           \midrule
        \multirow{3}{*}{\cite{you}} & -              & 87.20             & 47.39          & 48.16          & 51.99             & 46.53            & 45.44           & 45.98          & 55.50           & 55.98          & 45.42                & 45.39          \\
                           & Random         & 87.17             & 47.75          & 53.15          & 65.50             & 47.44            & 46.76           & 67.79          & 66.55           & 71.60          & 46.77                & 45.38          \\
                           & \textbf{Hedge} & 86.73             & \textbf{50.91} & \textbf{68.06} & \textbf{72.09}    & \textbf{53.46}   & \textbf{55.50}  & \textbf{73.59} & \textbf{75.45}  & \textbf{75.32} & \textbf{52.71}       & \textbf{48.50} \\
                           \midrule
        \multirow{3}{*}{\cite{fastat}} & -              & 83.34             & 47.57          & 47.17          & 50.20             & 46.23            & 43.78           & 44.48          & 53.68           & 54.15          & 43.75                & 43.71          \\
                           & Random         & 83.36             & 48.21          & 52.57          & 64.16             & 47.69            & 45.33           & 66.28          & 63.98           & 68.41          & 45.70                & 43.99          \\
                           & \textbf{Hedge} & 81.76             & \textbf{52.31} & \textbf{66.03} & \textbf{68.94}    & \textbf{55.09}   & \textbf{55.42}  & \textbf{69.87} & \textbf{71.80}  & \textbf{71.54} & \textbf{53.91}       & \textbf{49.13} \\
                           \midrule
        \multirow{3}{*}{\cite{mma}} & -              & 84.36             & 52.38          & 53.28          & 58.39             & 49.67            & 42.24           & 43.27          & 55.82           & 51.49          & 41.94                & 41.83          \\
                           & Random         & 83.92             & 52.68          & 57.94          & 67.10             & 50.66            & 44.97           & 67.98          & 64.65           & 66.52          & 44.48                & 42.93          \\
                           & \textbf{Hedge} & 82.99             & \textbf{54.59} & \textbf{66.56} & \textbf{69.62}    & \textbf{54.74}   & \textbf{49.95}  & \textbf{68.12} & \textbf{70.68}  & \textbf{68.64} & \textbf{48.80}       & \textbf{43.49} \\
		\bottomrule
	\end{tabular}}
	\vskip -0.1in
\end{table*}

\newpage

\textbf{IDs on the RobustBench:} 

Gowal2020Uncovering\_28\_10\_extra, 

Zhang2020Geometry, 

Gowal2020Uncovering\_34\_20,

Wu2020Adversarial, 

Pang2020Boosting, 

Cui2020Learnable\_34\_20,

Zhang2020Attacks, 

Rice2020Overfitting,

Huang2020Self, 

Cui2020Learnable\_34\_10, 

Chen2020Adversarial, 

Chen2020Efficient, 

Sitawarin2020Improving, 

Engstrom2019Robustness, 

Zhang2019You, 

Wong2020Fast, 
Ding2020MMA

\subsection{Extra Visualization of Figure~\ref{fig:visual}}
\label{app:more-visual}

\begin{figure}[h!]
	\centering
	\includegraphics[width=0.99\linewidth]{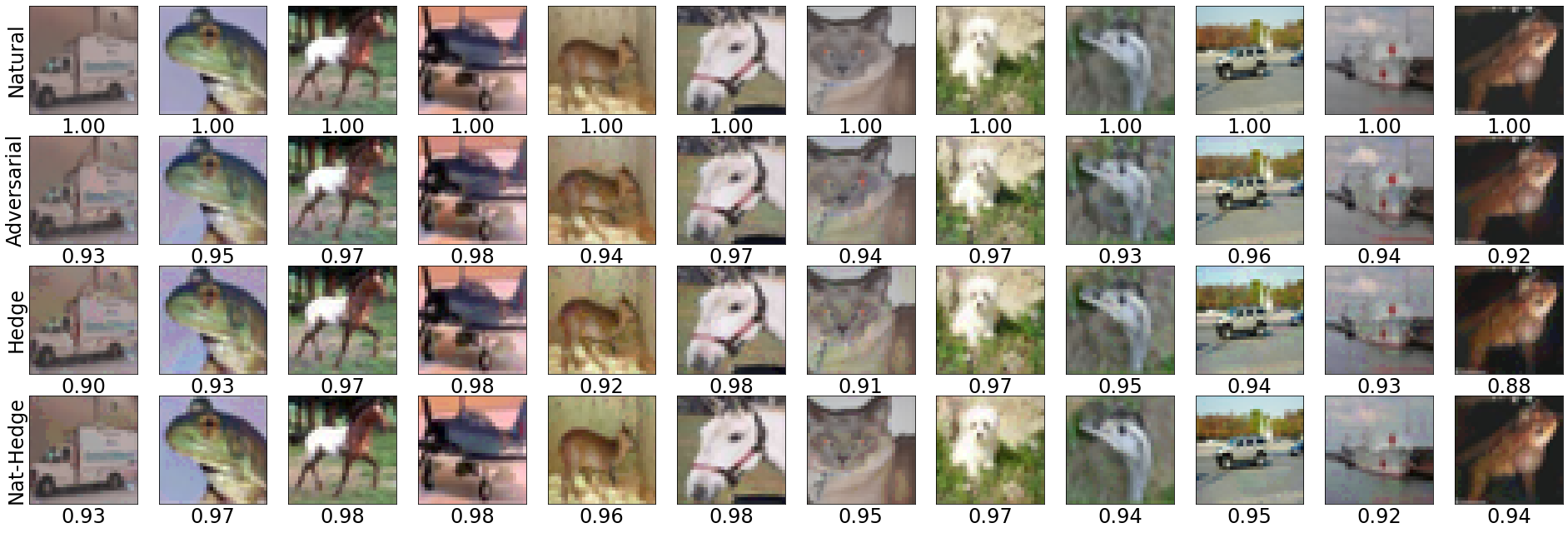}
	\caption{Extra visualizations for Figure~\ref{fig:visual} on examples from FtoT.}
	\label{fig:vis-f2t}
\end{figure}

\begin{figure}[h!]
	\centering
	\includegraphics[width=0.99\linewidth]{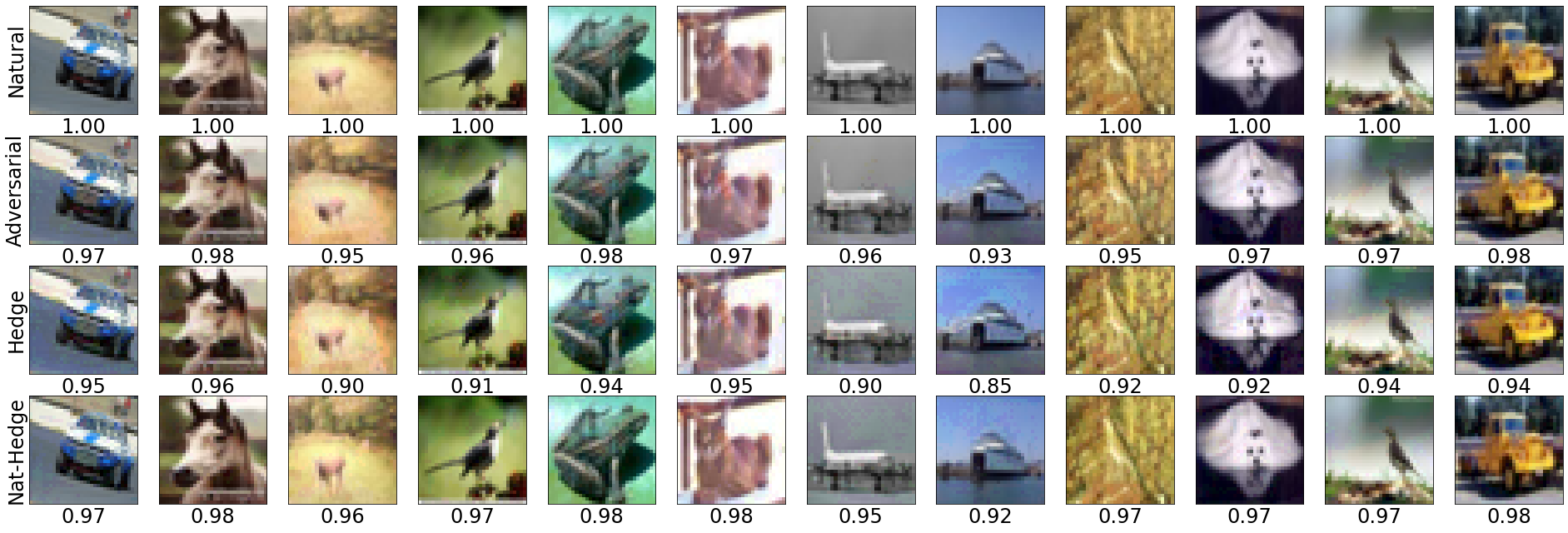}
	\caption{Extra visualizations for Figure~\ref{fig:visual} on examples from TtoF.}
	\label{fig:vis-t2f}
\end{figure}

We present extra visualizations for Section~\ref{sec:visual} on CIFAR10. Figure~\ref{fig:vis-f2t} presents examples of FtoT for the TRADES model. Namely, the TRADES model classifies the natural examples in the first row and then wrongly classifies the adversarial examples in the second row. After perturbing the adversarial examples with Hedge Defense, the generated hedge examples in the third row are correctly classified. In the first row, we also present the visualizations for applying Hedge Defense to the natural examples in the fourth row. In contrast, Figure~\ref{fig:vis-t2f} presents examples that belong to TtoF. The SSIM scores are shown beneath the images.

\subsection{Other Possibilities of Searching for the Defensive Perturbation}

In this part, we test several intuitive and alternative designs that may help Hedge Defense. We first introduce some randomness in Hedge Defense. Previous works~\cite{PinotERCA20} have pointed out that such randomness may help defensive robustness, although not all randomness techniques will help~\cite{Xie18}. Instead of simultaneously attacking all the classes, we randomly select a class to attack in each iteration, as shown below. This approach steadily achieves similar improvements to the standard Hedge Defense. More interestingly, on the undefended standard model, this random Hedge Defense can achieve around $1/C$ robust accuracy against the PGD attack. We will present a complete evaluation of this alternative method in a future version. Another method we tested is directly maximizing the KL-Divergence to the uniform distribution. Not surprisingly, it achieves the exact same performance as our Algorithm~1. 

\begin{algorithm}[H]
\small
\caption{\ Hedge Defense with Randomness}
\label{alg:hd-rand}
 \begin{algorithmic}[1]
    \STATE {{\bfseries Input}: the coming input $\xb'$,  the number of iterations $K$, the step size $\eta$,  the deep network $f(\cdot)$,  and the defensive radius $\epsilon_\text{d}$.}
    \STATE{\emph{// $ \cU(\mathbf{-1},\mathbf{1}) $ generates a uniform noise}}
    \STATE {{\bfseries Initialization}: $\xb''_0 \gets \xb' + \epsilon_\text{d} \ \cU(\mathbf{-1},\mathbf{1}) $}.
    \FOR{$k = 1 \dots K$}
    \STATE{\emph{// Randomly draw a class $c_\text{draw}$ from $1$ to $C$,}}
    \STATE{$\xb''_k \gets \xb''_{k-1} + $ \\ $ \quad \eta\cdot\sign(\nabla_{\xb''_{k-1}} \cL(f(\xb''_{k-1}),c_\text{draw}))$;}
    \STATE{\emph{// $\Pi$ is the projection operator.}}
    \STATE{$\xb''_k \gets \Pi_{\mathbb{B}(\xb',\epsilon_\text{d})} (\xb''_k)$;}
    \ENDFOR
    \STATE {{\bfseries Output}: the safer prediction $f(\xb''_K)$.}
 \end{algorithmic}
\end{algorithm}

\subsection{Reversing Hedge Defense as a label-free Attack}

\label{app:label-free}

\begin{algorithm}[H]
\small
\caption{\ The Label-Free Attack}
\label{alg:label-free}
 \begin{algorithmic}[1]
    \STATE {{\bfseries Input}: the coming input $\xb'$,  the number of iterations $K$, the step size $\eta$,  the deep network $f(\cdot)$,  and the defensive radius $\epsilon_\text{d}$.}
    \STATE{\emph{// $ \cU(\mathbf{-1},\mathbf{1}) $ generates a uniform noise}}
    \STATE {{\bfseries Initialization}: $\xb''_0 \gets \xb' + \epsilon_\text{d} \ \cU(\mathbf{-1},\mathbf{1}) $}.
    \FOR{$k = 1 \dots K$}
    \STATE{\emph{// Sum the losses for all the classes,}}
    \STATE{\emph{// then update $\xb''_k$ with gradient descending.}}
    \STATE{$\xb''_k \gets \xb''_{k-1} - $ \\ $ \quad \eta\cdot\sign(\nabla_{\xb''_{k-1}}\sum_{c=1}^{C} \cL(f(\xb''_{k-1}),c))$;}
    \STATE{\emph{// $\Pi$ is the projection operator.}}
    \STATE{$\xb''_k \gets \Pi_{\mathbb{B}(\xb',\epsilon_\text{d})} (\xb''_k)$;}
    \ENDFOR
    \STATE {{\bfseries Output}: the safer prediction $f(\xb''_K)$.}
 \end{algorithmic}
\end{algorithm}

In Section~\ref{sec:related}, we have mentioned that Hedge Defense essentially pushes examples away from the uniform distribution. Now we demonstrate that its reversing process, pulling examples to the uniform distribution, is also a form of adversarial attacks that do not require knowing the ground-truth labels. This label-free attack has the same loss function as Hedge Defense. However, when updating examples, it adopts the gradient descending method instead of the gradient ascending in Hedge Defense (Line~7 in Algorithm~\ref{alg:label-free}). We test its effectiveness on the undefended standard model and the robust model of TRADES in Table~\ref{table:label-free}. Apparently, due to the lack of knowing the ground-truth label, our label-free attack is not as powerful as those who can access the label. On the adversarially-trained models, its performance is also not very promising. Thus, its usage is relatively limited. However, the fact that the reverse Hedge Defense is an attacking process helps us understand that why Hedge Defense can bring extra robustness, which is the intention of the empirical evaluations in this part.

\begin{table}[h!]
	\caption{The lable-free attack on the standard model and the TRADES model.}
	\label{table:label-free}
	\centering
	\vskip 0.05in
\resizebox{\textwidth}{!}{
	\begin{tabular}{ cccc } 
		\toprule
		Method 
		& Model 
		& Natural Accuracy (\%)
		& Robust Accuracy (\%)\\
		\midrule
		\multirow{2}{*}{Label-Free Attack (The Reverse Hedge Defense)} 
		& Standard
		& 94.78 & 21.06 \\ 
		& TRADES
		& 84.92 & 79.97 \\ 
		\bottomrule
	\end{tabular}}
	\vskip -0.1in
\end{table}

\subsection{L2 norm}
\label{app:l2}

In this part, we present evaluations on the relatively weaker attacks of $\ell_2$ norm attacks on CIFAR10. As shown by Table~\ref{table:l2}, the robust accuracy against $\ell_2$ norm attacks is much higher than those of the $\ell_\infty$ norm attacks. Recent works~\cite{Tramer20} have illustrated that the $\ell_2$ norm may not keep the semantics of $\xb$ unchanged. 

\begin{table*}[h!]
	\caption{$\ell_2$ norm attacks with $\epsilon_\text{a}=0.5$ on CIFAR10.}
	\label{table:l2}
	\centering
	\vskip 0.05in
	\resizebox{\textwidth}{!}{
	\begin{tabular}{ ccccccccc } 
		\toprule
		Model & Method 
		& \tabincell{c}{Nat-\\Acc.} 
		& \tabincell{c}{APGD\\CE}  
		& \tabincell{c}{APGD\\T}
		& FAB & Square 
		& \tabincell{c}{Auto\\Attack}
		& \tabincell{c}{Worst\\Case} \\
		\midrule
		\midrule
        \multirow{3}{*}{Wu2020Adversarial~\cite{awp}}   
        & -              & 88.51 & 74.72          & 73.66          & 73.85          & 80.28          & 73.66          & 73.66          \\
        & Uniform        & 88.51 & 75.26          & 74.19          & \textbf{80.73} & 83.62          & 74.41          & 74.08          \\
        & \textbf{Hedge} & 87.52 & \textbf{76.71} & \textbf{76.02} & 80.59          & \textbf{84.12} & \textbf{76.17} & \textbf{75.91} \\
        \midrule
        \multirow{3}{*}{Rice2020Overfitting~\cite{overfitting}}
        & -              & 88.67 & 68.55          & 67.68          & 68.01          & 78.28          & 67.68          & 67.68          \\
        & Uniform        & 88.73 & 69.08          & 68.35          & \textbf{78.79} & 82.68          & 68.45          & 68.14          \\
        & \textbf{Hedge} & 87.10 & \textbf{73.55} & \textbf{73.25} & 78.53          & \textbf{84.01} & \textbf{73.09} & \textbf{72.68} \\
        \midrule
        \multirow{3}{*}{Ding2020MMA~\cite{mma}}         
        & -              & 88.02 & 66.21          & 66.09          & 66.36          & 76.08          & 66.09          & 66.09          \\
        & Uniform        & 88.06 & 66.47          & 66.41          & 77.04          & 79.36          & 66.30          & 66.17          \\
        & \textbf{Hedge} & 86.82 & \textbf{70.09} & \textbf{70.62} & \textbf{77.07} & \textbf{82.44} & \textbf{70.07} & \textbf{69.79} \\
		\bottomrule
	\end{tabular}}
	\vskip -0.1in
\end{table*}

\subsection{Extra Results of Analytical Experiments and Ablation Study}
We provide extra results of the analytical experiments and the ablation study in Section~\ref{sec:exp}.

\begin{figure}[h!]
		\centering

		\subfigure[]{
			\includegraphics[width=0.23\linewidth]{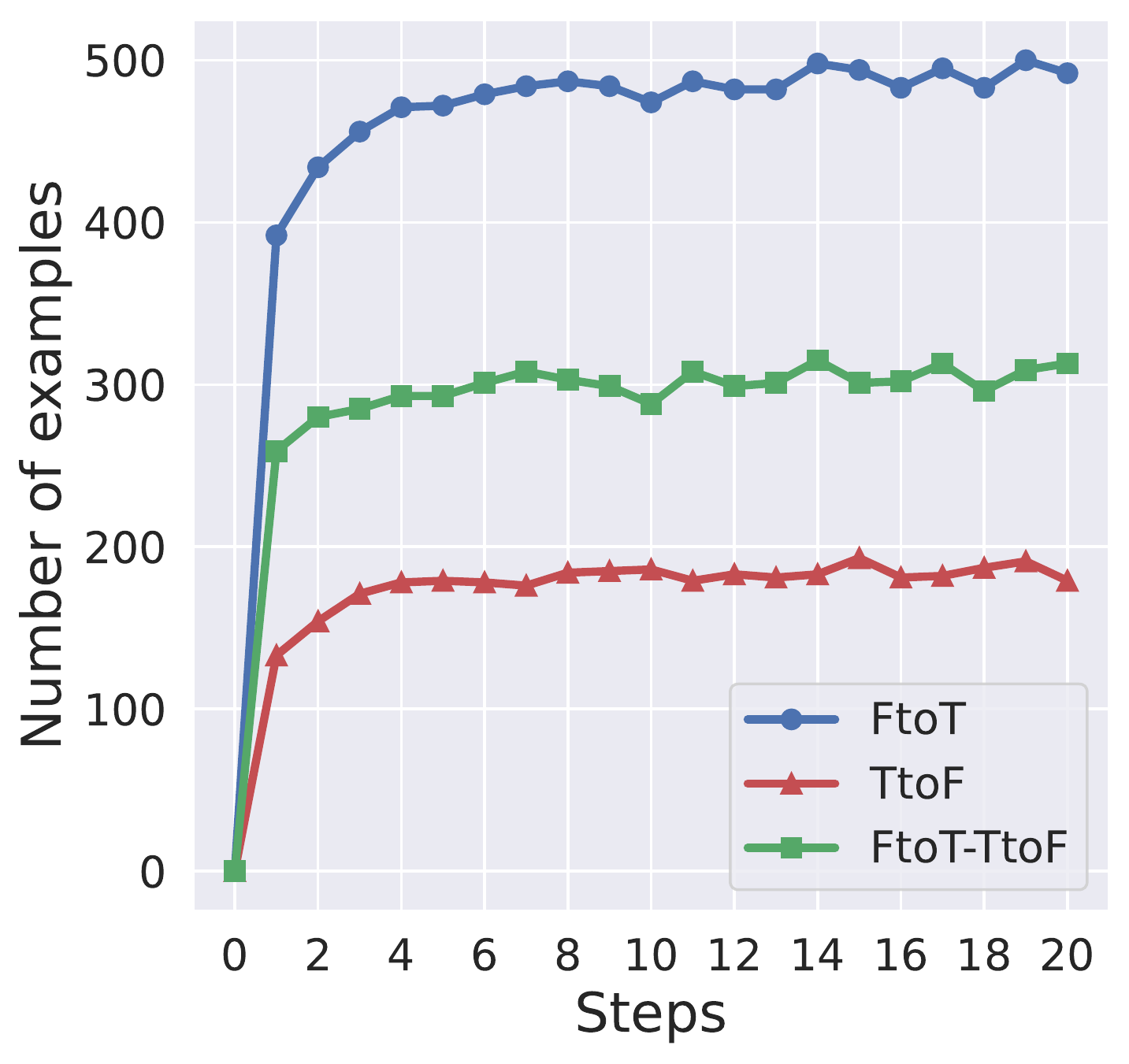}}
		\subfigure[]{
			\includegraphics[width=0.225\linewidth]{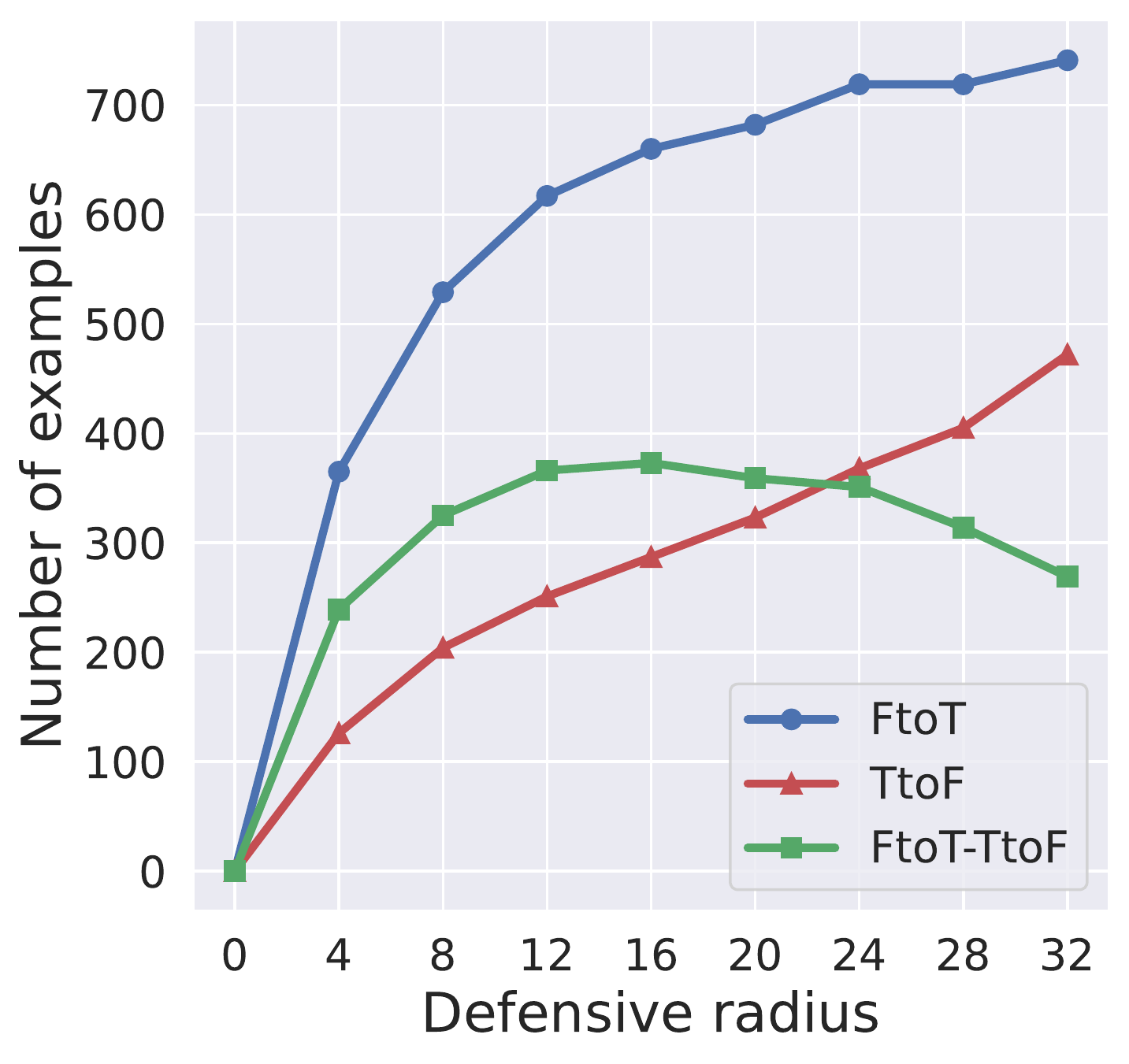}}
		\subfigure[]{
			\includegraphics[width=0.23\linewidth]{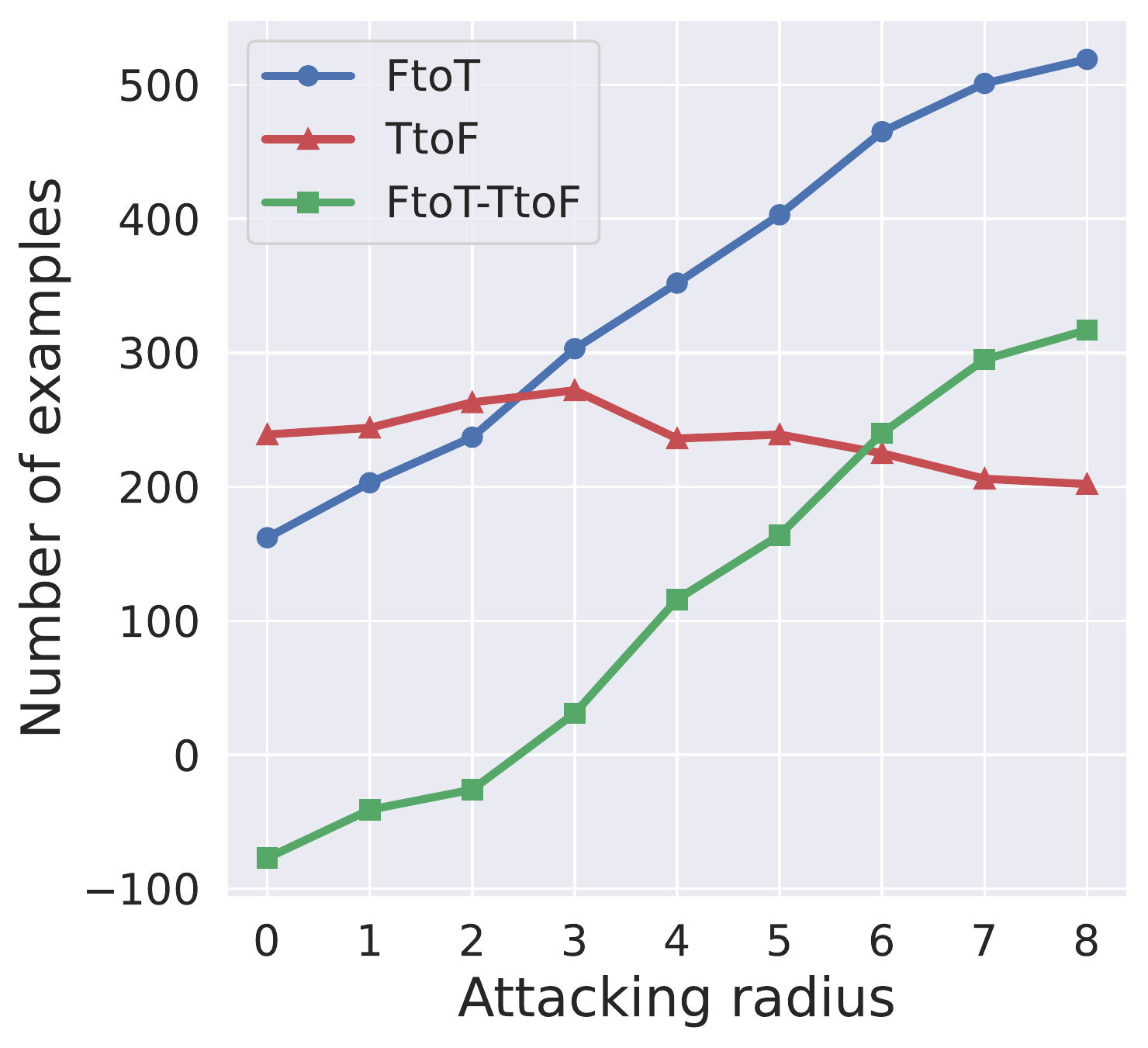}}
		\subfigure[]{
			\includegraphics[width=0.25\linewidth]{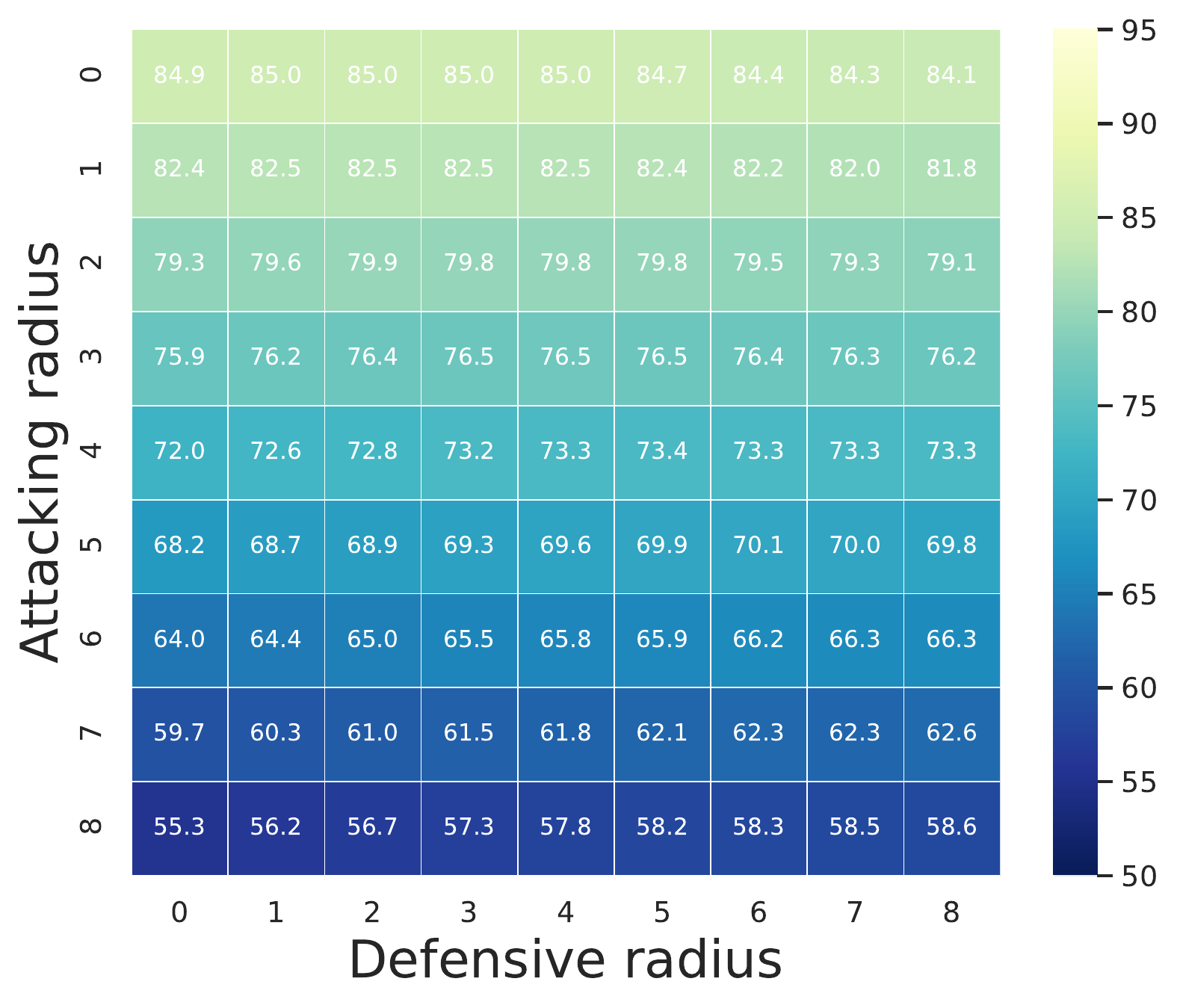}}
	
		\caption{A PGD version for Figure~\ref{fig:three}. }
		\vskip -0.1in
\end{figure}

\begin{figure}[h!]
	\centering
	\includegraphics[width=0.85\linewidth]{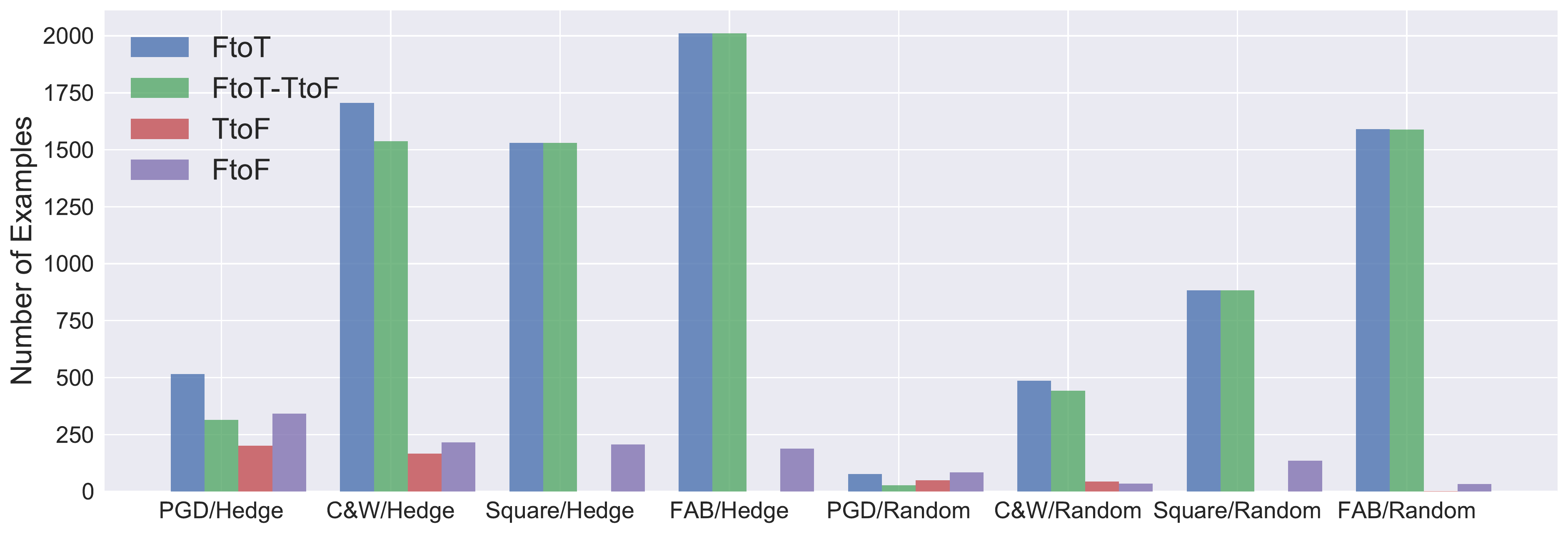}
	\caption{Extra results for Figure~\ref{fig:trade-off}}
    \label{fig:fig8}
\end{figure}

\subsection{Directly Attacking the Predictions of Adversarial Examples}

\begin{figure}[h!]
	\centering
	\includegraphics[width=0.6\linewidth]{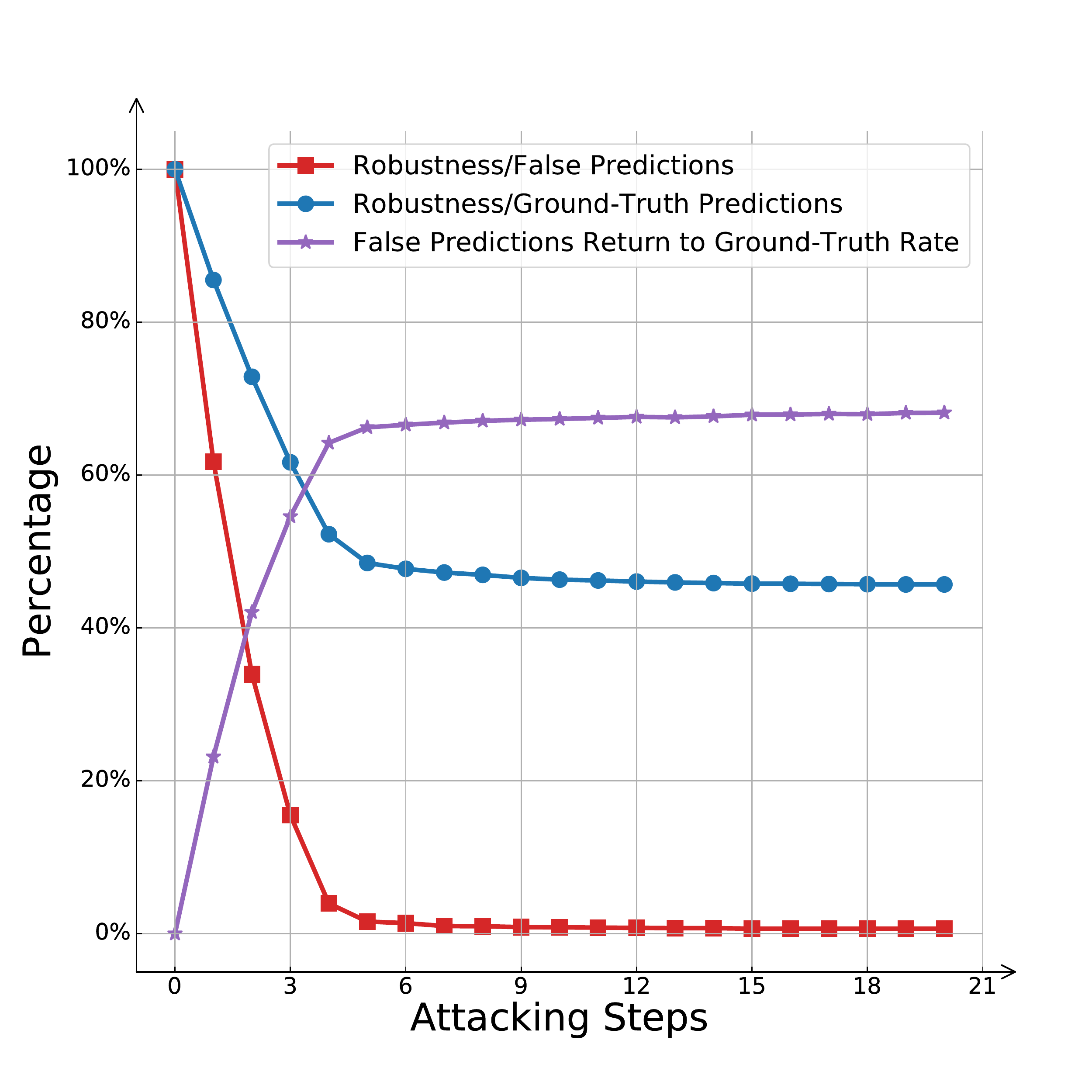}
	\caption{Attacking the predictions of adversarial examples.}
	\vskip -0.1in
\end{figure}

In this section, we provide empirical evidence that can help understand and verify our claims in Section~\ref{sec:intro}: On adversarially-trained models, the unprotected false classes are more vulnerable to attacks than the protected ground-truth class.

On the CIFAR10 test set, for the 8492 naturally accurate examples of the official TRADES model, we use PGD attack to generate their adversarial examples. 2908 cases are successfully attacked and generate false predictions, while 5584 of them withstand the attack and produce correct predictions. If we further attack the predictions of adversarial examples, 2889 out of the 2908 cases (red) will be successfully attacked, whereas 66\% of them (1928/2908, purple) will return to the correct predictions. In contrast, the 5584 adversarial examples with correct predictions can better resist attacks (blue line), with only 2551 examples being successfully attacked.

\subsection{Attempts to Attack Hedge Defense}
\label{app:attempt-attack}

In the evaluations of the main content, the adversarial attacks directly proceed on the static model without any defensive perturbation. We test an intuitive design that intends to counter Hedge Defense: On each step of attacks, we add a single-step defensive perturbation on the adversarial example after the attacking perturbation so that the attacks may explore the more complex cases of hedge examples. However, such an approach will hugely affect the attacking process and increase the robust accuracy to more than $80\%$.

Next, we consider the possibility of directly attacking the combination of Hedge Defense (as a pre-processing) and the deep networks. We adopt a simplified denotation to help readers better understand this process. Denote the feed-forward process in Algorithm~\ref{alg:hd} as $h(\xb''_k)=\sum_{c=1}^{C} \cL(f(\xb''_{k}),c)$, and the corresponding first order derivative with respect to the input as $\partial h(\xb''_k)/\partial \xb''_k$. Then the unified predictor $H(\xb')$ of deep networks and our Hedge Defense as a pre-processing can be formulated as:
\[
H =  f \circ M(\xb')
\],
where $M$ is a pre-processing with  $M(\xb') = \xb'+\sum_{k=1}^{K}\frac{\partial h(\xb''_k)}{\partial \xb''_k}$

\begin{equation}
\small
H(\xb')=
f(\xb'+\sum_{k=1}^{K}\frac{\partial h(\xb''_k)}{\partial \xb''_k})
=f(\xb''_K)
.
\end{equation}
We omit components like $\sign(\cdot)$ and the projection operation $\Pi$ in Algorithm~\ref{alg:hd}, which will not affect our conclusion. Then, a direct attack should compute the first order derivative of the above function with respect to the input $\xb'$.
\begin{equation}
\begin{aligned}
\small
\frac{\partial \cL(H(\xb'),y)}{\partial \xb'}
&=
\frac{\partial \cL(f(\xb''_K),y)}{\partial \xb'}
\\
&=
\frac{\partial \cL(f(\xb''_K),y)}{\partial \xb''_K}
\cdot
\frac{\partial \xb''_K}{\partial \xb'}
\\
&=
\frac{\partial \cL(f(\xb''_K),y)}{\partial \xb''_K}
\cdot
\frac{\partial (\xb'+\sum_{k=1}^{K}\frac{\partial h(\xb''_k)}{\partial \xb''_k})}{\partial \xb'} 
\\
&=
\frac{\partial \cL(f(\xb''_K),y)}{\partial \xb''_K}
\cdot
(
1+
\sum_{k=1}^{K}\frac{\partial^{2} h(\xb''_k)}{\partial \xb''_k \partial \xb'} 
)
.
\end{aligned}
\end{equation}

Empirical evaluations show that the magnitude of $\sum_{k=1}^{K}\frac{\partial^{2} h(\xb''_k)}{\partial \xb''_k \partial \xb'} $ is extremely small but its computation can be very expensive. Therefore, we omit this part and get:
\begin{equation}
\small
\frac{\partial \cL(H(\xb'),y)}{\partial \xb'}
\approx
\frac{\partial \cL(f(\xb''_K),y)}{\partial \xb''_K}
.
\end{equation}

This means, on each attacking step of our direct attack on Hedge Defense, instead of computing the derivative with respect to $\xb'$, one should first conduct the iteration in Algorithm~\ref{alg:hd} and then apply the computed gradient with respect to $\xb''_K$ to $\xb'$:

\begin{algorithm}[H]
\small
\caption{\ Attack Hedge Defense}
 \begin{algorithmic}[1]
    \STATE {{\bfseries Input}: the coming input $\xb$,  the number of defensive iterations $K$,  the number of attacking iterations $T$, the step size $\eta$,  the deep network $f(\cdot)$, the attacking radius $\epsilon_\text{a}$, and the defensive radius $\epsilon_\text{d}$.}
    \STATE{\emph{// $ \cU(\mathbf{-1},\mathbf{1}) $ generates a uniform noise}}
    \STATE {{\bfseries Initialization}: $\xb'_0 \gets \xb + \epsilon_\text{a} \ \cU(\mathbf{-1},\mathbf{1}) $}.
    \FOR{$t = 1 \dots T$}
    \STATE{$\xb''_{(t-1),0} \gets \xb'_{t-1} + \epsilon_\text{d} \ \cU(\mathbf{-1},\mathbf{1})$}
    \FOR{$k = 1 \dots K$}
    \STATE{$\xb''_{(t-1),k} \gets \xb''_{(t-1),(k-1)} + $ \\ $ \quad \eta\cdot\sign(\nabla_{\xb''_{(t-1),(k-1)}}\sum_{c=1}^{C} \cL(f(\xb''_{(t-1),(k-1)}),c))$;}
    \STATE{\emph{// $\Pi$ is the projection operator.}}
    \STATE{$\xb''_{(t-1),k} \gets \Pi_{\mathbb{B}(\xb',\epsilon_\text{d})} (\xb''_{(t-1),k})$;}
    \ENDFOR
    \STATE{$\xb'_t \gets \xb'_{t-1} +  \quad \eta\cdot\sign(\nabla_{\xb''_{(t-1),K}} \cL(f(\xb''_{(t-1),K}),y))$;}
    \STATE{$\xb'_t \gets \Pi_{\mathbb{B}(\xb,\epsilon_\text{a})} (\xb'_t)$;}
    \ENDFOR
    \STATE {{\bfseries Output}: the adversarial example $\xb'_T$.}
 \end{algorithmic}
\end{algorithm}

We test the above algorithms on RST~\cite{rst}. As shown by the table below, the defense-aware attack affects Hedge Defense and decreases its improvement by $0.6\%$. Making the total improvement of Hedge Defense becomes $65.78\%-63.17\%=2.61\%$. Meanwhile, the attacking performance of the defense-aware attack on the Direct Prediciton, predicting the generated adversarial examples without the defensive perturbation, hugely degrades. Moreover, the robust accuracy rises to $67.08\%$. We will investigate this defense-aware attack in future work.

\begin{table}[h!]
	\caption{Defense-aware attack on Hedge Defense.}
	\centering
	\vskip 0.05in
	\resizebox{\textwidth}{!}{
	\begin{tabular}{ cccc } 
		\toprule
		Model 
		& Attack 
		& Direct Prediction (\%)
		& Prediction of Hedge Defense (\%)\\
		\midrule
		\multirow{2}{*}{RST~\cite{rst}} 
		& Attack the Model
		& 63.17 & 66.38 \\ 
		& Attack Hedge Defense
		& 67.08 & 65.78 \\ 
		\bottomrule
	\end{tabular}}
	\vskip -0.1in
\end{table}

\subsection{About Adopted Assets}
\label{app:liscense}

We provide our codes in the supplementary material. The license of the mainly used resources, RobustBench~\cite{robustbench}, can be found at:

\url{https://github.com/RobustBench/robustbench/blob/master/LICENSE}

\subsection{Clarification on Ethical Concerns}
\label{app:ethical}

All assets adopted by this work are standard benchmark sets and do not relate to ethical issues like containing personally identifiable information or offensive content, acquiring data without the consent of the owners.

\newpage

\section{More Interpretations on the Linear Model}
\label{app:more-theory}

\subsection{Excluding the Non-Robust Examples in $(-\epsilon,\epsilon)$}

In Section~\ref{sec:theory}, we study Hedge Defense on a binary classification problem, where a robust solution exists (the ground-truth classifier). This design aims to simulate real-world tasks like image classification, where we assume that there is a robust solution~\cite{Stutz19}. Many works aim at analyzing the causes of the intriguing adversarial examples on deep neural networks and suggest that deep networks are non-robust against attacks due to poor generalizations~\cite{robustbench}. Others~\cite{Stutz19} find that improving adversarial robustness may implicitly help the generalization of deep networks. \cite{fgsm} also points out that, on the logistic regression model, adversarial training will degenerate to the LASSO regularization. Our work advocates the opinions in these works. It can be told that, in the problem we construct in Section~\ref{sec:theory}, the non-robustness comes from learning a biased classifier instead of the ground-truth one. Thus, delving into its mechanism, it is still a problem focusing on model generalization. In addition, there is another series of theoretical works focusing on studying the intrinsically non-robust examples~\cite{rst,tes19,trades}. In their framework, the ground-truth classifier is a non-robust one. Thus, for any classifier, the robust errors of these models have a lower bound~\cite{rst} that is larger than zero, and the non-robust examples can never be eliminated.

\begin{figure*}[h!]
		\centering
		\subfigure[]{
			\label{fig:x1}
			\includegraphics[width=0.18\linewidth]{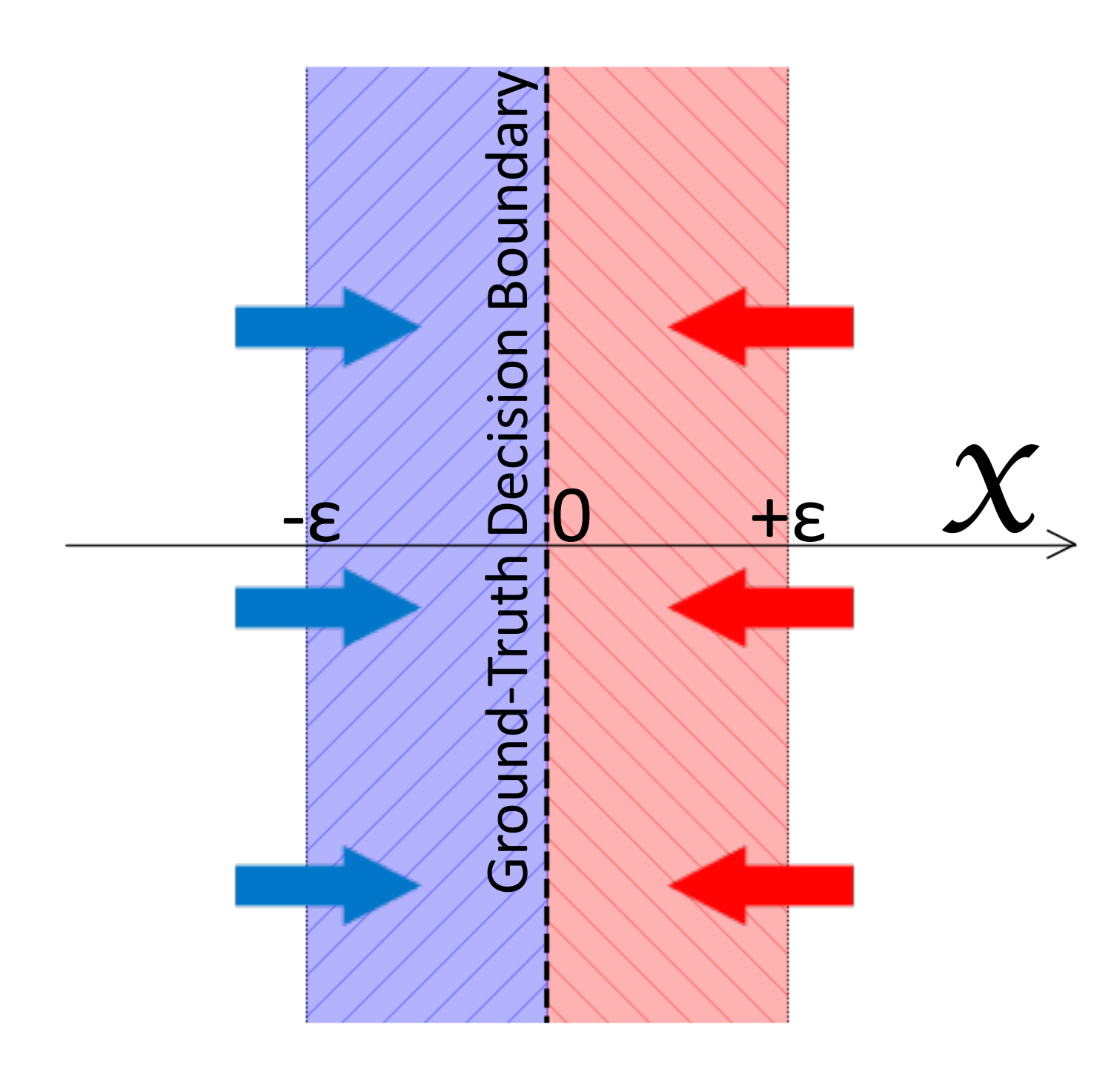}}
		\subfigure[]{
			\label{fig:x2}
			\includegraphics[width=0.18\linewidth]{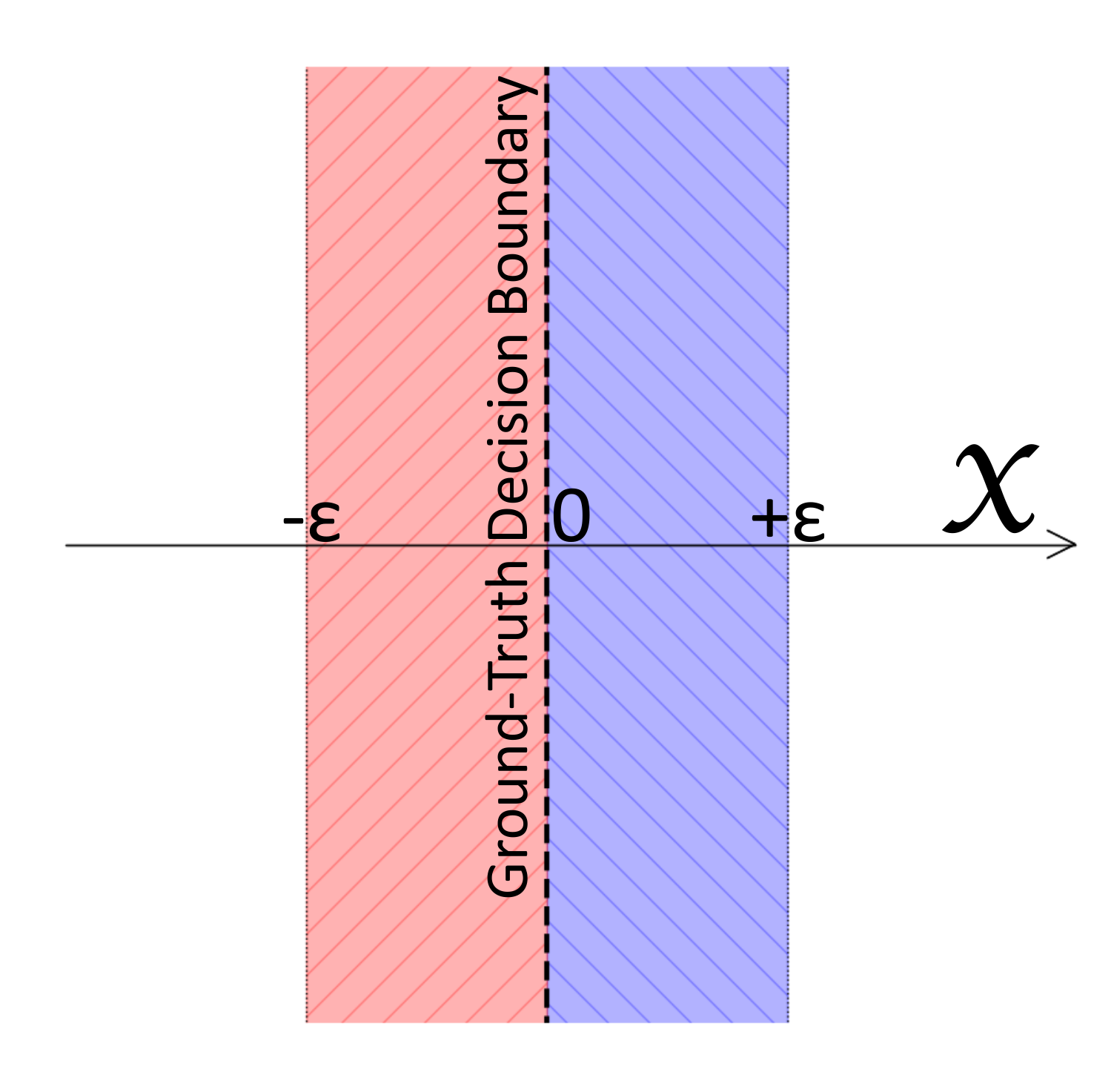}}
		\subfigure[]{
			\label{fig:x3}
			\includegraphics[width=0.18\linewidth]{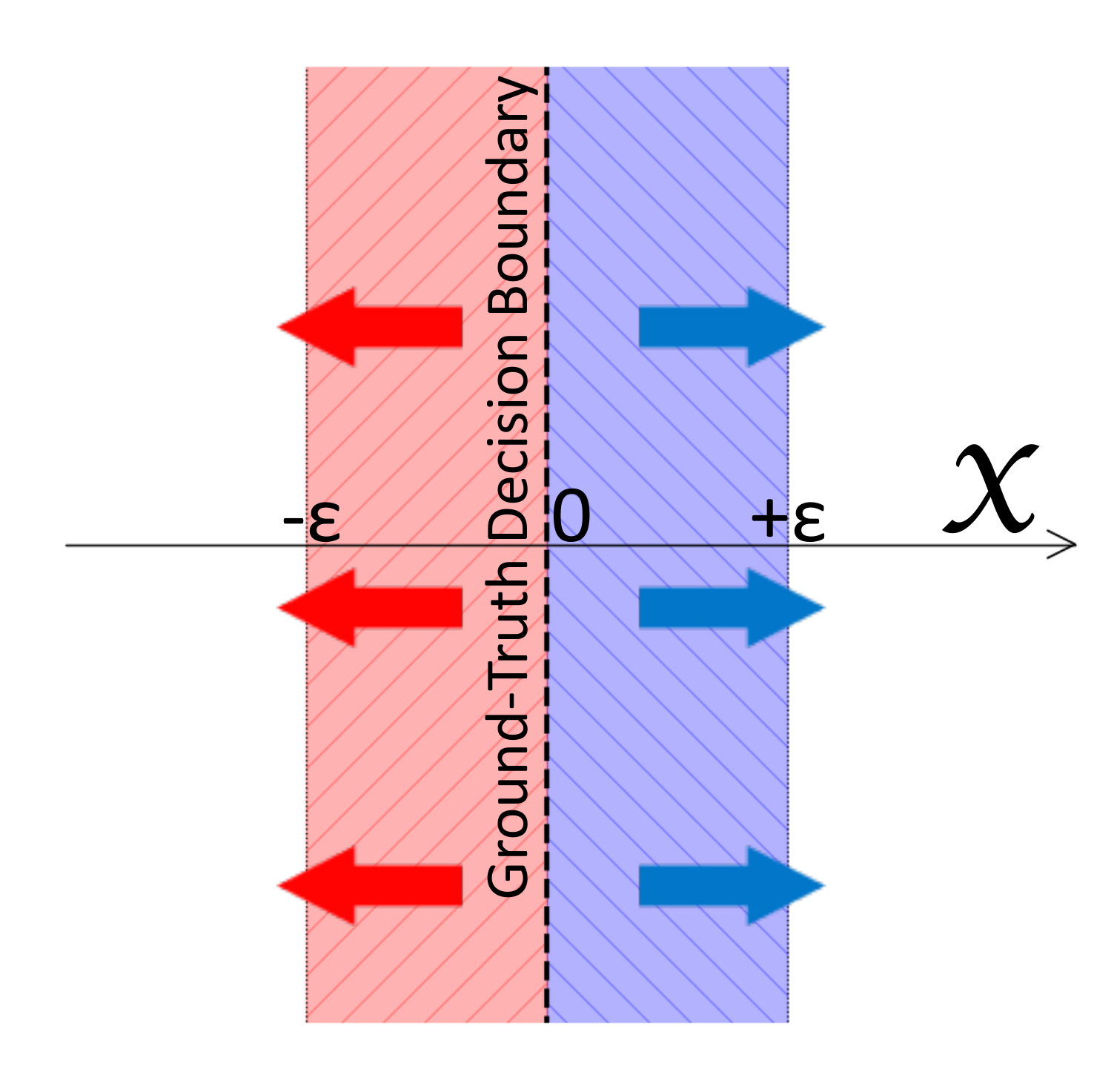}}
		\subfigure[]{
			\label{fig:x4}
			\includegraphics[width=0.18\linewidth]{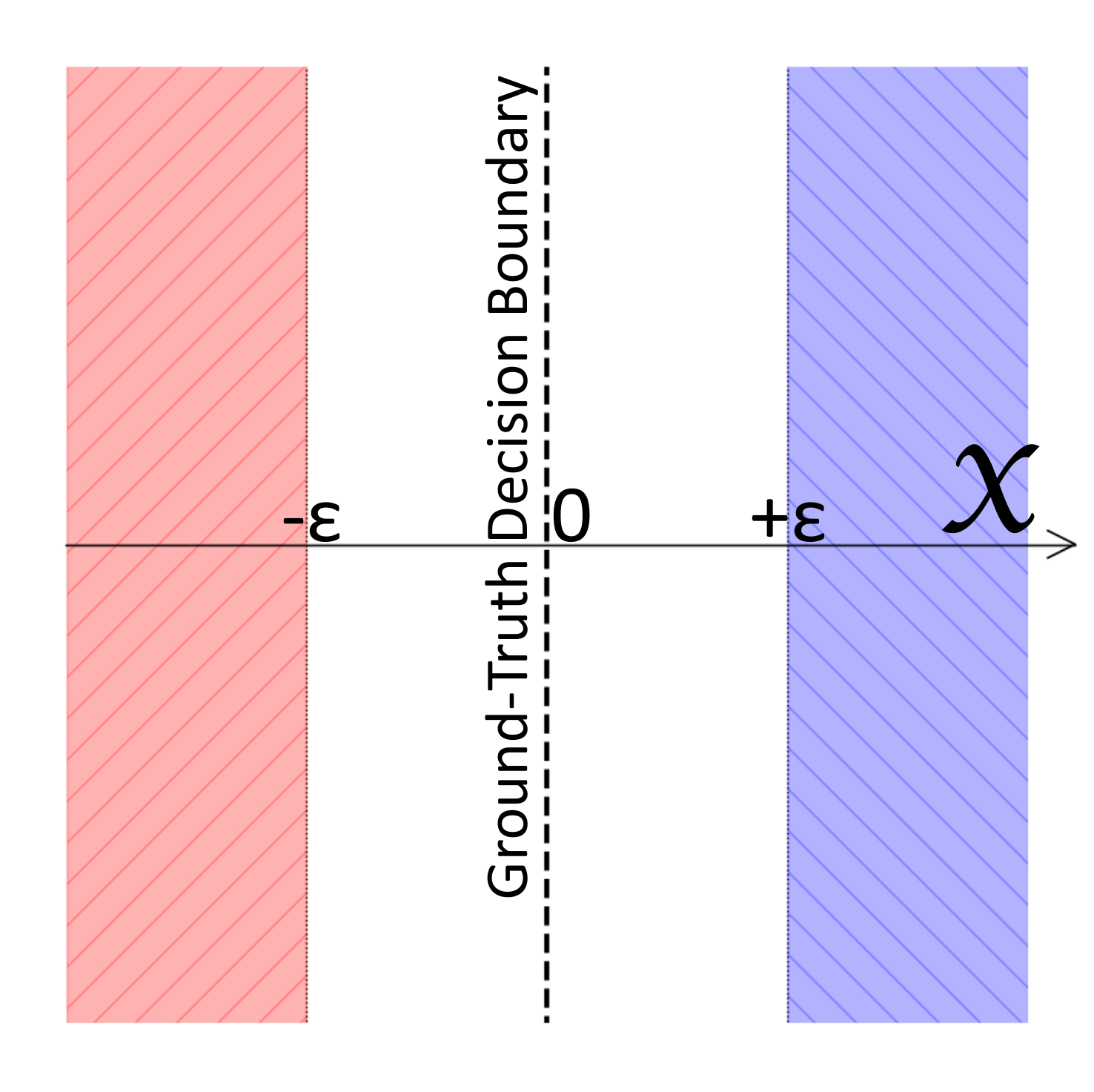}}

		\caption{Interpretation of Hedge Defense on the intrinsically non-robust examples.}
		\vskip -0.1in
\end{figure*}

Finally, we examine what will happen to the intrinsically non-robust examples in $(-\epsilon,\epsilon)$. As shown in Figure~\ref{fig:x1}, we apply the FGM attack to the examples in $(-\epsilon,\epsilon)$. In Figure~\ref{fig:x2}, these examples are pushed across the ground-truth decision boundary and thus wrongly classified. Then, if we further apply Hedge Defense on these attacked adversarial examples, the generated defensive perturbation will have the same direction as the attacking perturbation (Figure~\ref{fig:x3}), pushing examples further away from their original positions (Figure~\ref{fig:x4}). Notice that, for these examples, Hedge Defense does not increase their errors since the examples have been already wrongly classified in Figure~\ref{fig:x3}. All the analyses are conducted on the ground-truth classifier, and there is nothing we can do about these examples because no classifier can separate whether the examples within $(-\epsilon,\epsilon)$ are adversarial (Figure~\ref{fig:x2}) or natural (Figure~\ref{fig:x1}). Namely, suppose the classifier receives an input as $x'=\epsilon/2$. This $x'$ can either be an adversarial example that originally locates at $x'=-\epsilon/2$ with label $y=-1$, or a natural example that stays at $x'=\epsilon/2$ all the time with label $y=+1$. Without extra information, a classifier cannot separate these two cases and output different labels for both of them.

\subsection{Hedge Defense for Natural Examples}

\begin{figure*}[h!]
		\centering
		\subfigure[]{
			\label{fig:y1}
			\includegraphics[width=0.22\linewidth]{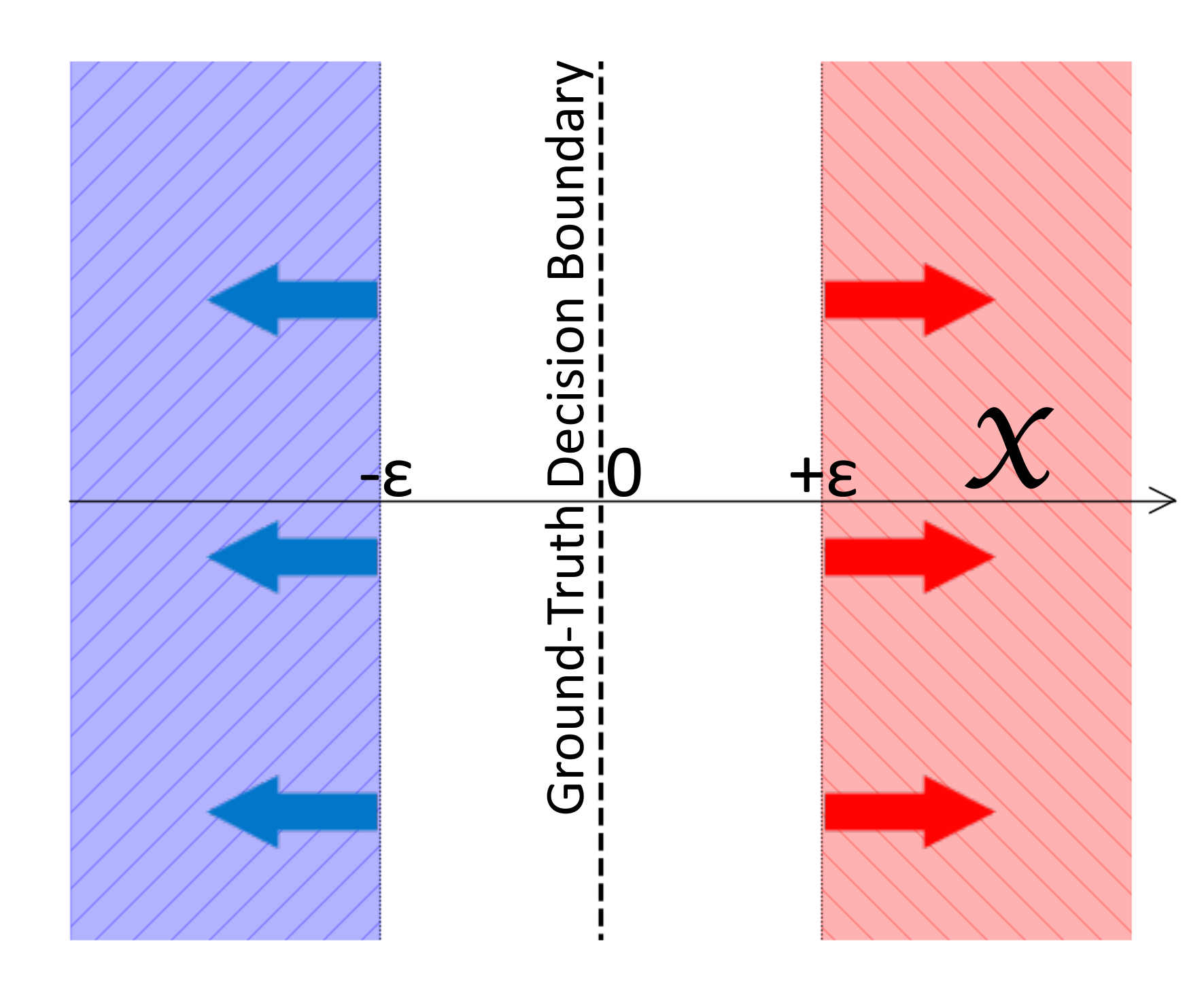}}
		\subfigure[]{
			\label{fig:y2}
			\includegraphics[width=0.22\linewidth]{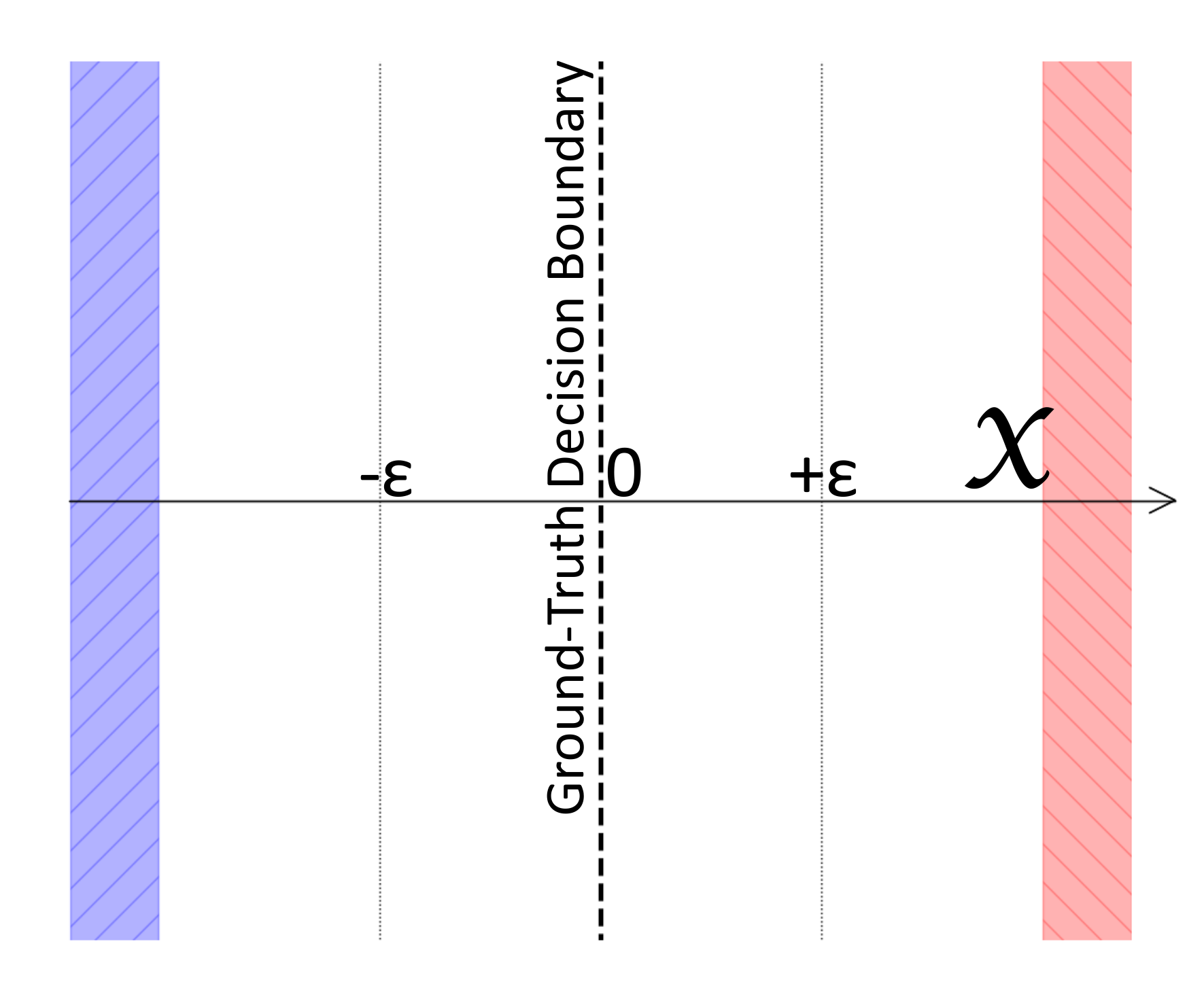}}
		\caption{Interpretation of Hedge Defense on the natural examples.}
		\vskip -0.1in
\end{figure*}

In Section~\ref{sec:theory}, we apply Hedge Defense after the FGM adversarial attack. Now we examine the case where we directly apply Hedge Defense to natural examples. As shown in Figure~\ref{fig:y1}, the generated defensive perturbation stays the same as those in Figure~\ref{fig:4e} and still pushes examples away from the decision boundary (Figure~\ref{fig:y2}). Therefore, the accuracy of natural examples is not changed, and both the ground-truth and the estimated classifier are still naturally accurate.

\subsection{For the Cases When $\zeta>\epsilon$}

\begin{figure*}[h!]
		\centering
		\subfigure[]{
			\label{fig:z1}
			\includegraphics[width=0.18\linewidth]{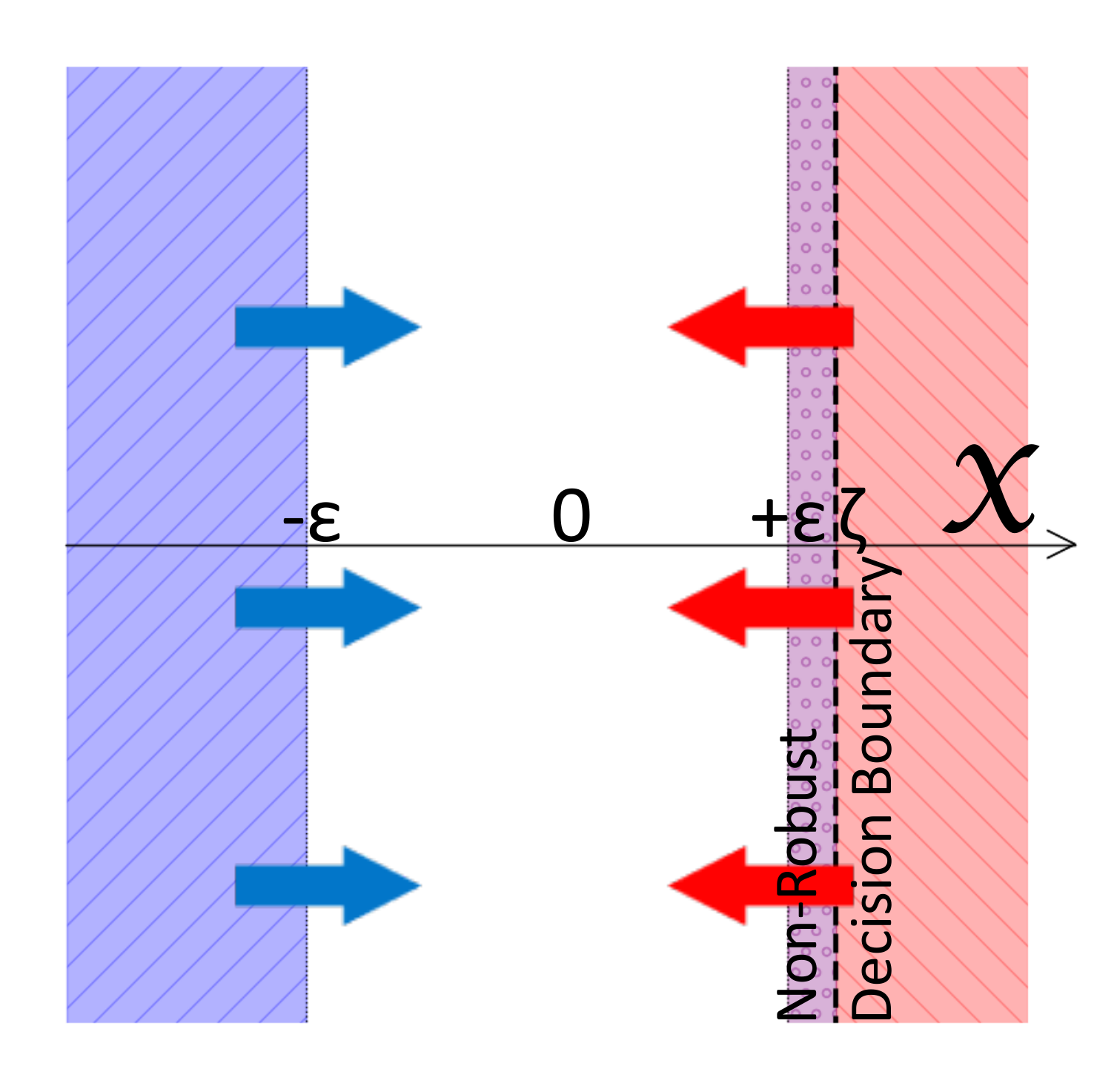}}
		\subfigure[]{
			\label{fig:z2}
			\includegraphics[width=0.18\linewidth]{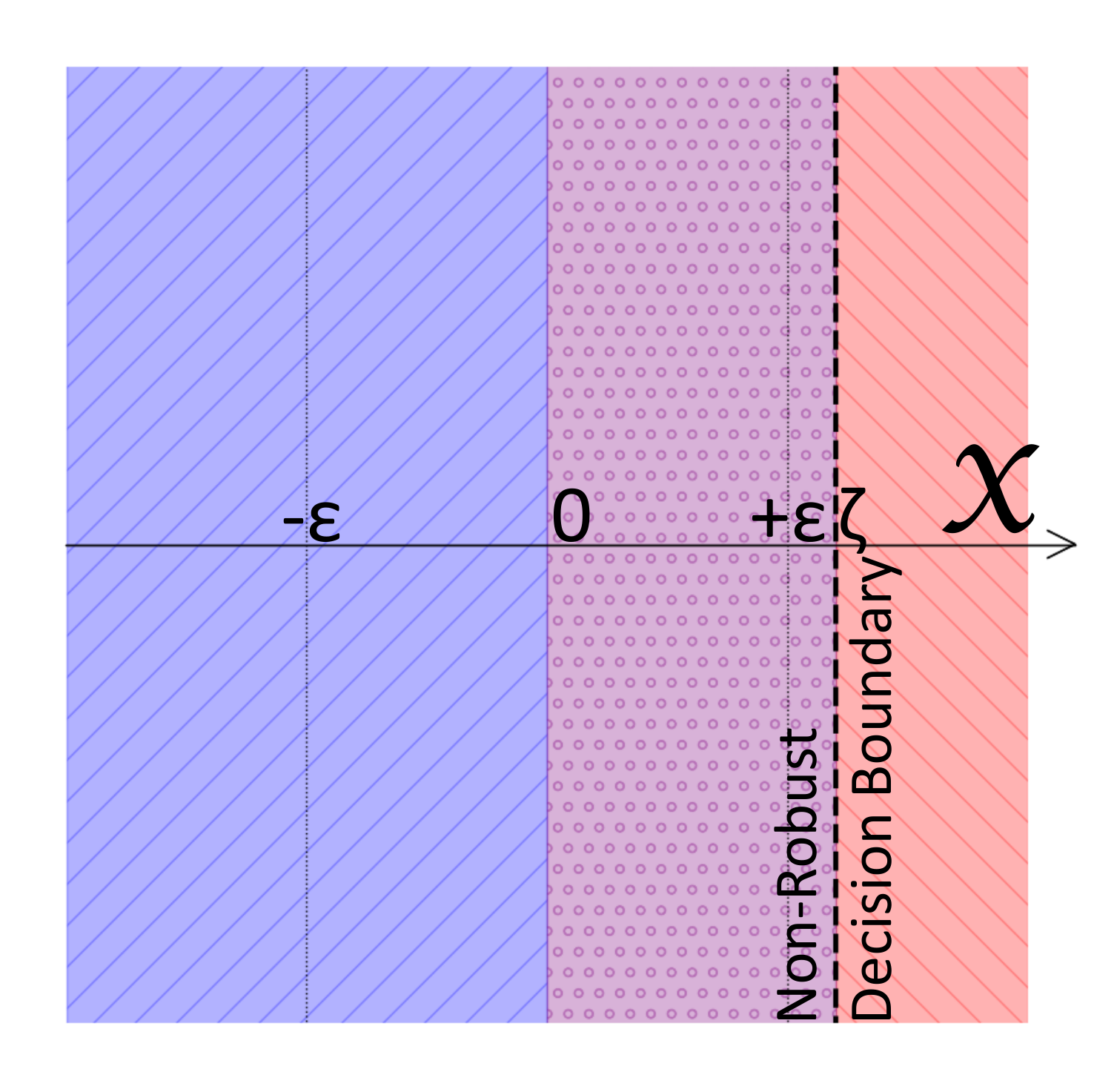}}
		\subfigure[]{
			\label{fig:z3}
			\includegraphics[width=0.18\linewidth]{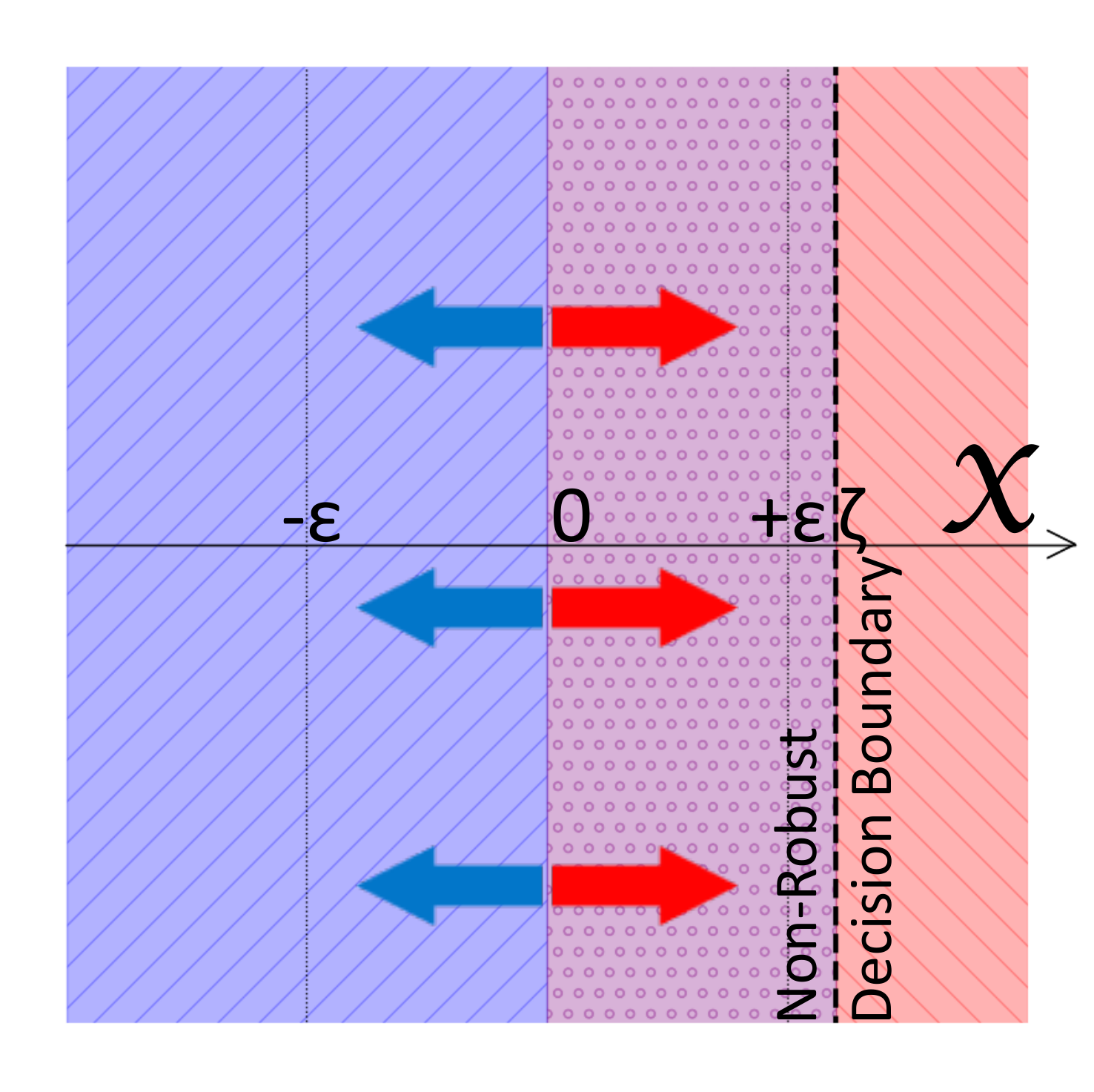}}
		\caption{Interpretation of Hedge Defense when $\zeta>\epsilon$.}
		\vskip -0.1in
\end{figure*}

In Section~\ref{sec:theory}, we assume the learned bias is smaller than $\epsilon$ so that the estimated classifier is a naturally accurate discriminator. On the other hand, when $\zeta>\epsilon$, Hedge Defense can still take effect. As shown in Figure~\ref{fig:z1}, the examples within $(\epsilon,\zeta)$ are wrongly classified before attacks happen. After attacks (Figure~\ref{fig:z2}), the width of the purple region represents the robust errors. The examples within $(0,\epsilon)$ are wrongly classified because of attacks, while the examples within $(\epsilon,\zeta)$ are wrongly classified because of failure to generalization~\cite{mart}. If we apply Hedge Defense on the adversarial examples (Figure~\ref{fig:z3}), the generated defensive perturbation will still push examples away from the decision boundary and recover examples to their positions in Figure~\ref{fig:z1}. Therefore, the robust errors from $(0,\epsilon)$ will be decreased. If the defensive perturbation can be larger than $\epsilon$, the natural errors in $(\epsilon,\zeta)$ can also be eliminated. Of course, such effectiveness of Hedge Defense establishes on the condition that $k_f \gg k_t > 0$. On adversarially-trained deep networks, examples far away from the natural point are not protected by adversarial training and no longer meet the condition of $k_f \gg k_t > 0$. Thus, the defensive radius of $\epsilon_\text{d}$ cannot be unrestrictedly large.

\subsection{Only the Estimated Function Needs to Meet $k_f \gg k_t > 0$}
In Section~\ref{sec:theory}, we require the ground-truth classifier to meet the condition that $k_f \gg k_t > 0$. However, Hedge Defense only requires that the estimated classifier meet this condition, and the resulted robustness can apply to any ground-truth classifier with a decision boundary at $x=0$, including the cases where $k_t \gg k_f > 0$ for the ground-truth classifier

\subsection{Multi-step Adversarial Attacks}
In Section~\ref{sec:theory}, we investigate the one-step attack and defense. However, since the model we studied is linear and we compute the gradient based on the Fast Gradient Method, which will apply a $\sign(\cdot)$ function to the computed gradient, the conclusion is that our analyses will not change on multi-step attacks.


\section{Detailed Discussions on the Closely-Related Works}
\label{app:more-related}

\subsection{Defensive Image Transformation~\cite{Guo18}}
\cite{Guo18} proposes defending deep networks with image transformations such as bit-depth reduction, JPEG compression, total variance minimization, and image quilting before feeding the image to the deep network. These transformations may wash off the influence of the adversarial perturbation. However, this method is not robust to white-box attacks, as shown in the next section.

\subsection{Random Image Pre-processing~\cite{Xie18}}
\cite{Xie18} tries to introduce random image pre-processing, including resizing and padding, to counter the effect of adversarial perturbations. They believe the randomness will prevent attacks from generating an effective universal attack. The authors consider three attack scenarios: vanilla attack (an attack on the original classifier), single-pattern attack (an attack assuming some fixed randomization pattern), and ensemble-pattern attack (an attack over a small ensemble of fixed randomization patterns). However, later works show that attackers can still optimize over the distribution of sampling and thus bypass the defense, as shown in the next section.

\subsection{Dynamic Models~\cite{rmc}}
\cite{rmc} proposes to dynamically optimize the model for a few steps before predicting during the testing stage to dodge a direct attack from the adversarial examples. To properly optimize the model, they need to collect testing data to form a small database so that later coming inputs can be properly optimized. They empirically verify their improvements on attacks like PGD and C\&W. In contrast, our Hedge Defense has a different, clear, and explainable optimization goal, allowing us to conduct partial theoretical analyses in Section~\ref{sec:theory}.

\subsection{Uniformly Perturbing Inputs as a Defense~\cite{PalV20}}

We here briefly introduce the theory that is built up in \cite{PalV20}. For the input space $\cX$, the label of any $\xb \in \mathcal{X}$ is defined by the groundtruth function $\sign(f(\xb)) \in \{-1,+1\}$. In \cite{PalV20}, $f(\xb)$ is assumed to be locally linear over $\cX$, which means for $\xb'$ that is within the $2\epsilon$ neighborhood of $\xb$, we have $f(\bx')=f_L(\bx')=f(\bx)+\nabla f(\bx)^{\top}(\bx'-\bx)$. Then, both attackers and defenders are allowed to conduct a single-shot perturbation 
simultaneously, and the utility of attackers $u_A$ can be defined as:
\begin{equation}
u_A(\xb, a(\xb), d(\xb)) = 
\begin{cases}
+1 \qquad \text{if} \ \sign(f_L(\bx)) \neq \sign(f_L(\bx+a(\bx)+d(\bx))), \\
-1 \qquad \text{otherwise},
\end{cases}  .
\end{equation}
$a(\xb)$ and $d(\xb)$ are respectively the perturbations from attackers and defenders, where $a(\xb),d(\xb) \in \cV $ and $\cV:=\{\vb: \|\vb\|_2 \leq \epsilon\}$.
Meanwhile, the score of defenders $u_D$ is simply the negative of the utility of attackers: $u_D(\bx, a, d)=-u_A(\bx, a, d)$.

Define the set of attack strategies $\mathcal{A}_A$ to be a function set $\{a|a:
\cX \to \cV \}$. Attacker are allowed to randomly draw strategies from $\cA_A$ and we define $\cP(\cA_A)$ to be the set of all probability densities over $\cA_A$. Similarly, we have $\cP(\cA_D)$ and $\cA_D$ on the defensive side. Then we have the following conclusion for the Fast Gradient Method attack: $a_\text{FGM}(x)=-\epsilon\frac{\sign(f(\xb))}{\|\nabla f(\xb)\|_2}\nabla f(\xb)$:
\begin{lemma}[Pal and Val \cite{PalV20}]
For any defensive strategy $s_D \in \cP (\cA_D)$, the FGM attack strategy $s_A \in \cP (\cA_A)$ achieves the largest utility, \ie, $\bar{u}_A(s_\text{FGM},s_D) \geq \bar{u}_A(s_A,s_D)$ for all $s_A \in \cP (\cA_A)$
\end{lemma}

Since defenders usually cannot get access to $\xb$, \cite{PalV20} investigate the scenario where the defensive strategies $\cA_D$ only consist of constant functions $\cA_D=\{d_\vb| \vb \in \cV, d_\vb:\cX \to \cV \ \text{s.t.} \forall \xb \in \cX, d_\vb(x) = \vb \}$. Then they find that the defense $s_\text{SMOOTH}$ that uniformly samples strategies from $\cA_D$ is an optimal response to $s_\text{FGM}$ and thus forms a Nash Equilibrium. The constant constraint is over strict and hinders the power of defenders, and \cite{PalV20} also stated that it is possible for defenders to design strategies when defenders are given $\xb+a(\xb)$. Thus, we consider to give defenders the same flexibility as the attackers and introduce our hedging defense in the next part.

Notice that the randomized smoothing discussed in \cite{PalV20} is different from the common concept of the randomized smoothing in \cite{rs}. In \cite{PalV20}, the random noise is sampled from the uniform distribution and is adopted for improved robustness. While in \cite{rs}, the Gaussian noise is used to estimate how certificated the robustness is.

The above work proves that if the defensive perturbation has to be generated simultaneously with the attacking perturbation, meaning the defensive perturbation is irrelevant to the input or adversarial examples, the uniform perturbation is an optimal solution under a relatively weak assumption (input locates within a locally-linear space). Also, it proves that the gradient-based attacks (PGD) are also the optimal attacks, enabling uniform perturbation to become a solid one. This work complements our findings, and Hedge Defense explores the scenario where the defensive perturbation is generated after the attacks and can get access to the input.

\section{Hedge Defense Does Not Belong to the Obfuscated Gradient Category}
\label{app:not-obfuscated}

A series of defenses termed the gradient masking intend to counter attacks by hiding the existence of adversarial examples. Namely, they try to make the gradient of the model chaotic so that standard attacks cannot properly search the adversarial examples even if they still exist. Particularly, the obfuscated gradient phenomena~\cite{obfuscated} is a special case of gradient masking. \cite{obfuscated} identifies that 7 out of the nine most powerful defenses are actually caused by the obfuscated gradient and thus breakable. From then on, whenever a new defense comes out, researchers may wonder is this new defense just another case of the obfuscated gradient. This section will show that our Hedge Defense or random perturbation does not fall into the category of the obfuscated gradient. Specifically, we first revisit the five characteristic behaviors of the defenses from the obfuscated gradient category and show that Hedge Defense does not have any of these behaviors. The five characteristic behaviors are proposed in Section3.1 of the original work~\cite{obfuscated}. Then, we investigate the three categories of the obfuscated gradient and demonstrate why Hedge Defense does not belong to any of them. Finally, we revisit the case study section (Section5) in the original work~\cite{obfuscated}. In that part, \cite{obfuscated} analyzes each of the seven invalid defenses and demonstrates how to break them. We will illustrate why these adapted attacks do not take effect on our Hedge Defense.

\subsection{The Five Characteristic Behaviors}

\paragraph{One-step attacks perform better than iterative attacks.}
We apply Hedge Defense against the one-step FGSM~\cite{fgsm} attack for the official model of TRADES on CIFAR10. The corresponding robust accuracy is $68.21\%$, much higher than that for the multi-step PGD attack ($58.99\%$). Thus, Hedge Defense does not have this behavior.

\paragraph{Black-box attacks are better than white-box attacks.}
In Table~\ref{table:cifar10}, all the robust accuracies of standard white-box attacks (PGD, C\&W) are much lower than those of the black-box attacks (Square, RayS), indicating white-box attacks are better than black-box attacks when facing Hedge Defense. Thus, Hedge Defense does not have this behavior.

\paragraph{Unbounded attacks do not reach 100\% success.}
We cancel the bound of the attacks ($\epsilon_\text{a}$) and do observe $100\%$ success of attacks. Notice that Hedge Defense is also bounded by $\epsilon_\text{d}$ so that it can be effortless to understand that Hedge Defense cannot substantially modify examples that are far away from the natural ones. Thus, Hedge Defense does not have this behavior.

\paragraph{Random sampling finds adversarial examples.}
We randomly sample examples in $\mathbb{B}(\xb,\epsilon_\text{a})$ and then apply Hedge Defense on them. The resulted accuracy is close to the one with the random perturbation in Table~\ref{table:cifar10}. Thus, Hedge Defense does not have this behavior.

\paragraph{Increasing distortion bound does not increase success.}
As shown in Figure~\ref{fig:noe-ar}, the increasing distortion bound ($\epsilon_\text{a}$) does increase attack successful rate. Thus, Hedge Defense does not have this behavior.

\subsection{The Three Types of Obfuscated Gradient}

\cite{obfuscated} divides the obfuscated gradient problem into three categories: \textbf{shattered gradients, stochastic gradients, and vanishing/exploding gradients}. Shattered gradients refer to the cases where the model cannot be directly differentiated. Stochastic gradients mean the defensive model has a random component, which will prevent the computation of gradients. Vanishing/exploding gradients refer to the cases in which the gradient is either too small or too large, causing the attacking direction to become inaccurate.

For the random perturbation, the shattered gradients problem and the vanishing/exploding gradients do not apply. For the stochastic gradients, we test the Expectation over Transformation (EOT) method of \cite{obfuscated} on the random perturbation model. EOT is designed to counter the potential randomness of a defensive model. Empirical results show that the EOT implementation of AutoAttack~\cite{autoattack} does not affect the performance of randomly perturbing the coming input. After all, the EOT method is not guaranteed to break all the potential randomness, and \cite{PinotERCA20} also shows that extra randomness can be helpful for robustness. For Hedge Defense, the stochastic gradients problem and the vanishing/exploding gradients do not apply. For the shattered gradients, we have tested the Backward Pass Differentiable Approximation (BPDA) method of \cite{obfuscated} in Appendix~\ref{app:attempt-attack} and show that Hedge Defense can withstand this examination.

\subsection{The Case Study of \cite{obfuscated}}

We study two most relevant previous works that have been proven to be ineffective because of the obfuscated gradient, defense via randomness~\cite{Xie18} and defense via pre-processing~\cite{Guo18}. For \cite{Xie18}, \cite{obfuscated} counters the randomness with EOT because the random module has a constrained distribution. Nevertheless, compared with the random resizing and padding, the uniform noises on adversarial examples are very primitive, leaving attackers with little room to optimize. \cite{Guo18} is also a pre-processing defense like Hedge Defense. \cite{obfuscated} breaks the shattered gradients problem by designing a differentiated approximate module (Backward Pass Differentiable Approximation) and acquiring alternative gradient information for attacks. However, we have shown in Appendix~\ref{app:attempt-attack} that our Hedge Defense can withstand the BPDA and still bring a benefit to robustness.

\section{Extra Discussions and Understandings}
\label{app:extra-discu}

Here we want to offer extra discussions and highlight the contributions of our work. 1) Considering the current development of adversarial robustness, proving that a defensive method is unbreakable is still infeasible. Even for the commonly believed powerful adversarial training methods, people are still unsure about its robustness against the upcoming attacks in the future. Thus, to support the robustness of our method, we provide evidence from various angles with our best effort, and all these evidences indicate that our defensive perturbation can withstand the existing attacks. 2) This work does not solely focus on developing a defense method. That is, our finding of the non-robustness feature of adversarial attacks is important as well, which indicates that there is still room for improvement in existing attacks. 3) Although we have not proven the certificated robustness of Hedge Defense, the random perturbation on adversarial examples has been thoroughly verified in \cite{PalV20}. Future works may consider this simple method as a standard manner of evaluating robust models. 4) Directly learning a robust model without any defensive pre-processing, \ie, Hedge Defense, is still more ideal and appealing. However, Hedge Defense provides a new perspective to examine the existing seemingly powerful adversarial attacks and help researchers better understand the competition between defenses and attacks. Future works that focus on the adversarial Game Theory~\cite{PinotERCA20,PalV20} may benefit from our findings and methods.




\end{document}